\documentclass[11pt]{article} % OR requires 11pt
\usepackage{amsmath,amsfonts,amsthm,amssymb,bbm}
\usepackage{graphicx}
\usepackage[figtopcap]{subfigure}
\usepackage{hyperref,setspace,natbib}
\usepackage{float}
\usepackage{booktabs,multicol,multirow}
\usepackage{cleveref}
\usepackage{threeparttable}
\usepackage{enumerate} %
\usepackage{bibunits}
\usepackage{xcolor}
\usepackage{hhline}
\usepackage{endnotes}
\let\footnote=\endnote
\usepackage{array}
\newcolumntype{C}[1]{>{\centering\let\newline\\\arraybackslash\hspace{0pt}}m{#1}}
\graphicspath{{Figures/}} 
\newtheorem{theorem}{Theorem}

\newtheorem{proposition}{Proposition}
\newtheorem{definition}{Definition}
\newtheorem{example}{Example}

\newtheorem{remark}{Remark}
\newtheorem{lemma}{Lemma}
\makeatletter

\newcommand{\Rmnum}[1]{\expandafter\@slowromancap\romannumeral #1@}
\makeatother
\newcommand{\possessivecite}[1]{\citeauthor{#1}'s (\citeyear{#1})}

\def\Halmos{\mbox{\quad$\square$}}

\def\proof#1{{\it #1}}
\def\endproof{}

\begin{document}

\title{~~~
% \\[-.75in]
\huge \bf On Consistency of Signature Using Lasso\thanks{The authors gratefully acknowledge Yacine Ait-Sahalia, Jose Blanchet, Steven Campbell, Matias Cattaneo, Nan Chen, Kay Giesecke, Paul Glasserman, Boris Hanin, Xuedong He, Jason Klusowski, Ciamac Moallemi, Hao Ni, Marcel Nutz, Markus Pelger, Shige Peng, Philip Protter, Ruodu Wang, Chao Zhang, Yufei Zhang, and seminar and conference participants at INFORMS 2024, ICIAM 2023, Bernoulli-IMS 11th World Congress in Probability and Statistics, the First INFORMS Conference on Financial Engineering and FinTech, the 2024 Workshop on Stochastic Control, Machine Learning and Quantitative Finance at Shanghai Jiao Tong University, the 2024 Workshop on Mathematical Finance and Insurance at Peking University, Princeton University, Stanford University, Columbia University, University of Waterloo, University of Oxford, and London School of Economics and Political Science for very helpful comments and discussion. Ruixun Zhang acknowledges research funding from the National Key R\&D Program of China (2022YFA1007900), the National Natural Science Foundation of China (72342004,12271013), the Fundamental Research Funds for the Central Universities (Peking University), and Yinhua Education Foundation.}
}

\author{
Xin Guo\thanks{Coleman Fung Chair Professor, UC Berkeley Department of Industrial Engineering and Operations Research. \url{xinguo@berkeley.edu} (email).},\ \  
Binnan Wang\thanks{Ph.D. Student, Peking University School of Mathematical Sciences and Laboratory for Mathematical Economics and Quantitative Finance. \url{wangbinnan@stu.pku.edu.cn} (email).},\ \ 
Ruixun Zhang\thanks{Assistant Professor and Boya Young Fellow, Peking University School of Mathematical Sciences, Center for Statistical Science, National Engineering Laboratory for Big Data Analysis and Applications, and Laboratory for Mathematical Economics and Quantitative Finance. \url{zhangruixun@pku.edu.cn} (email).},\ \ 
and Chaoyi Zhao\thanks{Postdoctoral Associate, MIT Sloan School of Management and Laboratory for Financial Engineering. \url{cy_zhao@mit.edu} (email).}
}

\date{March 13, 2025 %\\[.25in]
% Draft -- please do not circulate without permission of the author.
}

\maketitle
\thispagestyle{empty}

\centerline{\bf Abstract} \baselineskip 14pt \vskip 10pt
\noindent
Signatures are iterated path integrals of continuous and discrete-time processes, and their universal nonlinearity linearizes the problem of feature selection in time series data analysis. This paper studies the consistency of signature using Lasso regression, both theoretically and numerically. We establish conditions under which the Lasso regression is consistent both asymptotically and in finite sample. Furthermore, we show that the Lasso regression is more consistent with the It\^o signature for time series and processes that are closer to the Brownian motion and with weaker inter-dimensional correlations, while it is more consistent with the Stratonovich signature for mean-reverting time series and processes. We demonstrate that signature can be applied to learn nonlinear functions and option prices with high accuracy, and the performance depends on properties of the underlying process and the choice of the signature. 

\vskip 20pt\noindent {\bf Keywords}: Signature transform, Lasso, Consistency, Correlation structure, Machine learning
%
% \vskip 10pt\noindent {\bf JEL Classification}:

\newpage

\thispagestyle{empty}
\tableofcontents
\newpage

\setcounter{page}{1}
\pagenumbering{arabic}
\setlength{\baselineskip}{1.5\baselineskip}
\onehalfspacing
\setcounter{equation}{0}
\setcounter{table}{0}
\setcounter{figure}{0}

\section{Introduction}

\textbf{Background and Problem Statement.} 
Originally introduced and studied in algebraic topology \citep{Chen1954,Chen1957}, the signature transform, sometimes referred to as the path signature or signature, has been adopted and further developed in rough path theory  \citep{lyons2007differential,friz2010multidimensional}.
The signature produced from a continuous or discrete time series is a vector of real-valued features that extracts rich and relevant information \citep{morrill2020generalised,lyons2022signature}. 

{ Signature has proven to be an attractive and powerful tool for feature generation and pattern recognition with state-of-the-art performance in a wide range of domains in operations research, such as medical prediction \citep{kormilitzin2017detecting,moore2019using,morrill2019signature,morrill2020utilization,morrill2021neural,bleistein2023learning,pan2023path}, transportation \citep{guTransportationMarketRate2024}, and finance \citep{lyons2014feature,lyons2019numerical,kalsi2020optimal,salvi2021higher,akyildirim2022applications,cuchiero2023signature,futter2023signature,lemahieu2023generating}.\footnote{Other examples include handwriting recognition \citep{yang2016rotation,yang2016dropsample,wilson2018path,kidger2020neural,ibrahim2022imagesig} and action recognition \citep{yang2016deepwriterid,li2017lpsnet,fermanian2021embedding,lee2022lord,yang2022developing,cheng2023skeleton}.} 
Comprehensive reviews of successful and potential applications of signatures in machine learning can be found in \cite{chevyrev2016primer,lyons2022signature}, and \cite{morenopino2024roughtransformers}.}

Most of the empirical success and theoretical studies of the signature are built upon its striking  {\em universal nonlinearity} property: any continuous (linear or nonlinear) function of the time series can be approximated arbitrarily well by a linear combination of its signature (see \Cref{appendix: UN_detail}). 
This property linearizes the problem of feature selection, and empirical studies demonstrate that the universal nonlinearity property gives the signature several advantages over neural-network-based nonlinear methods \citep{levin2016learning,lyons2022signature,pan2023path,bleistein2023learning,guTransportationMarketRate2024}. 
First, training linear models of signature does not require the engineering of neural network architectures; 
second, the linear model allows for interpretability (we show an example in \Cref{subsec:optionpricing}).%

Despite the rapidly growing literature on the \emph{probabilistic} characteristics of signature and its successful application in machine learning, studies on the \emph{statistical} properties of the signature method are limited with a few exceptions such as \cite{kiraly2019kernels} and \cite{morrill2020generalised}.%
\footnote{\cite{kiraly2019kernels} explore the statistical properties of signature in the context of kernel learning for paths, while \cite{morrill2020generalised} summarize the statistical characteristic that signature can be regarded as an analog of the moment-generating function for time series.}
In particular, universal nonlinearity can be expressed under different definitions of signature, raising the question of which definition has better statistical properties for different processes and time series. To our knowledge, most empirical studies in the literature simply use a default definition regardless of the specific context and characteristics of the data. However, using an inappropriate signature definition may lead to suboptimal performance, as we demonstrate in this paper. 

Given the universal nonlinearity which legitimizes the regression analysis with signature, and given the popularity of Lasso regression \citep{tibshirani1996regression} to learn a sparse model of signature,%
\footnote{Examples include \cite{lyons2014rough,chevyrev2016primer,levin2016learning,moore2019using,sugiura2020machine,lemercier2021distribution,sugiura2021simple,lyons2022signature,bleistein2023learning,cuchiero2023signature}, and \cite{lemahieu2023generating}.}
the main focus of this paper is to understand the statistical properties of different forms of signature in Lasso regression given different time series data.
In particular, we study the statistical consistency in feature selection, a fundamental property for Lasso regression to achieve both explainability and good out-of-sample model performance \citep{zhao2006model,bickel2009simultaneous,wainwright2009sharp}. 

\textbf{Main Results and Contribution.} This paper studies the consistency of Lasso regression with signature both theoretically and numerically. We compare the two most widely used definitions of signature: It\^o and Stratonovich. We focus on two representative classes of Gaussian processes: multi-dimensional Brownian motion and Ornstein--Uhlenbeck (OU) process, and their respective discrete-time counterparts, i.e., random walk and autoregressive (AR) process.  These data-generating processes are simple enough to allow for analytical results while being fundamental in a number of domains ranging from machine learning \citep{song2019generative,ho2020denoising} and operations management \citep{asmussen2003applied,zhang2018degradation} to finance \citep{black1973pricing,merton1973theory} and biology \citep{martins1994estimating,hunt2007relative}.

Our contributions are multi-fold. First, we establish a probabilistic uniqueness of the universal nonlinearity { given an order of truncated signature} (Theorem \ref{thm:uniqueness}), which suggests that any feature selection procedure needs to recover this unique linear combination of signature to achieve good predictive performance.

Second, to analyze the consistency of Lasso regression with signature, we explicitly derive the correlation structure of signature for the aforementioned processes.  For Brownian motion, the correlation structure is shown to be block diagonal for the It\^o signature (\Cref{prop: structure_coef}), and to have a special odd--even alternating structure for the Stratonovich signature (\Cref{prop: structure_coef_S_integral}). In contrast,  the OU process exhibits this odd--even alternating structure for either choice of the signature (\Cref{prop: structure_coef_S_integral}).
 
Third, we establish conditions under which the Lasso regression with signature is provably consistent both asymptotically and in finite sample (Theorems \ref{prop: sufficient_irrepresentable}--\ref{th:samplestratonovich}), based on the classical notions of sign consistency and $l_\infty$ consistency \citep{zhao2006model,wainwright2009sharp}. 

Furthermore, numerical experiments show that, the Lasso regression with the It\^o signature is more consistent for time series and processes that are closer to Brownian motion and with weaker inter-dimensional correlations, while it is more consistent with the Stratonovich signature for processes with stronger mean reversion. 
In general, higher consistency rates yield better predictive performance.

Finally, we demonstrate that the signature can be applied to learn nonlinear functions and option prices with high accuracy. We compare stock options with interest rate options to highlight that performance depends on the properties of the underlying process.
This method is interpretable because the signature allows for learning a set of Arrow--Debreu state prices that are used for transfer-learning the prices of any general financial derivatives. 
These results demonstrate the practical relevance of our analysis.
 
Overall, our study takes a small step toward understanding the statistical properties of signatures for regression analysis. It fills one of the gaps between the theory and practice of signatures in machine learning. { Our findings have significant implications for various applications in operations research by guiding the selection of the appropriate signature definition to achieve better statistical properties and predictive performance. 
For example, our study provides a theoretical foundation for the signature-based adaptive-Lasso technique that has been recently developed and implemented by Amazon for transportation marketplace rate forecasting and financial planning \citep{guTransportationMarketRate2024}. This simple and novel model is reported to have generated tens of millions of monetary benefits for Amazon, and demonstrates strong potential for a wide range of applications, especially when compared to existing models such as ARIMA in terms of dealing with nonstationary  data and deep neural network approach in terms of interpretability and for limited and fragmented data. }  

\textbf{Notation.} 
Here, we define the vector and matrix norms used throughout the paper. For a vector $\mathbf{x} = (x_1,\dots,x_n)^\top \in \mathbb{R}^n$, we define $\|\mathbf{x}\|_1 = |x_1| + \cdots + |x_n|$, $\|\mathbf{x}\|_2 = \sqrt{x_1^2 + \cdots + x_n^2}$, and $\|\mathbf{x}\|_\infty = \max_{1 \leq i \leq n}|x_i|$; for a matrix $A \in \mathbb{R}^{m\times n} $, we define $\|A\|_1 = \max_{1 \leq j \leq n}\sum_{i=1}^m |a_{ij}|$, $\|A\|_2  = \sqrt{\Lambda_{\max} (A^\top A)} $, and $\|A\|_\infty = \max_{1 \leq i \leq m}\sum_{j=1}^n |a_{ij}|$, where $\Lambda_{\max}(\cdot)$ calculates the largest eigenvalue of a matrix, while $\Lambda_{\min}(\cdot)$ represents its smallest eigenvalue.

\textbf{Outline.}
The rest of this paper is organized as follows. \Cref{sec:framework} introduces the problem and key technical background. \Cref{sec:theoretical_results} presents the main theoretical results, including the uniqueness of universal nonlinearity, the correlation structure of signature, and the consistency of signature using Lasso regression. \Cref{sec:simulation} presents a simulation study to gain additional insights. \Cref{sec:application} applies our results to learning nonlinear functions and option prices. Finally, \Cref{sec:conclusion} concludes.

\section{Background}
\label{sec:framework}
In this section, we present the technical background to study the consistency of feature selection with signature transform using Lasso regression. 

\subsection{Definition of Signature Transform} \label{subsec: signatureandUNproperty}

Consider a $d$-dimensional continuous-time stochastic process $\mathbf{X}_t = (X_t^1, X_t^2,\dots, X_t^d)^\top \in \mathbb{R}^d$, $0\leq t \leq T$ on a probability space $\left(\Omega,\mathcal{F},\left\{\mathcal{F}_t\right\}_{t \geq 0},\mathbb{P}\right)$.\footnote{In this paper, we mainly consider $\mathbf{X}_t$ as a continuous-time process for the neatness of our theoretical analysis. However, our simulations and numerical applications in Sections \ref{sec:simulation} and \ref{sec:application} demonstrate that our theoretical results are also applicable to discrete time series.} Its signature or signature transform is defined as follows.

\begin{definition}[Signature]
For $k\geq 1$ and $i_1,\dots,i_k \in \{1,2,\dots,d\}$, the $k$-th order signature component of the process $\mathbf{X}$ with index $(i_1,\dots,i_k)$ from time 0 to $t$ is defined as
\begin{equation} \label{equ: def_of_signature}
    S(\mathbf{X})_{t}^{i_1,\dots,i_k} = \int_{0<t_1<\dots<t_k<t}  \mathrm{d} X_{t_1}^{i_1} \cdots \mathrm{d} X_{t_k}^{i_k},  \quad 0 \leq t \leq T.
\end{equation}
The 0-th order signature component of $\mathbf{X}$ from time 0 to $t$ is defined as $S(\mathbf{X})_{t}^{0} = 1$ for any $0 \leq t \leq T$. The signature of $\mathbf{X}$ is the collection of all the signature components of $\mathbf{X}$. The signature of $\mathbf{X}$ with orders truncated to $K$ is the collection of all the signature components of $\mathbf{X}$ with orders no more than $K$. 
\end{definition}

The $k$-th order signature component of $\mathbf{X}$ given by \eqref{equ: def_of_signature} is its $k$-fold iterated path integral along the indices $i_1, \dots, i_k$. For a given order $k$, there are $d^k$ choices of indices $(i_1,\dots,i_k)$, and therefore the number of all $k$-th order signature components is $d^k$. 

The integral in \eqref{equ: def_of_signature} can be specified using different definitions. For example, if $\mathbf{X}$ is a deterministic process, it can be defined via the Riemann/Lebesgue integral. If $\mathbf{X}$ is a multi-dimensional Brownian motion, it is a stochastic integral defined by either the It\^o integral or the Stratonovich integral. Throughout the paper, for clarity, we write 
\begin{equation*}
    S(\mathbf{X})_{t}^{i_1,\dots,i_k, I} = \int_{0<t_1<\dots<t_k<t}  \mathrm{d} X_{t_1}^{i_1} \cdots \mathrm{d} X_{t_k}^{i_k} = \int_{0<s<t} S(\mathbf{X})_{s}^{i_1,\dots,i_{k-1}, I} \mathrm{d} X_{s}^{i_k}
\end{equation*}
when using the It\^o integral, and 
\begin{equation*}
    S(\mathbf{X})_{t}^{i_1,\dots,i_k, S} = \int_{0<t_1<\dots<t_k<t}  \mathrm{d} X_{t_1}^{i_1} \circ \cdots \circ\mathrm{d} X_{t_k}^{i_k} = \int_{0<s<t} S(\mathbf{X})_{s}^{i_1,\dots,i_{k-1}, S} \circ \mathrm{d} X_{s}^{i_k}
\end{equation*}
when using the Stratonovich integral. For ease of exposition, we refer to the signature of $\mathbf{X}$ as the It\^o (resp. Stratonovich) signature if the integral is defined in the sense of the It\^o (resp. Stratonovich) integral.

\subsection{Universal Nonlinearity of Signature}\label{appendix: UN_detail}

One of the remarkable properties of the signature is its universal nonlinearity \citep{levin2016learning,kiraly2019kernels,fermanian2021embedding,lemercier2021distribution,lyons2022signature}.%
\footnote{Signature also enjoys several other nice probabilistic properties under mild conditions. First, all expected signature components of a stochastic process characterize the distribution of the process \citep{chevyrev2016characteristic,chevyrev2022signature}. Second, the signature of a process uniquely determines the path of the underlying process up to a tree-like equivalence \citep{hambly2010uniqueness,le2013stratonovich,boedihardjo2014uniqueness}.
}
It is particularly relevant for feature selection in statistical and machine learning, where one needs to find or learn a (nonlinear) function $f$ that maps the path of $\mathbf{X}$ to a target label $y$. Examples include learning diagnosis or signals from medical time series such as the electrocardiogram \citep{morrill2019signature,morrill2020utilization,morrill2021neural}, forecasting transportation marketplace rates from the time series of supply, demand, and macroeconomic factors \citep{guTransportationMarketRate2024}, and learning a nonlinear payoff or pricing function for financial derivatives given the time series of the underlying asset prices \citep{hutchinson1994nonparametric,bertsimas2001hedging,lyons2020non}.

The following theorem of \cite{cuchiero2023signature} outlines the universal nonlinearity.

\begin{theorem}[Universal nonlinearity, {\citet[Theorem 2.12]{cuchiero2023signature}}]\label{th:UN}
    Let $\mathbf{X}_t$ be a continuous $\mathbb{R}^d$-valued semimartingale and $\mathcal{S}$ be a compact subset of paths of the time-augmented process $\tilde{\mathbf{X}}_t = \begin{pmatrix}
      t , {\mathbf{X}}_t^\top 
    \end{pmatrix}^\top$ from time 0 to $T$.%
    \footnote{See Appendix \ref{appendix:timeaug} for details of the time augmentation.}
    Assume that $f: \mathcal{S} \to \mathbb{R}$ is a real-valued continuous function. Then, for any $\varepsilon>0$, there exists a linear functional $L: \mathbb{R}^\infty \to \mathbb{R}$ such that
    \begin{equation*}
        \sup_{s \in \mathcal{S}}\left| f (s) - L(\mathrm{Sig}(s)) \right| < \varepsilon,
    \end{equation*}
    where $\mathrm{Sig}(s)$ is the signature of $s$. 
\end{theorem}

By universal nonlinearity, any continuous function $f$ can be approximated arbitrarily well by a linear combination of the signature of $\mathbf{X}$. This lays the foundation for learning the relationship between the time series $\mathbf{X}$ and a target label $y$ using a linear regression.

\subsection{Feature Selection with Signature Using Lasso Regression} \label{subsec: selectingsignaturesLasso}

Consider $N$ pairs of samples, $( \mathbf{X}_{1}, y_1),  ( \mathbf{X}_{2}, y_2), \dots,  ( \mathbf{X}_{N}, y_N)$, where $\mathbf{X}_{n} = \{\mathbf{X}_{n,t}\}_{0\leq t \leq T}$ is the $n$-th path realization of $\mathbf{X}_t$ for $n=1,2,\dots,N$. Given a fixed order $K\geq 1$,  assume that $(\mathbf{X}_{n}, y_n)$ satisfies the following regression model
\begin{equation} \label{equ: linearregression}
    y_n = \beta_0 + \sum_{i_1 =1}^d \beta_{i_1} S(\mathbf{X}_{n})^{i_1}_T +
    \sum_{i_1,i_2 =1}^d \beta_{i_1,i_2} S(\mathbf{X}_{n})^{i_1,i_2}_T + \dots +
    \sum_{i_1,\dots,i_K =1}^d \beta_{i_1,\dots,i_K} S(\mathbf{X}_{n})^{i_1,\dots,i_K}_T +\varepsilon_n,
\end{equation}
where $\{\varepsilon_n\}_{n=1}^N$ are independent and identically distributed errors following a normal distribution with zero mean and finite variance. Here the number of predictors, i.e., the signature components of various orders, is $\frac{d^{K+1}-1}{d-1}$, including the 0-th order signature component $S(\mathbf{X})^{0}_T=1$, whose coefficient is $\beta_0$.

Recall that the goal of Lasso regression is to identify a sparse set of true predictors/features among all the predictors included in linear regression \eqref{equ: linearregression}. A predictor has a zero beta coefficient if it is not in the true model. 
We use $A^*_k$ to represent the set of all signature components of order $k$ with nonzero coefficients in \eqref{equ: linearregression}, and define the set of true predictors $A^*$ by
\begin{equation} \label{equ: activeset}
    A^* = \bigcup_{k=0}^K A_k^* := \bigcup_{k=0}^K \{ (i_1,\dots,i_k): \beta_{i_1,\dots,i_k} \neq 0  \}.
\end{equation}
Here, we begin the union with $k=0$ to include the 0-th order signature for notational convenience.

Given a tuning parameter $\lambda>0$ and $N$ samples, the Lasso estimator identifies the true predictors using
\begin{align}
    \hat{{\boldsymbol \beta}}^N(\lambda) = \arg\min_{\hat{\boldsymbol \beta}}\Bigg[  \sum_{n=1}^N \Bigg( y_n -& \hat{\beta}_0 - \sum_{i_1 =1}^d \hat{\beta}_{i_1} \tilde{S}(\mathbf{X}_{n})^{i_1}_T -
    \sum_{i_1,i_2 =1}^d \hat{\beta}_{i_1,i_2} \tilde{S}(\mathbf{X}_{n})^{i_1,i_2}_T - \cdots \nonumber\\
    & -
    \sum_{i_1,\dots,i_K =1}^d \hat{\beta}_{i_1,\dots,i_K} \tilde{S}(\mathbf{X}_{n})^{i_1,\dots,i_K}_T \Bigg)^2 + \lambda \left\|  \hat{\boldsymbol \beta} \right\|_1   \Bigg], \label{equ: lasso}
\end{align}
where $\hat{\boldsymbol \beta}$ is the vector containing all coefficients $\hat{\beta}_{i_1,\dots,i_k}$. Here, $\tilde{S}(\mathbf{X}_{n})$ represents the standardized version of ${S}(\mathbf{X}_{n})$  across $N$ samples by the $l_2$-norm. That is, for any index $(i_1,\dots,i_k)$,
\begin{equation*} \label{equ: standardization}
    \tilde{S}(\mathbf{X}_{n})^{i_1,\dots,i_k}_T = \frac{ {S}(\mathbf{X}_{n})^{i_1,\dots,i_k}_T }{  \sqrt{ \sum_{m=1}^N \left[{S}(\mathbf{X}_{m})^{i_1,\dots,i_k}_T \right]^2 \Big/ N } }, \quad n=1,2,\dots,N.%
\footnote{We perform this standardization for two reasons. First, the Lasso estimator is sensitive to the magnitudes of the predictors \citep{hastie2009elements} and the magnitudes of different orders of signature components are different \citep{lyons2007differential}, therefore standardization is necessary to ensure that the coefficients of different orders of signature are on the same scale and can be compared directly.
Second, the sample covariance matrix is now equivalent to the sample correlation matrix, allowing us to focus on the correlation structure of the signature components in the subsequent analysis.}
\end{equation*}

{ The Lasso estimator depends on the choice of $K$. The universal nonlinearity demonstrates that the linear combination of \emph{all} components of the signature of $\mathbf{X}$ can be used to approximate $f$. However, due to computational constraints, we must truncate the signature to a finite order $K$ in the implementation of Lasso regression. 
% The choice of $K$ involves a tradeoff between model complexity and performance.
In theory, one can exploit the signature approximation in \cite{dupire2022functional} and the recent results on Taylor expansions of signatures in \cite{Christa2024} to develop an error-bound analysis for the choice of $K$. \cite{fermanian2022functional} also provides a practical approach to choose $K$ based on the tradeoff between the approximation error and the number of coefficients in the regression model. In practice, it has also been documented that a small order $K$ usually suffices to achieve satisfactory performances \citep{morrill2020generalised,lyons2022signature,guTransportationMarketRate2024}. For example, \cite{guTransportationMarketRate2024} show that $K=3$ is sufficient for forecasting models of transportation rates in Amazon.
}

\subsection{Consistency and the Irrepresentable Condition of Lasso Regression} \label{subsec: consistency}

Our goal is to study the consistency of feature selection with signature using the Lasso estimator in \eqref{equ: lasso}. Broadly speaking, consistency means that the Lasso estimator converges to the true coefficients as the number of samples increases. In this section, we introduce two widely used notions of Lasso consistency from the literature 
and discuss the corresponding conditions required for each notion of consistency.

\cite{zhao2006model} propose the sign consistency for Lasso regression, which requires that the signs of all components of the Lasso estimator match those of the true coefficients as the number of samples increases without bound.  
\cite{wainwright2009sharp} studies the consistency of Lasso regression by requiring that the $l_\infty$ distance between the true and the estimated coefficients is bounded.

In the context of Lasso regression with signature, the sign consistency and the $l_\infty$ consistency of Lasso are defined as follows. 
\begin{definition}[Sign consistency] \label{def: signconsistency}
Lasso regression is (strongly) sign consistent if there exists $\lambda_N$, a function of sample number $N$, such that
\begin{equation*}
    \lim_{N\to+\infty} \mathbb{P} \left( \mathrm{sign} \left( \hat{{\boldsymbol \beta}}^N(\lambda_N) \right) = \mathrm{sign} ({\boldsymbol \beta} )  \right) = 1,
\end{equation*}
where $\hat{{\boldsymbol \beta}}^N(\cdot)$ is the Lasso estimator given by \eqref{equ: lasso}, ${\boldsymbol \beta}$ is a vector containing all beta coefficients of the true model \eqref{equ: linearregression}, and the function $\mathrm{sign}(\cdot)$ maps positive entries to $1$, negative entries to $-1$, and $0$ to $0$.
\end{definition}

\begin{definition}[$l_\infty$ consistency] \label{def: norm bound}
There exists a function of $\lambda_N$, $g(\lambda_N)$, such that the Lasso regression 
satisfies the $l_\infty$ bound
\begin{equation*}
    \left\|  \hat{{\boldsymbol \beta}}^N(\lambda_N) - \tilde{\boldsymbol \beta}  \right\|_\infty \leq g(\lambda_N),
\end{equation*}
where $\hat{{\boldsymbol \beta}}^N(\cdot)$ is the Lasso estimator given by \eqref{equ: lasso} and $\tilde{\boldsymbol \beta}$ is a vector containing all standardized beta coefficients of the true model whose component with index $(i_1,\dots,i_k)$ is given by
\begin{equation*}
    \tilde{\beta}_{i_1,\dots,i_k} = \beta_{i_1,\dots,i_k} \cdot \sqrt{ \frac{1}{N}\sum_{m=1}^N \left[{S}(\mathbf{X}_{m})^{i_1,\dots,i_k}_T \right]^2 } .
\end{equation*}
\end{definition}

As discussed in \cite{wainwright2009sharp}, if the support of $\hat{{\boldsymbol \beta}}^N(\lambda_N)$ is contained within the support of ${\boldsymbol \beta}$ and the absolute values of all beta coefficients for predictors in $A^*$ are greater than $g(\lambda_N)$, the $l_\infty$ consistency implies the sign consistency.

To guarantee the consistency of Lasso, \citet{zhao2006model} and \cite{wainwright2009sharp} propose the following two irrepresentable conditions, respectively. 

\begin{definition}[Irrepresentable condition]\label{def: irrepresentable_model}
The feature selection in \eqref{equ: linearregression} satisfies irrepresentable condition I if there exists a constant $\gamma \in (0,1]$ such that
\begin{equation*} % \label{equ: Corr_Ito_BM}
    \text{I.}\quad\quad
     \left\| {\Delta}_{A^{*c},A^*} {\Delta}_{A^*,A^*}^{-1} \mathrm{sign}( {\boldsymbol \beta}_{A^*} ) \right\|_\infty \leq 1-\gamma, %\mathbf{1} - {\boldsymbol \eta} ,
\end{equation*}
and satisfies irrepresentable condition II if there exists a constant $\gamma \in (0,1]$ such that
\begin{equation*} 
    \text{II.}\quad\quad
     \left\| {{\Delta}}_{A^{*c},A^*} {{\Delta}}_{A^*,A^*}^{-1} \right\|_\infty \leq 1-\gamma ,
\end{equation*}
where $A^*$ is given by \eqref{equ: activeset},  $A^{*c}$ is the complement of $A^*$, ${\Delta}_{A^{*c},A^*}$ (${\Delta}_{A^*,A^*}$) represents the correlation matrix\footnote{In this paper, in line with \cite{zhao2006model}, all covariances and correlation coefficients are defined to be uncentered. Specifically, for random variables $X$ and $Y$, we define their covariance as $\mathbb{E}[XY]$, and their correlation coefficient as $\mathbb{E}[XY]/\sqrt{\mathbb{E}[X^2]\mathbb{E}[Y^2]}$. One can easily extend our results to the centered case.} between all predictors in $A^{*c}$ and $A^*$ ($A^*$ and $A^*$), and ${\boldsymbol \beta}_{A^*}$ represents a vector formed by beta coefficients for all predictors in $A^*$. 
\end{definition}
The irrepresentable conditions in Definition \ref{def: irrepresentable_model} intuitively mean that irrelevant predictors in $A^{*c}$ cannot be adequately represented by the true predictors in $A^*$, implying weak collinearity between the predictors. \citet{zhao2006model} demonstrate that the irrepresentable condition I is almost a necessary and sufficient condition for the Lasso regression to be sign consistent. \cite{wainwright2009sharp} proves that the irrepresentable condition II is a sufficient condition for the $l_\infty$ consistency of Lasso regression under specific technical assumptions. The irrepresentable condition II is slightly stronger than the irrepresentable condition I.

In the context of signature, predictors in the linear regression \eqref{equ: linearregression} are correlated and have special correlation structures that differ from previous studies on Lasso \citep{zhao2006model,bickel2009simultaneous,wainwright2009sharp}. 
We show in the following section that in fact their correlation structures vary with the underlying process $\mathbf{X}$ and the choice of integrals in the definition of signature \eqref{equ: def_of_signature}. These different correlation structures lead to different statistical consistencies.% for different processes

\section{Theoretical Results}
\label{sec:theoretical_results}
This section presents the main theoretical results. \Cref{subsec:uniqueness} shows the \emph{uniqueness} of universal nonlinearity in a probabilistic sense. 
Section \ref{sec: correlationstructure} characterizes the correlation structures between signature components. 
Section \ref{sec: consistency} presents the results of consistency in signature selection, both asymptotically ($N=\infty$) and for a finite sample ($N<\infty$).

As outlined in \Cref{fig:outline} for our results, the statistical consistency of signature using Lasso regression depends on two factors---the underlying processes $\mathbf{X}$ (Brownian motion or OU process) and the definition of signature (It\^o or Stratonovich).

\centerline{[Insert \Cref{fig:outline} approximately here.]}

\subsection{Uniqueness of Universal Nonlinearity}\label{subsec:uniqueness}
The universal nonlinearity in \Cref{th:UN} shows the \textit{existence} of a linear combination of signature components to approximate any function $f$. We provide the following Theorem \ref{thm:uniqueness} to complement the universal nonlinearity, which demonstrates the \textit{uniqueness} of this linear combination in a probabilistic sense { given an order of truncated signature}. To the best of our knowledge, Theorem \ref{thm:uniqueness} has not appeared in the literature.

\begin{theorem}[Uniqueness] \label{thm:uniqueness}
Given $K \geq 1$, let $S=(S_1,S_2,\dots,S_p)^\top$ be the vector of the signature of a stochastic process $\mathbf{X}$ with orders truncated to $K$, and assume $S$ has a non-degenerate joint distribution. Consider two different linear combinations of signature components, $L_a = \sum_{i=1}^p a_iS_i$ and $L_b = \sum_{i=1}^p b_iS_i$, such that $a_i\neq b_i$ for at least some $i$. Then, there exists a constant $\theta>0$ such that, for any $\eta \in (0,\overline{\eta})$,
\begin{equation}\label{equ:unique1}
    \mathbb{P} \left( \left|L_a - L_b \right|>\eta \right)\geq P_\theta^*(\eta) > 0.
    \end{equation}
Furthermore, if $f$ is a function that maps $\mathbf{X}$ to a real value such that $\left|f(\mathbf{X})-L_a \right|\leq \varepsilon$ almost surely for a constant $\varepsilon < \overline{\eta}$, then for any $\eta \in (0,\overline{\eta}-\varepsilon)$,
\begin{equation}\label{equ:unique}
    \mathbb{P}\left( \left|f(\mathbf{X})-L_b \right| > \eta \right)\geq  P_\theta^*(\eta+\varepsilon) > 0.
\end{equation}
Here
\begin{align*}
&P_\theta^*(\eta) = \left(1-\frac{1}{\theta} \right) \cdot \frac{  \frac{\mathbb{E} \left[ {\left(\sum_{i=1}^p c_iS_i\right)}^2 \Bigl| \|S\|_2 \leq \theta\sqrt{p\|\Sigma\|_2} \right]}{\theta \|C\|_\infty p \sqrt{\|\Sigma\|_2}} - \eta}{\theta \|C\|_\infty p \sqrt{\|\Sigma\|_2} - \eta} ,
\\&\overline{\eta} = \min\left\{ \frac{\mathbb{E} \left[ {\left(\sum_{i=1}^p c_iS_i\right)}^2 \Bigl| \|S\|_2 \leq \theta\sqrt{p\|\Sigma\|_2} \right]}{\theta \|C\|_\infty p \sqrt{\|\Sigma\|_2}} ,\theta \|C\|_\infty p \sqrt{\|\Sigma\|_2}\right\},
\end{align*}
with $c_i=a_i-b_i$, $C = (c_1,c_2,\dots,c_p)^\top$, 
and $\Sigma=\mathbb{E}(SS^\top)$.

    \label{th:1}
\end{theorem}

Theorem \ref{thm:uniqueness} has important implications for selecting signature components using Lasso regression. In particular, \eqref{equ:unique} shows that when a nonlinear function $f$ is approximated by a linear combination of signature components $L_a$, there is always a positive probability that a different linear combination $L_b$ has a positive gap from $f$, which implies that $L_a$ is the unique linear combination to approximate $f$ { given an order of truncated signature $K$.\footnote{
This uniqueness is only for a fixed value of $K$. In fact, if $L_a$ and $L_b$ are linear combinations of signature components with orders truncated to $K$ and $K+1$, respectively, the true function $f$ could always be better approximated by $L_b$.}}
Therefore, given $f$, it is important for any feature selection procedure to recover this unique linear combination of signature components to achieve statistical consistency in feature selection.

{ There is a strand of literature focusing on whether signatures can uniquely determine the path of the underlying process; see \cite{hambly2010uniqueness}, \cite{le2013stratonovich}, and \cite{boedihardjo2014uniqueness}. This literature investigates the one-to-one correspondence between $\mathbf{X}$ and its signature. This is different from Theorem \ref{thm:uniqueness}, which characterizes the one-to-one correspondence between $f(\mathbf{X})$ and the linear combination of signature components. 
}

\subsection{Correlation Structure of Signature} \label{sec: correlationstructure}

\begin{figure}[htbp]
    \centering
    \caption{Outline of main theoretical results.}
    \label{fig:outline}
    \subfigure{
        \includegraphics[width=0.95\linewidth]{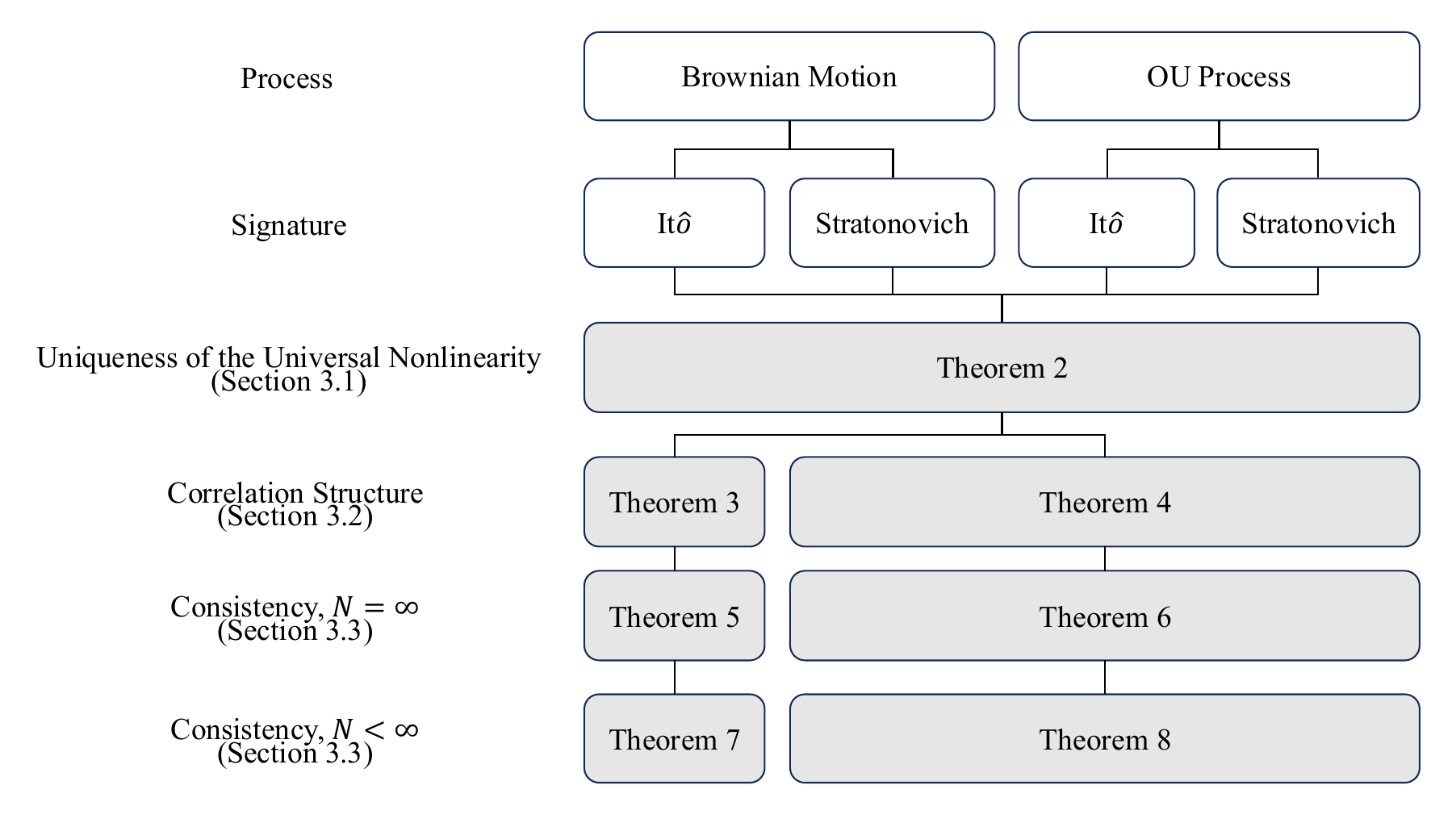}
    }
   
\end{figure}

Now we study the correlation structure of the four combinations of processes and signatures in \Cref{fig:outline}. 
Throughout the paper, we define $\mathbf{X}=\{\mathbf{X}_t\}_{t\ge 0}$ as a $d$-dimensional Brownian motion on a probability space $\left(\Omega,\mathcal{F},\left\{\mathcal{F}_t\right\}_{t \geq 0},\mathbb{P}\right)$  if
    \begin{equation} \label{equ: MultiBM_setup}
        \mathbf{X}_t = (X_t^1, X_t^2,\dots, X_t^d)^\top = \Gamma (W_t^1, W_t^2,\dots, W_t^d)^\top,
    \end{equation}
    where $W_t^1, W_t^2, \dots, W_t^d$ are mutually independent 1-dimensional standard Brownian motions on $\mathbb{R}$, and $\Gamma$ is a matrix independent of $t$. In particular,  $\langle X_t^i,X_t^j\rangle = \rho_{ij} \sigma_i \sigma_j t$ with $\rho_{ij} \sigma_i \sigma_j = (\Gamma \Gamma^\top )_{ij}$, where $\sigma_i^2t$ is the variance of $X_t^i$ and $\rho_{ij} \in [-1,1]$ is the inter-dimensional correlation between $X_t^i$ and $X_t^j$.

We say that $\mathbf{X}=\{\mathbf{X}_t\}_{t\ge 0}$ is a $d$-dimensional OU process on a probability space $\left(\Omega,\mathcal{F},\left\{\mathcal{F}_t\right\}_{t \geq 0},\mathbb{P}\right)$ if
    \begin{equation} \label{equ: OU_setup}
        \mathbf{X}_t = (X_t^1, X_t^2,\dots, X_t^d)^\top = \Gamma (Y_t^1, Y_t^2,\dots, Y_t^d)^\top,
    \end{equation}
where $\Gamma$ is a $d \times d$ matrix independent of $t$, and $Y_t^1, Y_t^2, \dots, Y_t^d$ are mutually independent 1-dimensional OU processes on $\mathbb{R}$  driven by stochastic differential equations
    \begin{equation*} \label{equ: OU_process}
        \mathrm{d} Y_t^i = - \kappa_i Y_t^i \mathrm{d} t + \mathrm{d} W_t^i, \quad Y_0^i = 0,
    \end{equation*}
for $i=1,2,\ldots,d$. Here $\kappa_i>0$ are parameters to control the speed of mean reversion and a higher $\kappa_i$ implies a stronger mean reversion. When $\kappa_i = 0$, $Y_t^i$ reduces to a standard Brownian motion.

\textbf{It\^o Signature of Brownian Motion.}
The following proposition gives the moments of the It\^o signature of a $d$-dimensional Brownian motion. 

\begin{proposition} \label{prop: moment_signatures}
Let $\mathbf{X}$ be a $d$-dimensional Brownian motion given by \eqref{equ: MultiBM_setup}. For $m, n \in \mathbb{Z}^+$ and $m\neq n$,
\begin{align*}
&\mathbb{E} \left[S(\mathbf{X})_{t}^{i_1,\dots,i_n,I} \right] = 0 ,\quad \mathbb{E} \left[S(\mathbf{X})_{t}^{i_1,\dots,i_n,I} S(\mathbf{X})_{t}^{j_1,\dots,j_n,I} \right] = \frac{t^n}{n!} \prod_{k=1}^n \rho_{i_k j_k} \sigma_{i_k} \sigma_{j_k},\\
&\mathbb{E} \left[S(\mathbf{X})_{t}^{i_1,\dots,i_n,I} S(\mathbf{X})_{t}^{j_1,\dots,j_m,I} \right] = 0 .\label{prop: moment_signatures: equ2}
\end{align*}
\end{proposition}
With Proposition \ref{prop: moment_signatures}, the following result explicitly characterizes the correlation structure of the It\^o signature for Brownian motion. 

\begin{theorem} \label{prop: structure_coef}
Let $\mathbf{X}$ be a $d$-dimensional Brownian motion given by \eqref{equ: MultiBM_setup}. If the signature is rearranged in recursive order (see Definition \ref{def: recursive order} in Appendix \ref{appendix: calculation_of_corr_BM}), then the correlation matrix for the It\^o signature of $\mathbf{X}$ with orders truncated to $K$ is a block diagonal matrix given by
\begin{equation}
\label{equ: Corr_Ito_BM}
    \Delta^1 = \mathrm{diag}\{ \Omega_0, \Omega_1, \Omega_2,\dots,\Omega_K \},
\end{equation}
where each diagonal block $\Omega_k$ represents the correlation matrix for all $k$-th order signature components given by
\begin{equation} \label{equ: Omega_k}
\Omega_k = \underbrace{ \Omega \otimes \Omega \otimes \cdots \otimes \Omega }_{k}, \quad k =1,2,\dots, K,
\end{equation}
and $\Omega_0 = 1$. Here $\otimes$ represents the Kronecker product and $\Omega$ is a $d\times d$ matrix with $\rho_{ij}$ being the $(i,j)$-th entry. 
\end{theorem}
Theorem \ref{prop: structure_coef} shows that the It\^o signature components of different orders are mutually uncorrelated, leading to a block diagonal correlation structure; the correlation between signature components of the same order has a Kronecker product structure determined by the correlation $\rho_{ij}$ of the Brownian motion.

{ The block diagonal structure of the correlation matrix has important statistical implications for the It\^o signature. In Section \ref{sec: consistency}, we demonstrate that the Lasso regression using signature as predictors is consistent if the correlation is weak (see Theorems \ref{prop: sufficient_irrepresentable} and \ref{th:sampleito}). However, when the correlation within each block is strong, Lasso may be unstable for signature components in the same block. In such cases, one may consider using methods such as sparse principal component analysis \citep{Zou2006,Leng2009} or scaled Lasso \citep{Arashi2021} to address multicollinearity and achieve a more stable Lasso estimation.
}

\textbf{Stratonovich Signature of Brownian Motion and Both Signatures of OU Process.}
We first provide the moments of the Stratonovich signature of Brownian motion. 

\begin{proposition} \label{prop: moment_signatures_S_integral}
Let $\mathbf{X}$ be a $d$-dimensional Brownian motion given by \eqref{equ: MultiBM_setup}. For $m, n\in \mathbb{Z}^+$, we have
\begin{align*}
&\mathbb{E} \left[S(\mathbf{X})_{t}^{i_1,\dots,i_{2n-1},S}\right] = 0 , \quad \mathbb{E} \left[S(\mathbf{X})_{t}^{i_1,\dots,i_{2n},S}\right] = \frac{1}{2^n} \frac{t^{n}}{n!} \prod_{k=1}^{n} \rho_{i_{2k-1}i_{2k}} \prod_{k=1}^{2n} \sigma_{i_k} ,\\
&\mathbb{E} \left[S(\mathbf{X})_{t}^{i_1,\dots,i_{2n},S} S(\mathbf{X})_{t}^{j_1,\dots,j_{2m-1},S}\right] = 0 , \label{equ: S_integral_odd_even_zero}
\end{align*}
and $\mathbb{E} \left[S(\mathbf{X})_{t}^{i_1,\dots,i_{2n},S} S(\mathbf{X})_{t}^{j_1,\dots,j_{2m},S}\right]$ and $\mathbb{E} \left[S(\mathbf{X})_{t}^{i_1,\dots,i_{2n-1},S} S(\mathbf{X})_{t}^{j_1,\dots,j_{2m-1},S}\right]$ can be calculated using formulas provided in Proposition \ref{prop: moment_signatures_S_integral_TEMP} in Appendix \ref{appendix: calculation_of_corr_BM}.
\end{proposition}

The calculation of moments for the OU process is more complicated than those for the Brownian motion, as discussed in Appendix \ref{appendix: calculation_of_corr_OU}. Nonetheless, the correlation matrices of both the It\^o and the Stratonovich signatures of the OU process exhibit the same odd--even alternating structure as that of the Stratonovich signature of the Brownian motion, which is given below.

\begin{theorem} \label{prop: structure_coef_S_integral}
Consider the Stratonovich signature of a $d$-dimensional Brownian motion given by \eqref{equ: MultiBM_setup}, or the It\^o or the Stratonovich signature of a $d$-dimensional OU process given by \eqref{equ: OU_setup}. The correlation matrix for the signature with orders truncated to $2K$ has an odd--even alternating structure given by
\begin{equation} \label{equ: BM_S_ALTERNATING}
    \Delta^2 = \begin{pmatrix}
    \Psi_{0,0} & 0 & \Psi_{0,2} & 0 & \cdots & 0 & \Psi_{0,2K} \\
    0 & \Psi_{1,1} & 0 & \Psi_{1,3} & \cdots & \Psi_{1,2K-1} & 0 \\
    \Psi_{2,0} & 0 & \Psi_{2,2} & 0 & \cdots & 0 & \Psi_{2,2K} \\
    0 & \Psi_{3,1} & 0 & \Psi_{3,3} & \cdots & \Psi_{3,2K-1} & 0 \\
    \vdots & \vdots & \vdots & \vdots & \ddots & \vdots & \vdots \\
    0 & \Psi_{2K-1,1} & 0 & \Psi_{2K-1,3} & \cdots & \Psi_{2K-1,2K-1} & 0 \\
    \Psi_{2K,0} & 0 & \Psi_{2K,2} & 0 & \cdots & 0 & \Psi_{2K,2K} 
    \end{pmatrix},
\end{equation}
where $\Psi_{m,n}$ is the correlation matrix between all $m$-th and $n$-th order signature components.\footnote{For the Stratonovich signature of a $d$-dimensional Brownian motion, $\Psi_{m,n}$ is given by Proposition \ref{prop: moment_signatures_S_integral_TEMP} in Appendix \ref{appendix: calculation_of_corr}.} 
In particular, if the indices of the signature components are rearranged with all odd-order signature components and all even-order signature components together respectively, the correlation matrix has a block diagonal form given by
\begin{equation} \label{equ:odd_even_struct}
   \tilde{\Delta}^2 = \mathrm{diag} \{ \Psi_{\mathrm{odd}}, \Psi_{\mathrm{even}} \},
\end{equation}
where 
\begin{equation} \label{equ: def_odd_even_matrix}
   \Psi_{\mathrm{odd}}=\begin{pmatrix}
    \Psi_{1,1} & \Psi_{1,3} & \cdots & \Psi_{1,2K-1}  \\
    \Psi_{3,1} & \Psi_{3,3} & \cdots & \Psi_{3,2K-1}\\
    \vdots & \vdots & \cdots & \vdots  \\
    \Psi_{2K-1,1} & \Psi_{2K-1,3} & \cdots & \Psi_{2K-1,2K-1}
    \end{pmatrix}, \quad\Psi_{\mathrm{even}}=\begin{pmatrix}
    \Psi_{0,0} & \Psi_{0,2} & \cdots & \Psi_{0,2K}  \\
    \Psi_{2,0} & \Psi_{2,2} & \cdots & \Psi_{2,2K}  \\
    \vdots & \vdots & \cdots & \vdots  \\
    \Psi_{2K,0} & \Psi_{2K,2} & \cdots & \Psi_{2K,2K} 
    \end{pmatrix}. 
\end{equation}
\end{theorem}

Theorems \ref{prop: structure_coef} and \ref{prop: structure_coef_S_integral} reveal a striking difference between the four combinations of processes and signatures in \Cref{fig:outline}. Specifically, a Brownian motion's It\^o signature components of different orders are uncorrelated, leading to a block diagonal correlation structure. In contrast, for the Stratonovich signature of Brownian motion and both signatures of the OU process, the components are uncorrelated only if they have different parity, leading to an odd--even alternating structure. This difference has significant implications for the consistency of the four combinations of processes and signatures, as will be discussed in Section \ref{sec: consistency}.

Finally, signature-based analyses sometimes consider time augmentation in which a time dimension $t$ is added to the original process $\mathbf{X}_t$ \citep{chevyrev2016primer,lyons2022signature}.
Appendix \ref{appendix:timeaug} provides the correlation structure and consistency results of the time-augmented processes.

\subsection{Consistency of Signature Using Lasso Regression} \label{sec: consistency}
This section investigates the consistency of feature selection using the four combinations of processes and signatures in \Cref{fig:outline}. 

\textbf{Asymptotic Results.}
The following theorem characterizes the conditions under which the irrepresentable condition holds for the It\^o signature of Brownian motion.  

\begin{theorem} \label{prop: sufficient_irrepresentable}
For a multi-dimensional Brownian motion given by \eqref{equ: MultiBM_setup}, consider its It\^o signature with orders truncated to $K$. Both irrepresentable conditions I and II hold if and only if they hold for each $\Omega_k$ in \eqref{equ: Omega_k}. In addition, both irrepresentable conditions I and II hold if
\begin{equation}
\label{equ:bound_irc}
    |\rho_{ij}| < \frac{1}{2q_{\max}-1},
\end{equation}
where $q_{\max}=\max_{0 \leq k \leq K} \{  \#A^*_k\}$ and $A^*_k$ is the set of true predictors of order $k$ defined in \eqref{equ: activeset}.
\end{theorem}

The sufficient condition \eqref{equ:bound_irc} in Theorem \ref{prop: sufficient_irrepresentable} requires that different dimensions of the multi-dimensional Brownian motion are not strongly correlated, with a sufficient bound given by \eqref{equ:bound_irc}. Empirically, it has been documented that a small $K$ suffices to provide a reasonable approximation in applications \citep{morrill2020generalised,lyons2022signature}. Therefore, $q_{\max}$ is typically small, which implies that the bound given by \eqref{equ:bound_irc} is fairly easy to satisfy.
In addition, Appendix \ref{appendix:tight_sufficient} discusses the tightness of this bound.

{ 
In fact, \citet[Corollary 2]{zhao2006model} demonstrate that any Lasso regression is consistent if the absolute values of the correlations between predictors are smaller than $1/(2q-1)$, where $q=\# A^*$ is the total number of true predictors in the Lasso regression. Our sufficient condition \eqref{equ:bound_irc} provides a much more relaxed upper bound compared to \possessivecite{zhao2006model} condition, thanks to the block diagonal correlation structure of the It\^o signature for Brownian motion given by Theorem \ref{prop: structure_coef}. This is because $A_k^*$ is the set of true predictors in the $k$-th block of \eqref{equ: Corr_Ito_BM}, and $q_{\max}$ is the maximum number of true predictors across all these blocks. In other words, even with a large number of all true predictors ($\# A^*$) in the Lasso regression, it remains consistent as long as the number of true predictors within each block ($\# A^*_k$) is relatively small.
} 

The following theorem characterizes the condition under which the irrepresentable condition holds for the Stratonovich signature of a Brownian motion and for both signatures of the OU process. 

\begin{theorem} \label{prop: irrepresentable_equiv}
Consider the Stratonovich signature of a $d$-dimensional Brownian motion given by \eqref{equ: MultiBM_setup}, or the It\^o or the Stratonovich signature of a $d$-dimensional OU process given by \eqref{equ: OU_setup}, with orders truncated to $2K$. Both irrepresentable conditions I and II hold if and only if they hold for both $\Psi_{\mathrm{odd}}$ and $\Psi_{\mathrm{even}}$ in \eqref{equ: def_odd_even_matrix}.
\end{theorem}

For these types of signatures, the irrepresentable condition may fail even when all dimensions of $\mathbf{X}$ are mutually independent, as is shown in Example \ref{exmp: fig: corr_S_rho_0} in Appendix \ref{appendix: calculation_of_corr}. Therefore, no sufficient conditions of the form \eqref{equ:bound_irc} can be established.
This implies that, for example, the Stratonovich signature of Brownian motion may exhibit lower consistency compared to its It\^o signature, which we confirm in \Cref{sec:simulation}.

\textbf{Finite Sample Results.}
Theorems \ref{prop: sufficient_irrepresentable} and \ref{prop: irrepresentable_equiv} characterize when the irrepresentable conditions hold for the population correlation matrix of signature, which implies the sign consistency (Definition \ref{def: signconsistency}) of Lasso regression when $N=\infty$. 
In practice, however, the number of sample paths is finite, i.e., $N < \infty$. Hence, the \emph{sample correlation matrix}, denoted by $\hat{\Delta}$, may deviate from the population correlation matrix $\Delta$. 
The following results demonstrate that the Lasso regression using signature maintains consistency \emph{with high probability} in finite sample under certain conditions.\footnote{This statement aligns with the convention of the literature of high-dimensional statistics; see, for example, \cite{wainwright2009sharp}, \cite{ravikumar2011high}, and \cite{vershynin2018high}.}
In addition, Appendix \ref{appendix:consistency_general_feature} discusses the consistency of Lasso regression with general predictors in finite sample.

\begin{theorem}
For a multi-dimensional Brownian motion given by \eqref{equ: MultiBM_setup}, consider a Lasso regression \eqref{equ: lasso} using the It\^o signature with orders truncated to $K$ as predictors.
Let $\rho = \max_{i\neq  j}\{|\rho_{ij}|\}$, $\sigma$ the volatility of $\varepsilon_n$ in \eqref{equ: linearregression}, $q_{\max} = \max_{0 \leq k \leq K} \{ \#A^*_k\}$, and $p$ the number of predictors in the Lasso regression. If \eqref{equ:bound_irc} holds 
and the sequence of regularization parameters $\{\lambda_N\}$ satisfies $\lambda_N > \frac{4\sigma(1-(q_{\max}-1)\rho)}{1-(2q_{\max}-1)\rho} \sqrt{\frac{2\ln p}{N}}$, then the following properties hold with probability greater than 
    \begin{equation}\label{equ:p_min_4}
        P_{\min}^1:= \left(1-\frac{8p^4\sigma_{\max}^4(\sigma_{\min}^4+c_1)}{N\xi^2\sigma_{\min}^4}\right)\left(1-4e^{-c_2N\lambda_N^2}\right) 
    \end{equation}
for some positive constants $c_1$ and $c_2$.
\begin{enumerate}[(a)]
    \item The Lasso regression has a unique solution $\hat{{\boldsymbol \beta}}^N(\lambda_N) \in \mathbb{R}^p$ with its support contained within the true support, and $\hat{{\boldsymbol \beta}}^N(\lambda_N)$ satisfies
 $$\left\|\hat{{\boldsymbol \beta}}^N(\lambda_N)-\tilde{\boldsymbol \beta}\right\|_\infty \leq \lambda_N \left[\frac{3-(2q_{\max}-3)\rho}{\left(1-(q_{\max}-1)\rho\right)\left(2+2\rho\right)} + 4\sigma\sqrt{\frac{2q_{\max}^{\frac{1}{2}}}{1-(q_{\max}-1)\rho}}  \right] =: h(\lambda_N);$$
 \item If in addition $\min_{i \in A^*} |\tilde{ \beta}_i| > h(\lambda_N)$, then $\mathrm{sign}(\hat{\boldsymbol \beta}^N(\lambda_N))=\mathrm{sign}(\tilde{\boldsymbol \beta})$.
\end{enumerate}
Here, $\xi = \min\left\{g_\Sigma^{-1}\left(\frac{\left(1-(2q_{\max}-1)\rho\right)\left(1-(q_{\max}-1)\rho\right)}{3-(2q_{\max}-3)\rho}\right),g_\Sigma^{-1}\left(\frac{1-(q_{\max}-1)\rho}{2\sqrt{pq_{\max}}}\right)\right\} >0$ with 
\begin{equation} \label{equ:def_g_x}
g_{\Sigma}(x) = \frac{2x\sigma_{\min}^2 (p-1)\rho}{(\sigma_{\min}^2- x) \left(2\sigma_{\min}^2- x \right) } + \frac{(p-1) x }{\sigma_{\min}^2- x },
\end{equation}
$\sigma_{\min} = \min_{1 \leq i \leq p}\sqrt{\Sigma_{ii}}$, $\sigma_{\max} = \max_{1 \leq i \leq p}\sqrt{\Sigma_{ii}}$, and $\Sigma$ the population covariance matrix of all predictors in {\eqref{equ: linearregression}}. 
\label{th:sampleito}
\end{theorem}

Part (a) of Theorem \ref{th:sampleito} demonstrates that, for a Brownian motion, when using the It\^o signature as predictors, the difference between the coefficients estimated using Lasso regression and the true values can be bounded, leading to the $l_\infty$ consistency. Part (b) shows that the sign consistency of Lasso regression holds if the magnitudes of true parameters are sufficiently large. Both results hold with a probability of at least $P_{\min}^1$. In particular, the lower bound probability \eqref{equ:p_min_4} characterizes how likely the Lasso regression can recover the true set of signature components. { This probability converges to 1 at a polynomial rate of $N^{-1}$ as the number of samples increases without bound. } Clearly, taking partial derivatives yields the following proposition, which illustrates how this probability varies with different parameters of the model.\footnote{Propositions \ref{pro:sampleito} and \ref{pro:samplestratonovich} follow directly from taking partial derivatives with respect to various parameters, and we therefore omit their proofs.} 
\begin{proposition}\label{pro:sampleito}
    Holding other parameters constant, the lower bound of probability $P_{\min}^1$ given by \eqref{equ:p_min_4}
\begin{enumerate}[(i)]
\item decreases with respect to $\rho$, $p$, and $q_{\max}$, which correspond to the upper bound of the inter-dimensional correlation of the Brownian motion, the number of predictors in the Lasso regression, and the number of true predictors, respectively;
\item increases with respect to $N$, the number of sample paths.
\end{enumerate}
\end{proposition}

Proposition \ref{pro:sampleito} demonstrates that the Lasso regression is (more likely to be) consistent when different dimensions of the Brownian motion are less correlated, or when there are fewer predictors and true predictors in the model, or when more samples are observed. These findings align with general intuition.

The following results demonstrate the consistency of Lasso regression when using the Stratonovich signature as predictors for Brownian motion, or using the It\^o or Stratonovich signature as predictors for the OU process.

\begin{theorem}
\label{th:samplestratonovich}
Consider a Lasso regression \eqref{equ: lasso} using 
the Stratonovich signature as predictors for a multi-dimensional Brownian motion given by \eqref{equ: MultiBM_setup}, or the It\^o or Stratonovich signature as predictors for a multi-dimensional OU process given by \eqref{equ: OU_setup}, with orders truncated to $2K$. 
Let $\sigma$ be the volatility of $\varepsilon_n$ in \eqref{equ: linearregression} and $p$ be the number of predictors in the Lasso regression. If the irrepresentable
condition II holds for both $\Psi_{\mathrm{odd}}$ and $\Psi_{\mathrm{even}}$ given by \eqref{equ: def_odd_even_matrix},
and the sequence of regularization parameters $\{\lambda_N\}$ satisfies $\lambda_N > \frac{4\sigma}{\gamma}\sqrt{\frac{2\ln p}{N}}$, then the following properties hold with probability greater than
\begin{equation} \label{equ:p_min_5}
P_{\min}^2:=\left(1-\frac{8p^4\sigma_{\max}^4(\sigma_{\min}^4+c_1)}{N\xi^2\sigma_{\min}^4}\right)\left(1-4e^{-c_2N\lambda_N^2}\right)    
\end{equation}
for some positive constants $c_1$ and $c_2$.
\begin{enumerate}[(a)]
\item     The Lasso regression has a unique solution $\hat{\boldsymbol \beta}^N(\lambda_N) \in \mathbb{R}^p$ with its support contained within the true support, and $\hat{\boldsymbol \beta}^N(\lambda_N)$ satisfies
 $$\left\|\hat{\boldsymbol\beta}^N(\lambda_N)-\tilde{\boldsymbol \beta}\right\|_\infty \leq \lambda_N \left[\frac{\zeta(2+2\alpha\zeta+\gamma)}{2+2\alpha\zeta} + \frac{4\sigma}{\sqrt{\frac{1}{2}C_{\min}}}\right] =: h(\lambda_N);$$

 \item If in addition $\min_{i \in A^*} |\tilde{ \beta}_i| > h(\lambda_N)$, then $\mathrm{sign}(\hat{\boldsymbol\beta}^N(\lambda_N))=\mathrm{sign}(\tilde{\boldsymbol\beta})$.
\end{enumerate}
Here, 
\begin{itemize}
    \item $\alpha = \left\|\Delta_{A^{*c}A^{*}} \right\|_{\infty} = \max\left\{\left\|\Psi_{\mathrm{odd},A^{*c}A^{*}}\right\|_{\infty},\left\|\Psi_{\mathrm{even},A^{*c}A^{*}}\right\|_{\infty} \right\}$;
    \item $\zeta = \left\|\Delta_{A^{*}A^{*}}^{-1}\right\|_{\infty} = \max\left\{\left\|\Psi_{\mathrm{odd},A^{*}A^{*}}^{-1}\right\|_{\infty},\left\|\Psi_{\mathrm{even},A^{*}A^{*}}^{-1}\right\|_{\infty}\right\}$;
    \item $C_{\min} = \Lambda_{\min}(\Delta_{A^{*}A^{*}}) = \min\left\{\Lambda_{\min}(\Psi_{\mathrm{odd},A^{*}A^{*}}),\Lambda_{\min}(\Psi_{\mathrm{even},A^{*}A^{*}})\right\} $;
    \item $\gamma = \min\left\{1-\left\|\Psi_{\mathrm{odd},A^{*c}A^{*}}\Psi_{\mathrm{odd},A^{*}A^{*}}^{-1}\right\|_{\infty},1-\left\|\Psi_{\mathrm{even},A^{*c}A^{*}}\Psi_{\mathrm{even},A^{*}A^{*}}^{-1}\right\|_{\infty} \right\}$;
    \item $\xi = \min\left\{g_\Sigma^{-1}\left(\frac{\gamma}{\zeta(2+2\alpha\zeta+\gamma)}\right),g_\Sigma^{-1}\left(\frac{C_{\min}}{2\sqrt{p}}\right)\right\} >0$;
    \item $g_{\Sigma}(\cdot)$ is defined by \eqref{equ:def_g_x}, $\sigma_{\min} = \min_{1 \leq i \leq p}\sqrt{\Sigma_{ii}}$, $\sigma_{\max} = \max_{1 \leq i \leq p}\sqrt{\Sigma_{ii}}$, and $\Sigma$ is the population covariance matrix of all predictors in {\eqref{equ: linearregression}}.
\end{itemize}
\end{theorem}

In comparison with the result for the It\^o signature of Brownian motion (Theorem \ref{th:sampleito}), Theorem \ref{th:samplestratonovich} is mathematically more involved as a result of the more complex correlation structure (see Theorems \ref{prop: structure_coef} and \ref{prop: structure_coef_S_integral}). { The lower bound probability \eqref{equ:p_min_5} also converges to 1 at a polynomial rate of $N^{-1}$ as the number of samples increases without bound. }Clearly, taking partial derivatives yields the following proposition, which shows how the lower bound probability $P_{\min}^2$ varies with the parameters. 

\begin{proposition}\label{pro:samplestratonovich}
    Holding other parameters constant, the lower bound of probability $P_{\min}^2$ given by \eqref{equ:p_min_5}
\begin{enumerate}[(i)]    
    \item decreases with respect to $\alpha$ and $p$, which correspond to the upper bound for the correlation between true predictors and false predictors 
    % , the $l_2$-norm of the correlation matrix of the true predictors, 
    and the number of predictors in the Lasso regression, respectively;
    
    \item increases with respect to $\gamma$ and $N$, which correspond to the degree of compliance with the irrepresentable condition II for the population correlation matrix and the number of samples, respectively.
\end{enumerate}
\end{proposition}
Like the result for the It\^o signature of Brownian motion (Proposition \ref{pro:sampleito}), Proposition \ref{pro:samplestratonovich} demonstrates that the Lasso regression is (more likely to be) consistent when there are fewer predictors in the model or when more samples are observed. Furthermore, a lower correlation between predictors and greater compliance with the irrepresentable condition both improve the consistency.

\section{Simulation}\label{sec:simulation}

We use numerical simulations to illustrate our theoretical results and gain additional insights into the consistency of Lasso regression for signature transform.\footnote{Appendix \ref{appendix: discrete} provides technical details, computational cost, more numerical experiments, and robustness checks for the simulations conducted in this section.}

\subsection{Consistency}
\label{sec:simulation_consistency}
Consider a two-dimensional ($d=2$) Brownian motion with inter-dimensional correlation $\rho$.\footnote{The choice of $d=2$ is consistent with the simulation setup commonly used in the literature on signatures; see, for example, \cite{chevyrev2016primer}.}
Assume that there are $q=\# A^*$ true predictors in the true model \eqref{equ: linearregression}, all of which are signature components up to order $K=4$. We follow the steps below to perform our experiment.
\begin{enumerate}
    \item Randomly choose $q$ true predictors from all $\frac{d^{K+1}-1}{d-1} = 31$ signature components; 
    \item Randomly set each beta coefficient of these true predictors from the standard normal distribution; 
    \item Generate 100 samples from this true model with error term $\varepsilon_n$ drawn from a normal distribution with mean zero and standard deviation 0.01; 
    \item Run a Lasso regression given by \eqref{equ: lasso} to select predictors based on these 100 samples; 
    \item Check whether the Lasso regression is sign consistent according to Definition~\ref{def: signconsistency}. 
\end{enumerate}
We then repeat the above procedure 1,000 times and calculate the \emph{consistency rate}, which is defined as the proportion of consistent results among these 1,000 experiments.

Figure \ref{fig: Experiment_3} shows the consistency rates for different values of inter-dimensional correlation $\rho$ and true predictors $q$, with Figure \ref{fig: Experiment_3_BM} for Brownian motion and Figure \ref{fig: Experiment_3_RW} for its discrete counterpart---the random walk. First, signatures for both Brownian motion and random walk are similar. They both exhibit higher consistency rates when the absolute value of $\rho$ is small, i.e., when the inter-dimensional correlation of either Brownian motion or random walk is weak. 
Second, as the number of true predictors $q$ increases, both consistency rates decrease. These findings are consistent with Theorem \ref{prop: sufficient_irrepresentable}, Theorem \ref{th:sampleito}, and Proposition \ref{pro:sampleito}.

\centerline{[Insert \Cref{fig: Experiment_3} approximately here.]}

Furthermore, the consistency rates for the It\^o signature are consistently higher than those for the Stratonovich signature, with $\rho$ and $q$ fixed. 
This is consistent with Theorems \ref{prop: structure_coef} and \ref{prop: structure_coef_S_integral}---signature components of different orders are uncorrelated using the It\^o signature but correlated using the Stratonovich signature. The collinearity between the Stratonovich signature components contributes to their lower consistency for Lasso regression. 

Appendix \ref{appendix:choiceofd} provides additional results for the impact of the number of dimensions $d$ and the number of samples $N$.

\subsection{Predictive Performance}
A higher consistency rate of the Lasso regression is desirable as it is associated with better predictive performance of the model, which we confirm in this section using out-of-sample data. 
 
To this end, we conduct additional simulations for Brownian motion and random walk, following a similar setup as in \Cref{sec:simulation_consistency}, with 200 samples generated from the true model for each experiment. These 200 samples are then equally divided into a training set and a test set, each containing 100 samples. Next, we run a Lasso regression on the training set and choose the tuning parameter $\lambda$ using 5-fold cross-validation. Finally, we calculate the out-of-sample mean squared error (OOS MSE) using the chosen $\lambda$ on the test set. 

\begin{figure}[htbp]
    \centering
    \caption{OOS MSE for the Brownian motion and the random walk with different values of inter-dimensional correlation $\rho$ and different numbers of true predictors $q$. Solid (dashed) lines correspond to the It\^o (Stratonovich) signature. \label{fig: Experiment_otherdef_MSE}}
    \subfigure[Brownian motion.\label{fig: Experiment_otherdef_MSE_BM}]{
        \includegraphics[width=0.45\linewidth]{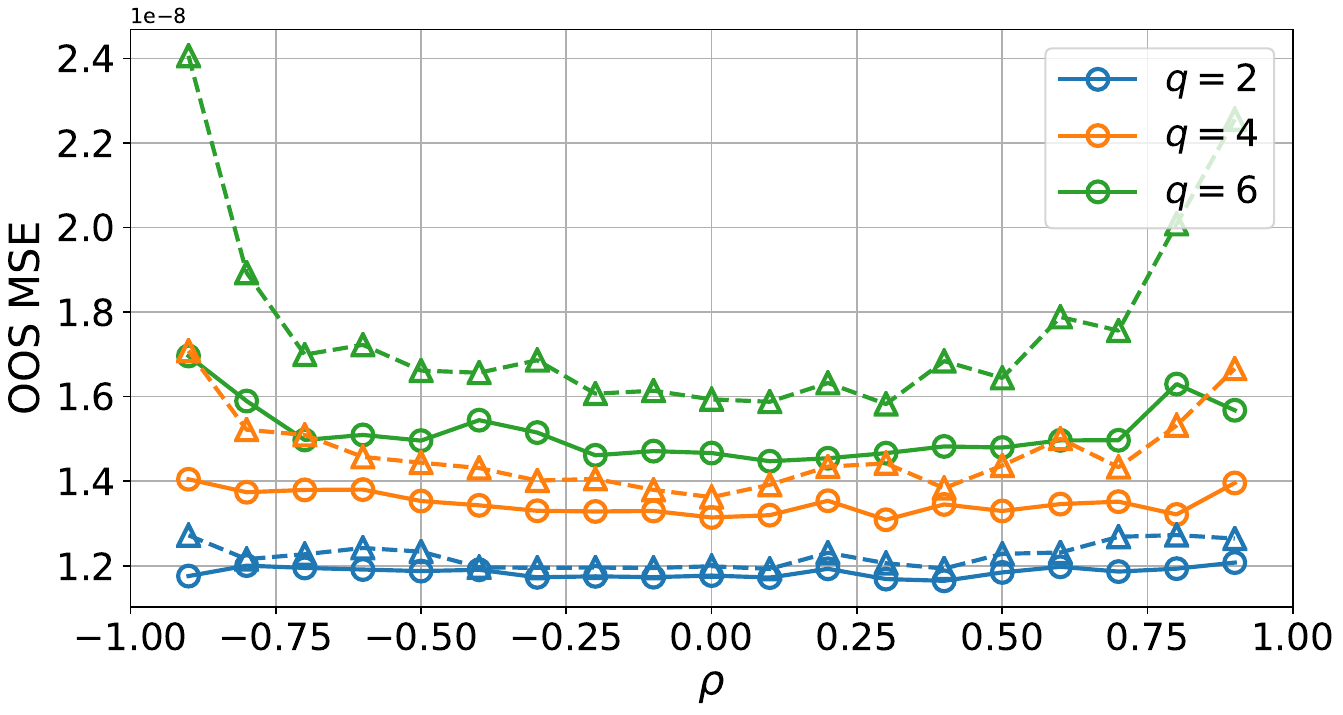}
    }
    % \hfill
    \subfigure[Random walk.\label{fig: Experiment_otherdef_MSE_RW}]{
        \includegraphics[width=0.45\linewidth]{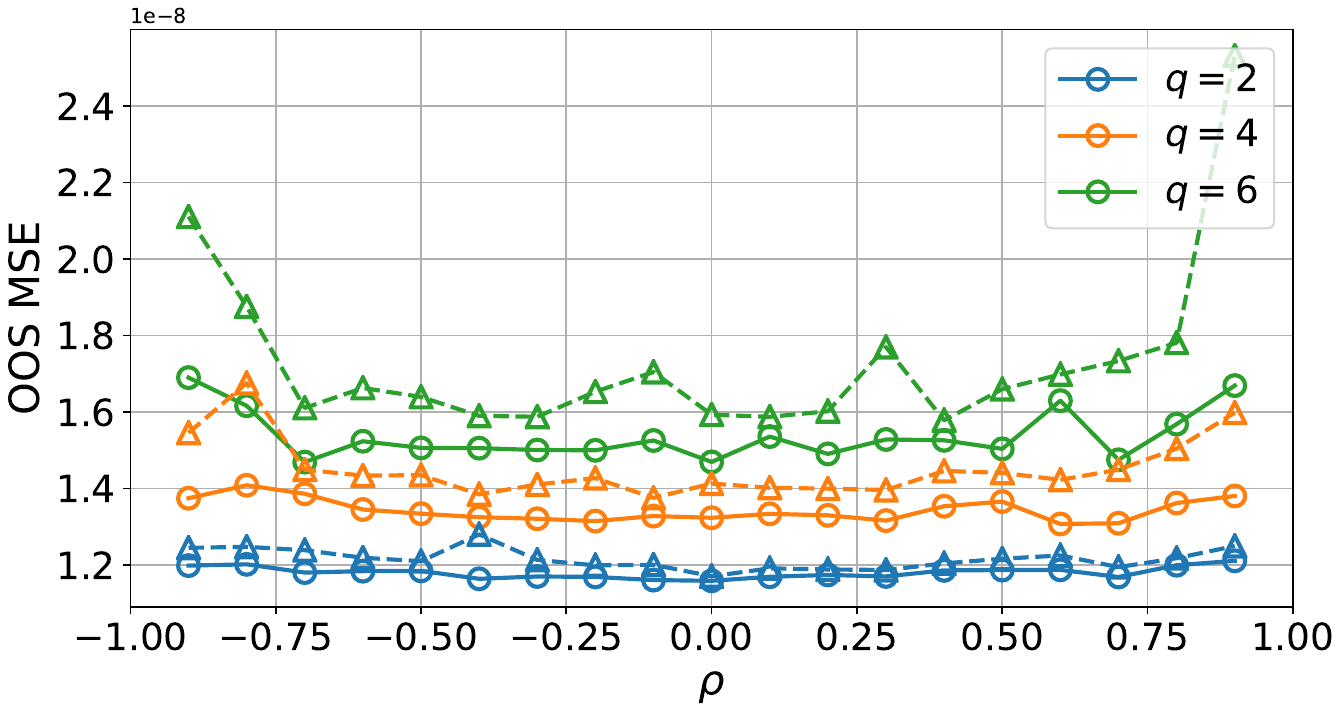}
    }
\end{figure}

Overall, this analysis confirms that the insights derived from sign consistency extend to predictive performance metrics. In particular, Figure \ref{fig: Experiment_otherdef_MSE} shows the OOS MSE for different values of the inter-dimensional correlation $\rho$, and different numbers of true predictors $q$. First, Lasso regression shows lower OOS MSE when the absolute value of $\rho$ is small, i.e., when the inter-dimensional correlations are weak. Second, as the number of true predictors $q$ increases, the OOS MSE increases. Finally, the It\^o signature has a lower OOS MSE compared to the Stratonovich signature with fixed $\rho$ and $q$. 

\centerline{[Insert \Cref{fig: Experiment_otherdef_MSE} approximately here.]}

\subsection{Impact of Mean Reversion}
\label{sec:simulation_mean_revert}
To study the impact of mean reversion on the consistency of Lasso regression, we run simulations for both the OU process and its discrete counterpart---the autoregressive AR(1) model with parameter $\phi$. Recall that higher values of $\kappa$ for the OU process and lower values of $\phi$ for the AR(1) model imply stronger mean reversion. We consider two-dimensional OU and AR(1) processes, with both dimensions sharing the same parameters ($\kappa$ and $\phi$). The inter-dimensional correlation matrix $\Gamma \Gamma^\top$ is randomly drawn from the Wishart$(2,2)$ distribution. All other setups are the same as the Brownian motion experiment in \Cref{sec:simulation_consistency}.

\begin{figure}[htbp]
    \centering
    \caption{Consistency rates for the OU process and the AR(1) model with different parameters ($\kappa$ and $1-\phi$) and different numbers of true predictors $q$. Solid (dashed) lines correspond to the It\^o (Stratonovich) signature. \label{fig: Experiment_12and13}}
    \subfigure[OU process.\label{fig: Experiment_12}]{
        \includegraphics[width=0.45\linewidth]{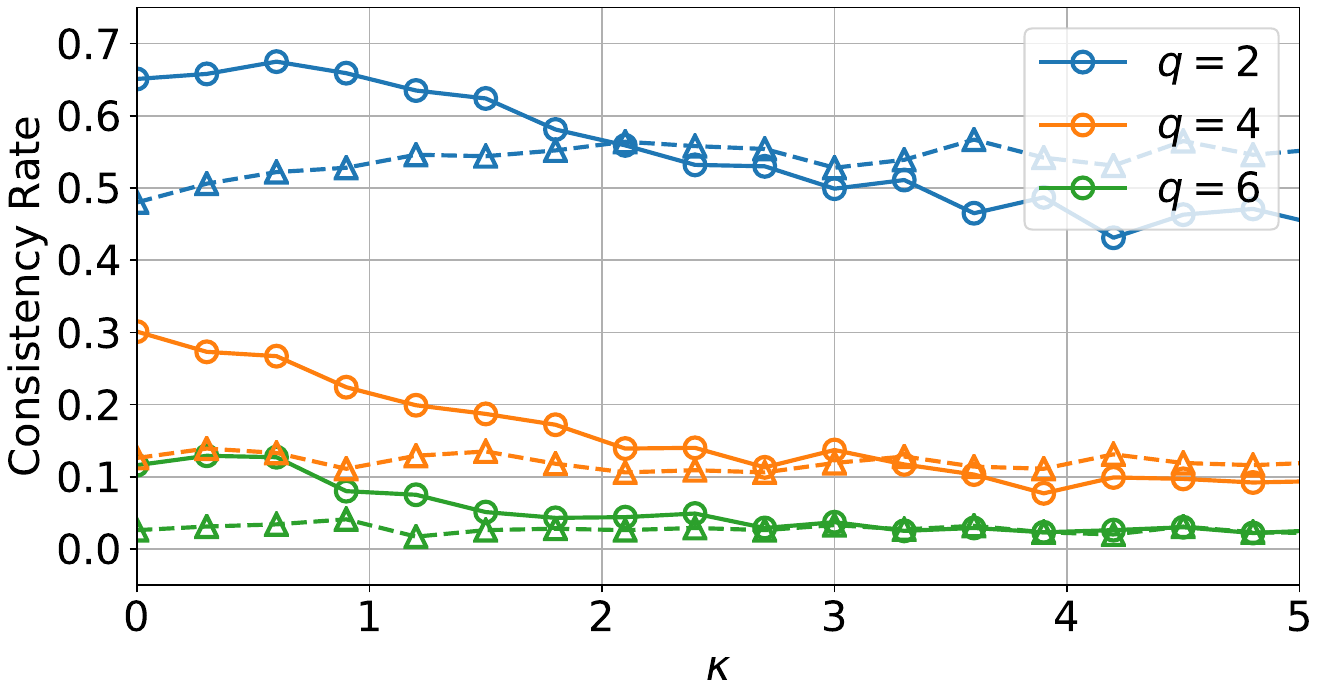}
    }
    % \hfill
    \subfigure[AR(1) model.\label{fig: Experiment_AR}]{
        \includegraphics[width=0.45\linewidth]{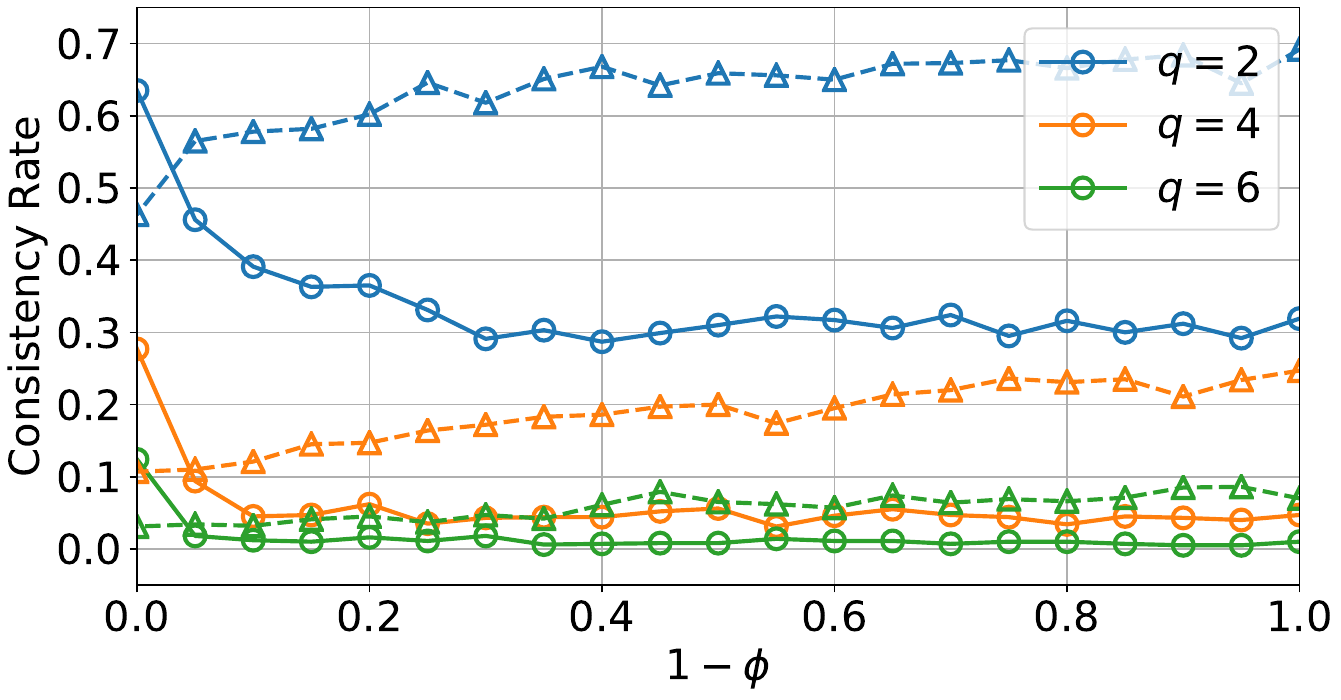}
    }
\end{figure}

Figure \ref{fig: Experiment_12and13} shows the simulation results for the consistency rates of both processes.
First, the It\^o signature reaches the highest consistency rate when $\kappa$ and $1-\phi$ approach $0$, which corresponds to a Brownian motion and a random walk. Second, when the process is sufficiently mean reverting, the Stratonovich signature has higher consistency rates than the It\^o signature.
Finally, Lasso regression becomes less consistent as the number of true predictors $q$ increases, a similar observation as in the experiment for Brownian motion.

\centerline{[Insert \Cref{fig: Experiment_12and13} approximately here.]}

Overall, these results suggest that, for processes that are sufficiently rough or mean reverting \citep{gatheral2018volatility}, using Lasso regression with the Stratonovich signature will likely lead to a higher statistical consistency and better out-of-sample predictive performance compared to the It\^o signature.  Appendix \ref{appendix: calculation_of_corr_OU} provides more theoretical explanations and Appendix \ref{appendix:ARIMA} examines the more complex ARIMA processes.

\section{Applications}\label{sec:application}

In this section, 
we use both the It\^o and the Stratonovich signatures to understand the implication of 
their statistical properties in real applications. 
In particular, Section \ref{subsec:learnpayoff} uses the signature transform to learn option payoffs based on its universal nonlinearity, and Section \ref{subsec:optionpricing} illustrates the application of the signature transform in option pricing.

\subsection{Learning Option Payoffs} \label{subsec:learnpayoff}

Option payoffs are nonlinear functions of the underlying asset. We first show that Lasso regression using signature as predictors can approximate these nonlinear functions well in terms of regression $R^2$. We then use the results derived in \Cref{sec: consistency} and \Cref{sec:simulation} to guide the selection between the It\^o and Stratonovich signatures. 

\subsubsection{Fitting Performance}\label{subsubsec:olsandsig}

We consider two underlying assets, $X^1_t$ and $X^2_t$, both of which following geometric Brownian motions with $X^1_0 = X^2_0=1$, $\mu_1=\mu_2=0$, and $\sigma_1=\sigma_2=0.2$. 
The correlation between the two assets is {0.6}. We consider the following {eight} option payoff functions with time to maturity $T=1$. In the simulation, we employ the Euler-Maruyama method for discretization and divide the time interval into 1000 steps.
    
\begin{enumerate}[(a)]
\item Call option ($d=1$): $\max(X^1_T-1.2,0)$;
\item Put option ($d=1$): $\max(0.8-X^1_T,0)$;
\item Asian option ($d=1$): $\max(\mathrm{mean}_{0\leq t \leq T}(X^1_t)-1.2,0)$;
\item Lookback option ($d=1$): $\max(\max_{0\leq t \leq T}(X^1_t)-1.2,0)$;
\item Rainbow option \Rmnum{1} ($d=2$): $\max(X^1_T-X^2_T,0)$;
\item Rainbow option \Rmnum{2} ($d=2$): $\max(\max(X^1_T,X^2_T)-1.2,0)$;
\item { Rainbow option \Rmnum{3} ($d=2$):
$\max(\max_{0\leq t \leq T}(X^1_t)-\max_{0\leq t \leq T}(X^2_t),0)$;}
\item { Rainbow option \Rmnum{4} ($d=2$): 
$\max(\mathrm{mean}_{0\leq t \leq T}(X^1_t)+\mathrm{mean}_{0\leq t \leq T}(X^2_t)-2.4,0)$.}
\end{enumerate}
The first two are standard options most commonly used in practice; the third and fourth have payoff functions that depend on the entire path of the underlying prices; { the last four} have payoff functions relying on multidimensional underlying paths{, with the fifth and sixth depending only on the terminal values and the seventh and eighth depending on the entire path.}

For each option payoff and for {$K = 6$},\footnote{As a robustness check, we have also conducted experiments for $K \in \{3,4,5,6,7,8,9,10\}$ if $d=1$ and $K \in \{3,4,5,6,7\}$ if $d=2$. The results are similar to the case of $K = 6$. } we perform Lasso regression using the following three different types of predictors.
\begin{enumerate}[(1)]
    \item The Stratonovich signature of the path of the underlying asset(s) with orders up to $K$ (denoted as ``Sig''); 
    \item $p\left(=\frac{d^{K+1}-1}{d-1}\right)$ randomly sampled points from the path of the underlying asset(s) (denoted as ``RSam''); 
    \item $p\left(=\frac{d^{K+1}-1}{d-1}\right)$ equidistant points from the path of the underlying asset(s) (denoted as ``USam'').
\end{enumerate}
The training set for the Lasso regression consists of { 200} simulated paths, and the test set consists of { 100} simulated paths. We repeat each experiment 200 times to derive confidence intervals for the estimates. 

\textbf{Results.} Figure \ref{fig:r2total} shows the relationship between $R^2$ and the penalization parameter of the Lasso regression $\lambda$, when using different types of predictors. 
Both in-sample and out-of-sample $R^2$ for Lasso regression with signature components as predictors consistently outperform those for Lasso regression with random sampling and equidistant sampling as predictors. This demonstrates the effectiveness of the signature transform in approximating various nonlinear payoff functions, thanks to its universal nonlinearity.

\centerline{[Insert \Cref{fig:r2total} approximately here.]}

\begin{figure}[htbp]
    \centering
    \caption{{Lasso paths with signatures as predictors.\label{fig:coeftotal}}}
    \subfigure[Call option.\label{fig:call_coef}]{
        \includegraphics[width=0.37\textwidth]{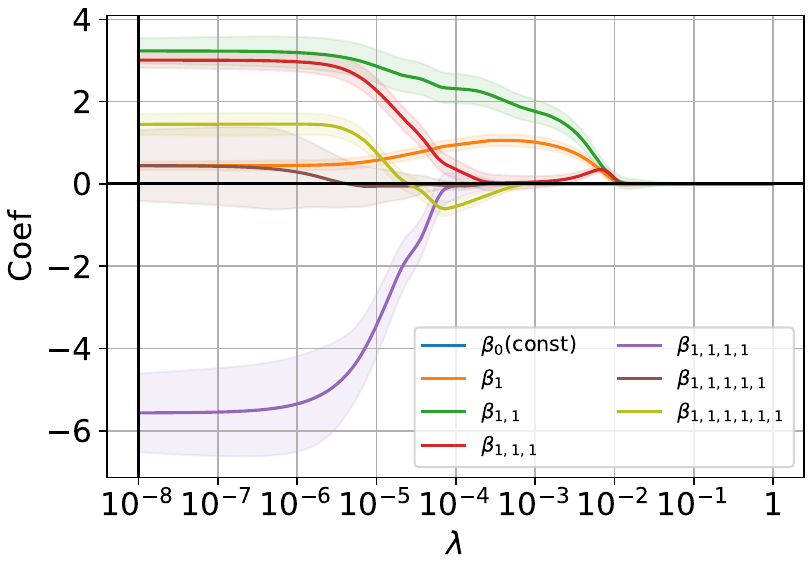}
    }
    % \hfill
    \subfigure[Put option.\label{fig:put_coef}]{
        \includegraphics[width=0.37\textwidth]{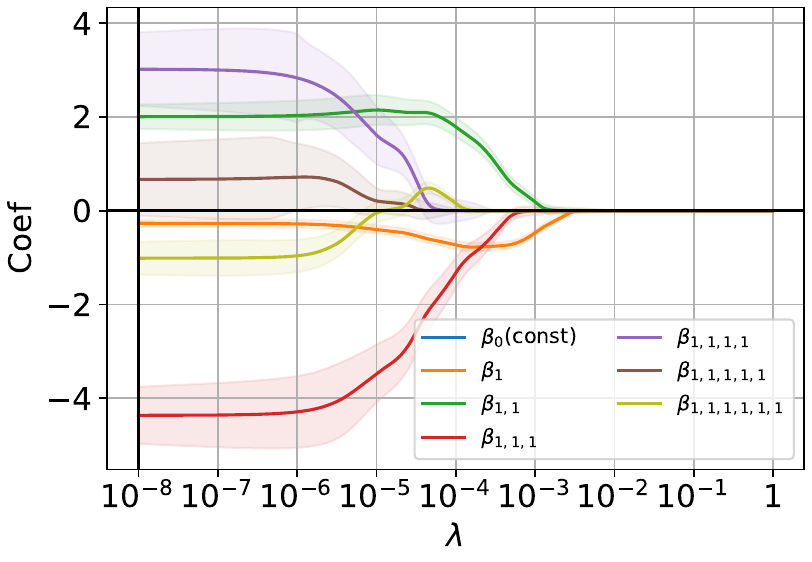}
    }
    % \hfill
    \subfigure[Asian option.\label{fig:asian_coef}]{
        \includegraphics[width=0.37\textwidth]{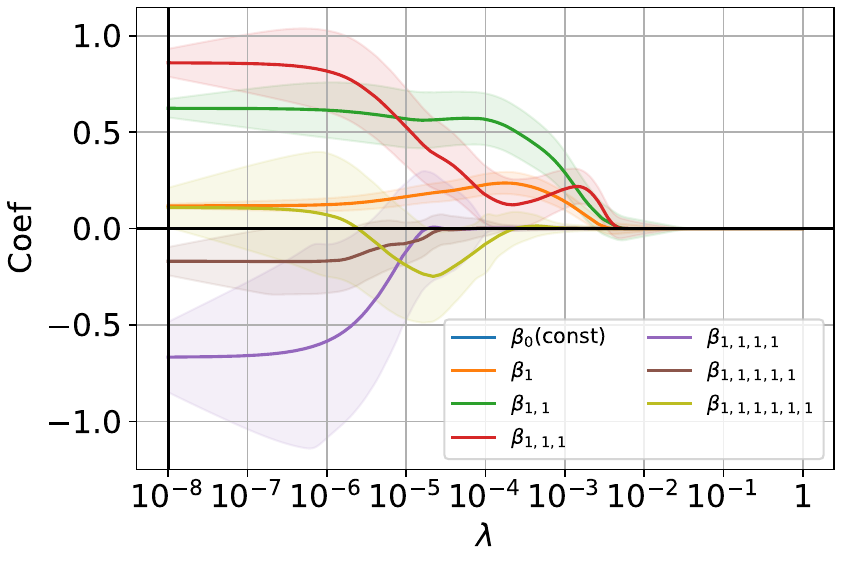}
    }
    % \hfill
    \subfigure[Lookback option.\label{fig:lookback_coef}]{
        \includegraphics[width=0.37\textwidth]{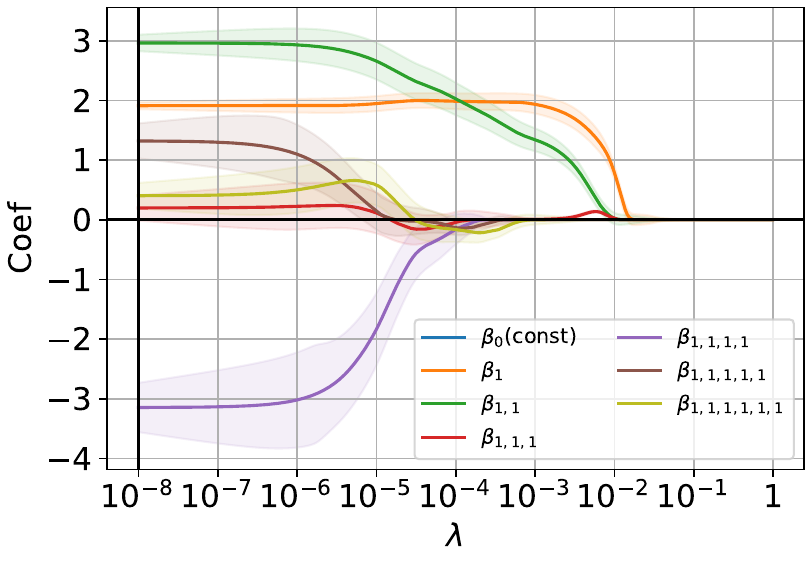}
    }
    % \hfill
    \subfigure[Rainbow option \Rmnum{1}.\label{fig:rainbow1_coef}]{
        \includegraphics[width=0.37\textwidth]{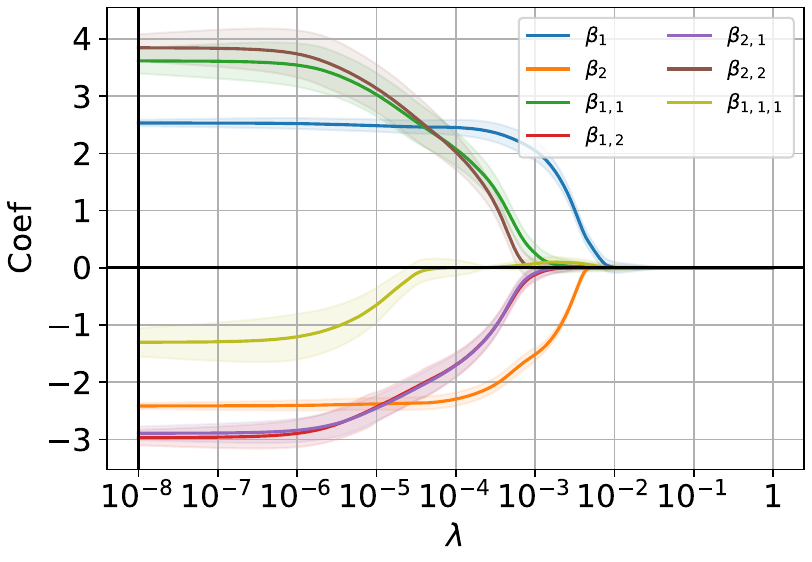}
    }
    % \hfill
    \subfigure[Rainbow option \Rmnum{2}.\label{fig:rainbow2_coef}]{
        \includegraphics[width=0.37\textwidth]{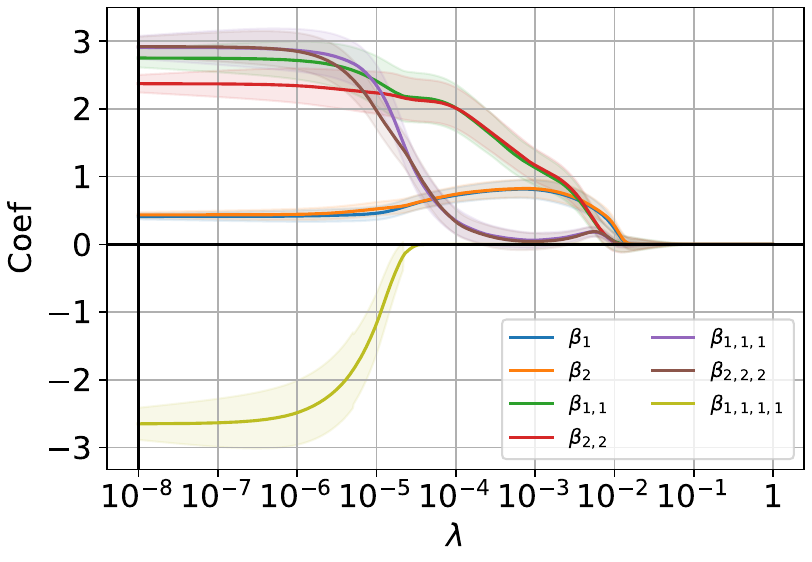}
    }
    % \hfill
    \subfigure[ Rainbow option \Rmnum{3}.\label{fig:rainbow3_coef}]{
        \includegraphics[width=0.37\textwidth]{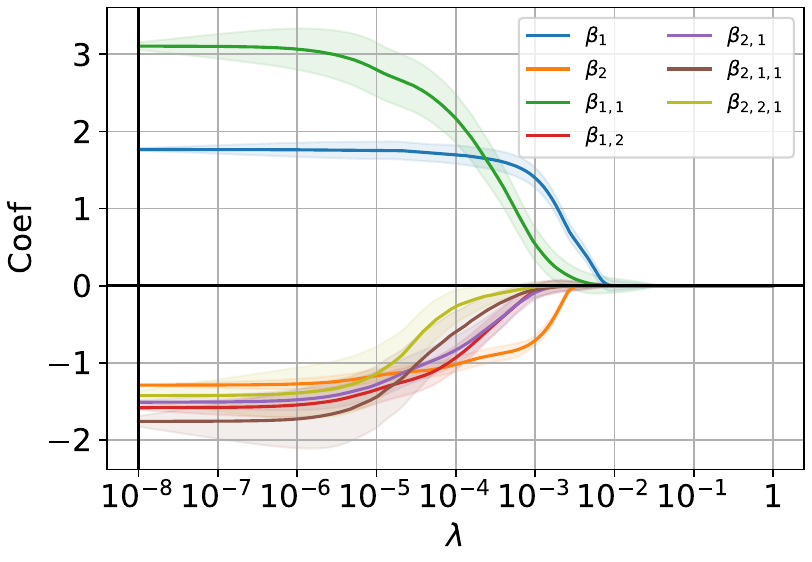}
    }
    % \hfill
    \subfigure[ Rainbow option \Rmnum{4}.\label{fig:rainbow4_coef}]{
        \includegraphics[width=0.37\textwidth]{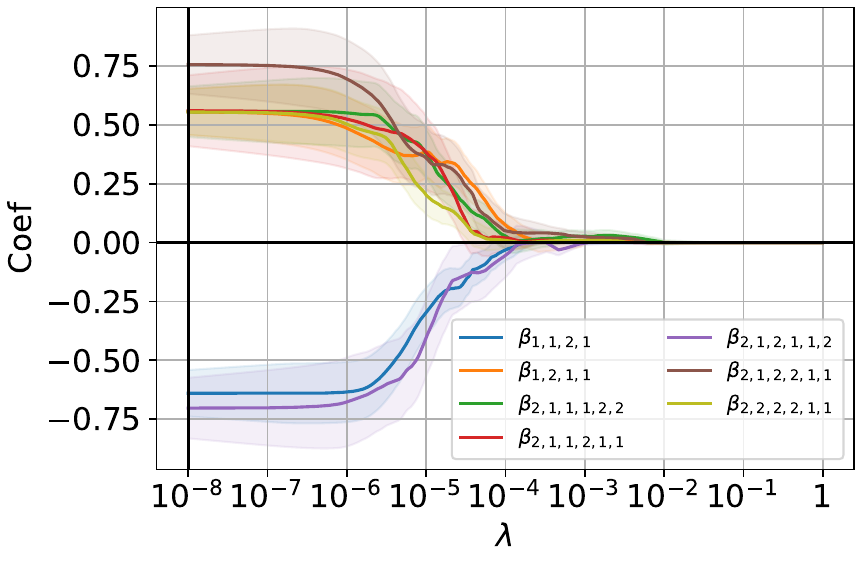}
    }
\end{figure}

Figure \ref{fig:coeftotal} further shows the Lasso paths as a function of the penalization parameter $\lambda$ when using signature components as predictors.\footnote{For rainbow options with two-dimensional underlying price processes, due to their large number of predictors in the Lasso regression, we only show the {seven} predictors with the largest estimated coefficients.} The fairly narrow range of the 90\% confidence intervals of the parameters indicates the stability of the estimated coefficients across repeated experiments, consistent with the uniqueness of the universal nonlinearity of signature (Theorem \ref{thm:uniqueness}).

\centerline{[Insert \Cref{fig:coeftotal} approximately here.]}

\subsubsection{Comparison Between Different Signatures}

We further compare the performance of the It\^o and the Stratonovich signatures in learning option payoffs. Given a discrete time series of an underlying asset $\{X_{t_j}\}_{j=1}^{1000}$ with $t_j = j/1,000$, we consider three numerical methods to calculate the signatures, summarized in Table \ref{tab:threemethods}. The first two are numerical methods for computing the It\^o and Stratonovich integrals, respectively.\footnote{Appendix \ref{appendix: discrete} provides technical details for the schemes of numerically computing both integrals.} The third method, called Linear, linearly interpolates the time series and then calculates the signature using Riemann/Lebesgue integrals, which is widely adopted in practice \citep{lyons2022signature}.
 
\centerline{[Insert \Cref{tab:threemethods} approximately here.]}

\begin{figure}[htbp]
    \centering
    \caption{{ Out-of-sample} $R^2$ when using different methods for computing signature given in Table \ref{tab:threemethods}.\label{fig:ItoSLinear_r2}}
    \subfigure[Brownian motion.\label{fig:brown_r2}]{
        \includegraphics[width=0.45\linewidth]{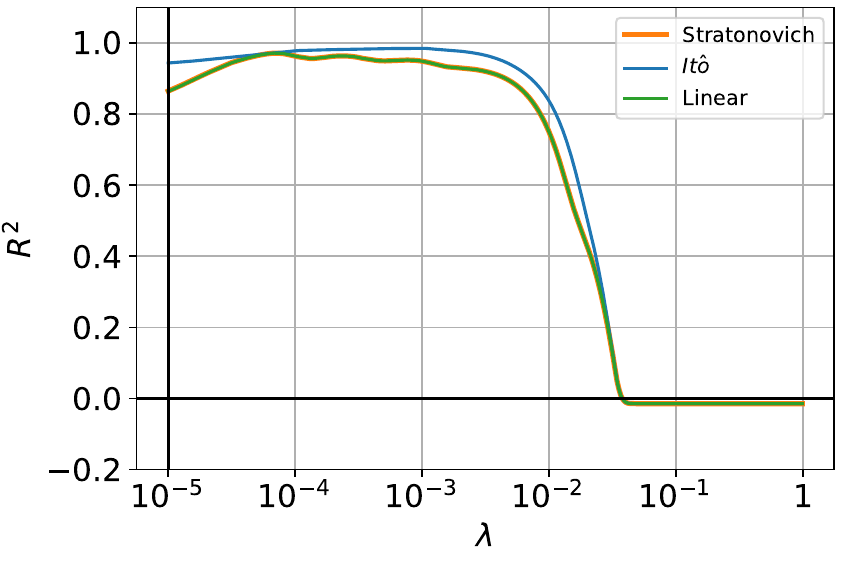}
    }
    % \hfill
    \subfigure[OU process.\label{fig:ou_r2}]{
        \includegraphics[width=0.45\linewidth]{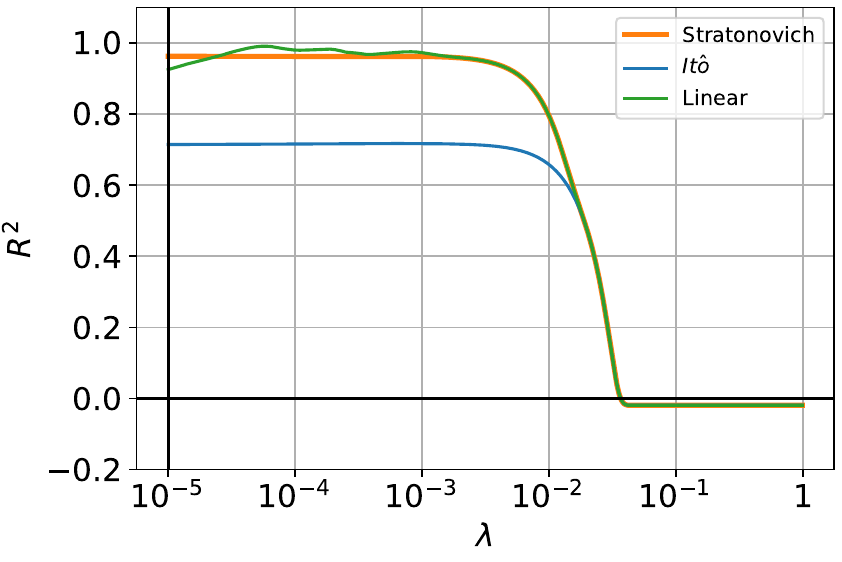}
    }
\end{figure}

We simulate two different types of processes for the underlying asset: a one-dimensional standard Brownian Motion and a one-dimensional standard OU process with mean-reverting parameter $\kappa=1$. As an example, we consider the payoff $\max(X_T,0)$ with time to maturity $T=1$.
Similar to the settings in Section \ref{subsubsec:olsandsig}, the training set for the Lasso regression consists of { 200} simulated paths, and the test set consists of {100} simulated paths. Each experiment is repeated 200 times and the average out-of-sample $R^2$ is shown in Figure \ref{fig:ItoSLinear_r2}.{\footnote{ We omit the values of in-sample $R^2$ as they are very close to the values of out-of-sample $R^2$.}}

\centerline{[Insert \Cref{fig:ItoSLinear_r2} approximately here.]}

\textbf{Results.} First, the It\^o signature of the Brownian motion outperforms its Stratonovich signature in Lasso regression. Second, the Stratonovich signature of the OU process outperforms its It\^o signature. These findings are consistent with our theoretical results in \Cref{sec: correlationstructure} that the It\^o signature components of a Brownian motion are more uncorrelated compared to the Stratonovich signature, as well as our simulation results in \Cref{sec:simulation_mean_revert}. Third, the performance of the Linear method in \Cref{tab:threemethods} is almost identical to the results for the Stratonovich signature, suggesting that for a mean-reverting time series, it may be reasonable to use the heuristic of linearly interpolating the time series and then calculating the signature using Riemann/Lebesgue integrals as in \cite{lyons2022signature}. If the underlying time series is closer to the path of a Brownian motion, using the It\^o signature may lead to improved performance compared to current heuristics.

\subsection{Option Pricing}\label{subsec:optionpricing}

The effectiveness of the signature in learning option payoffs in the previous section suggests a new way to price and hedge options.\footnote{For example, \cite{arribas2018derivatives} and \cite{lyons2019numerical} price options using securities whose payoffs are signature. \cite{lyons2020non} use the signature transform to perform option hedging. \cite{bayraktar2024deep} propose a deep learning framework to price options using signature. } 
In this section, we follow the method proposed by \cite{lyons2019numerical} to use the signature to price stock options and interest rate options, which are two of the most important and widely traded options in the equity and fixed income markets, respectively. In addition, the dynamics of the prices of their underlying assets have different statistical properties---the former resembling a Brownian motion, while the latter resembling a mean-reverting process.

\subsubsection{Method}

We consider a set of $m$ options actively traded in the market with different payoffs $A_1, A_2, \dots, A_m$, and their prices are observable (referred to as ``source options''). Our goal is to determine the prices of a different set of $n$ options with payoffs $B_1, B_2, \dots, B_n$ (referred to as ``target options''). Both sets of options share the same underlying asset with path $\mathbf{X}_t$. Therefore, their payoffs, $A_i$ and $B_j$, are (different) functions of $\mathbf{X}$. For simplicity, we assume that they also share the same maturity. 

To price the target options, we consider the first $K$ signature components of the underlying asset, $S_0(\mathbf{X}), S_1(\mathbf{X}), \dots, S_K(\mathbf{X})$. The universal nonlinearity implies that
\begin{equation}\label{equ:reg_A}
    A_i(\mathbf{X}) \approx a_{i,0} S_0(\mathbf{X}) + a_{i,1} S_1(\mathbf{X}) + \dots + a_{i,K} S_K(\mathbf{X}), \quad i=1,2,\dots, m
\end{equation}
and
\begin{equation}\label{equ:reg_B}
    B_j(\mathbf{X}) \approx b_{j,0} S_0(\mathbf{X}) + b_{j,1} S_1(\mathbf{X}) + \dots + b_{j,K} S_K(\mathbf{X}), \quad j=1,2,\dots, n.
\end{equation}
The coefficients $a_{i,\cdot}$ and $b_{j,\cdot}$ can be estimated using Lasso regression based on data from both sets of options because their payoffs $A_i(\mathbf{X})$ and $B_j(\mathbf{X})$ are known given underlying paths. 
% By taking expectations for \eqref{equ:reg_A} and \eqref{equ:reg_B} in the risk-neutral world and discounting to the present, we obtain
Financial assets with identical payoffs must have identical prices assuming no arbitrage, thus from \eqref{equ:reg_A} and \eqref{equ:reg_B} we obtain the prices of the options
\begin{equation}\label{equ:price_A}
    p(A_i) \approx a_{i,0} p(S_0) + a_{i,1} p(S_1) + \dots + a_{i,K} p(S_K), \quad i=1,2,\dots, m
\end{equation}
and
\begin{equation}\label{equ:price_B}
    p(B_j) \approx b_{j,0} p(S_0) + b_{j,1} p(S_1) + \dots + b_{j,K} p(S_K), \quad j=1,2,\dots, n,
\end{equation}
where $p(\cdot)$ denotes the price of a derivative. Because $p(A_i)$ are observable from source options, we can estimate $p(S_0),\dots,p(S_K)$ using \eqref{equ:price_A}, and then predict $p(B_j)$ for target options using \eqref{equ:price_B}.

We point out that this method is interpretable because the signature linearizes the problem of feature selection.
In particular, $p(S_0), p(S_1), \dots, p(S_K)$ can be understood as the prices of $K$ latent derivatives whose payoff functions are given by the first $K$ signature components of the underlying asset, $S_0, S_1, \dots, S_K$. This is analogous to the Arrow--Debreu state prices of \possessivecite{ross:1976} arbitrage pricing theory, and therefore we refer to these latent derivatives as ``signature derivatives.'' 

Based on this framework, we summarize the procedure for estimating $p(B_j)$ in the following steps.
\begin{enumerate}[(i)]
    \item Simulate $N$ paths of $\mathbf{X}$: $\mathbf{X}_1, \mathbf{X}_2, \dots, \mathbf{X}_N$;
    \item Calculate the payoffs of source options $A_i(\mathbf{X}_1),A_i(\mathbf{X}_2),\dots,A_i(\mathbf{X}_N)$ for $i=1,2,\dots,m$, and target options $B_j(\mathbf{X}_1),B_j(\mathbf{X}_2),\dots,B_j(\mathbf{X}_N)$ for $j=1,2,\dots,n$;
    \item Calculate the corresponding signature $S_k(\mathbf{X}_1),S_k(\mathbf{X}_2),\dots,S_k(\mathbf{X}_N)$ for $k=0,1,\dots,K$;
    \item For each $i$ or $j$, estimate $a_{i,\cdot}$ and $b_{j,\cdot}$ using Lasso regression based on \eqref{equ:reg_A} and \eqref{equ:reg_B}, respectively, where the predictors are $S_k(\mathbf{X}_1),S_k(\mathbf{X}_2),\dots,S_k(\mathbf{X}_N)$ for $k=0,1,\dots,K$, the dependent variables are $A_i(\mathbf{X}_1),\dots,A_i(\mathbf{X}_N)$ or $B_j(\mathbf{X}_1),\dots,B_j(\mathbf{X}_N)$, and the regression is performed with $N$ samples;
    \item Estimate the prices of signature derivatives $p(S_0), p(S_1), \dots, p(S_K)$ using ordinary least squares based on \eqref{equ:price_A}, where the predictors are $a_{i,\cdot}$, the dependent variables are $p(A_i)$, and the estimation uses $m$ samples;
    \item For each $j$, calculate the price of the target option $p(B_j)$ using \eqref{equ:price_B} directly.
\end{enumerate}

\subsubsection{Stock Options} \label{subsubsec:stock}

Following the Black--Scholes--Merton framework \citep{black1973pricing,merton1973theory}, we assume that the underlying asset $\mathbf{X}$ follows a geometric Brownian motion in the risk-neutral world with initial price $100$, risk-free rate $2\%$, dividend yield $0\%$, and volatility $20\%$. The source options are vanilla European calls and puts with strikes at $90, 92, 94, \dots, 110$ ($m=22$), priced by the Black--Scholes--Merton formula. The target options are vanilla European calls and puts with strikes at $91, 93, 95, \dots, 109$ ($n=20$), and their true prices are determined using the Black--Scholes--Merton formula. The times to maturity for all these options are set to be $2.5$ years. 

The simulation is conducted as follows. We simulate $N=1,000$ paths for the underlying asset, and the step size for simulating the underlying path is $1/252$ (one trading day). For each Lasso regression, signature components with orders up to 4 are used as predictors ($K=4$), and the penalization parameter is determined using five-fold cross-validation. 

The experiment is repeated 100 times, and the estimation error is computed for each experiment. The relative error is measured using the average of $|\hat{p}(B_j) - p(B_j)| / p(B_j)$ across all target options, where $\hat{p}(B_j)$ and $p(B_j)$ are the estimated price and true price of $B_j$, respectively. 

\begin{figure}[htbp]
    \centering
    \caption{Estimated prices versus the true prices for stock options.\label{fig:Scatter_optionpricing}}
    \subfigure[It\^o signature.\label{fig:Scatter_optionpricing_I}]{
        \includegraphics[width=0.45\linewidth]{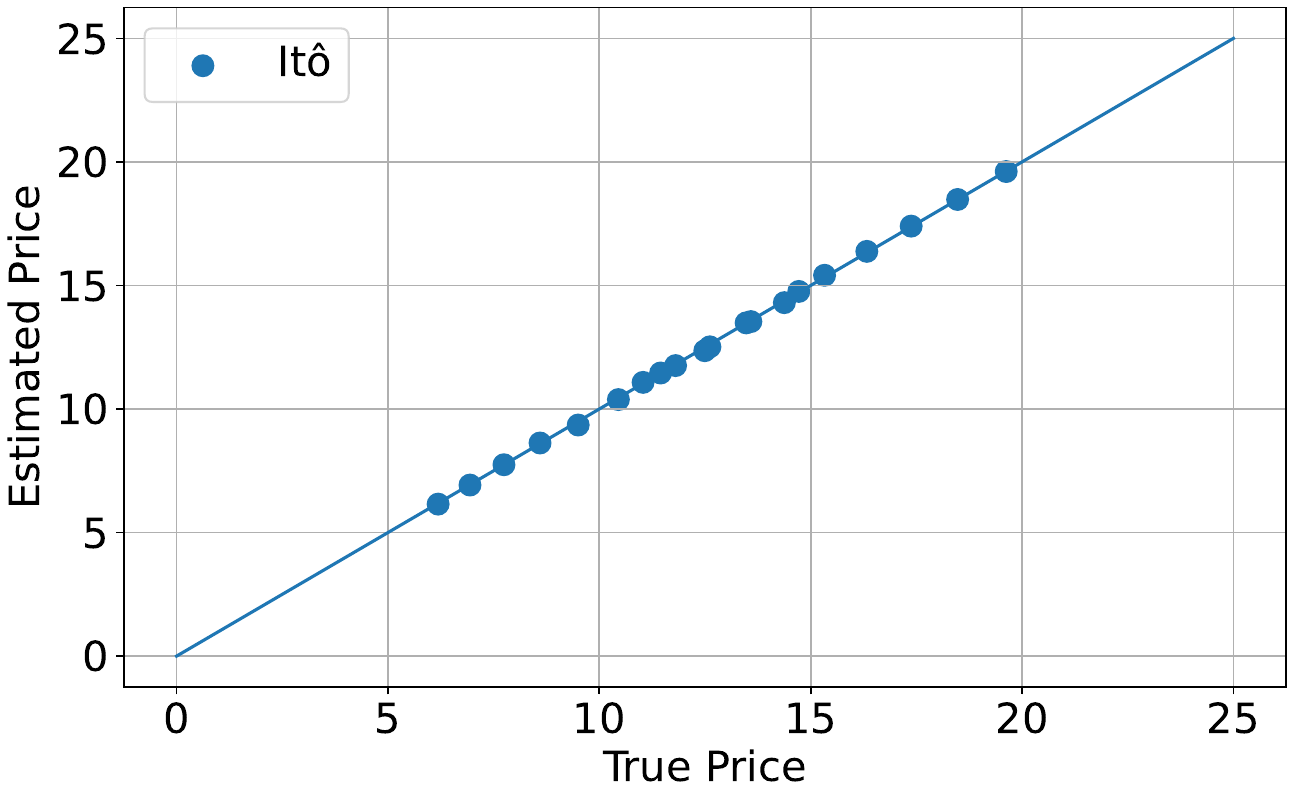}
    }
    % \hfill
    \subfigure[Stratonovich signature.\label{fig:Scatter_optionpricing_S}]{
        \includegraphics[width=0.45\linewidth]{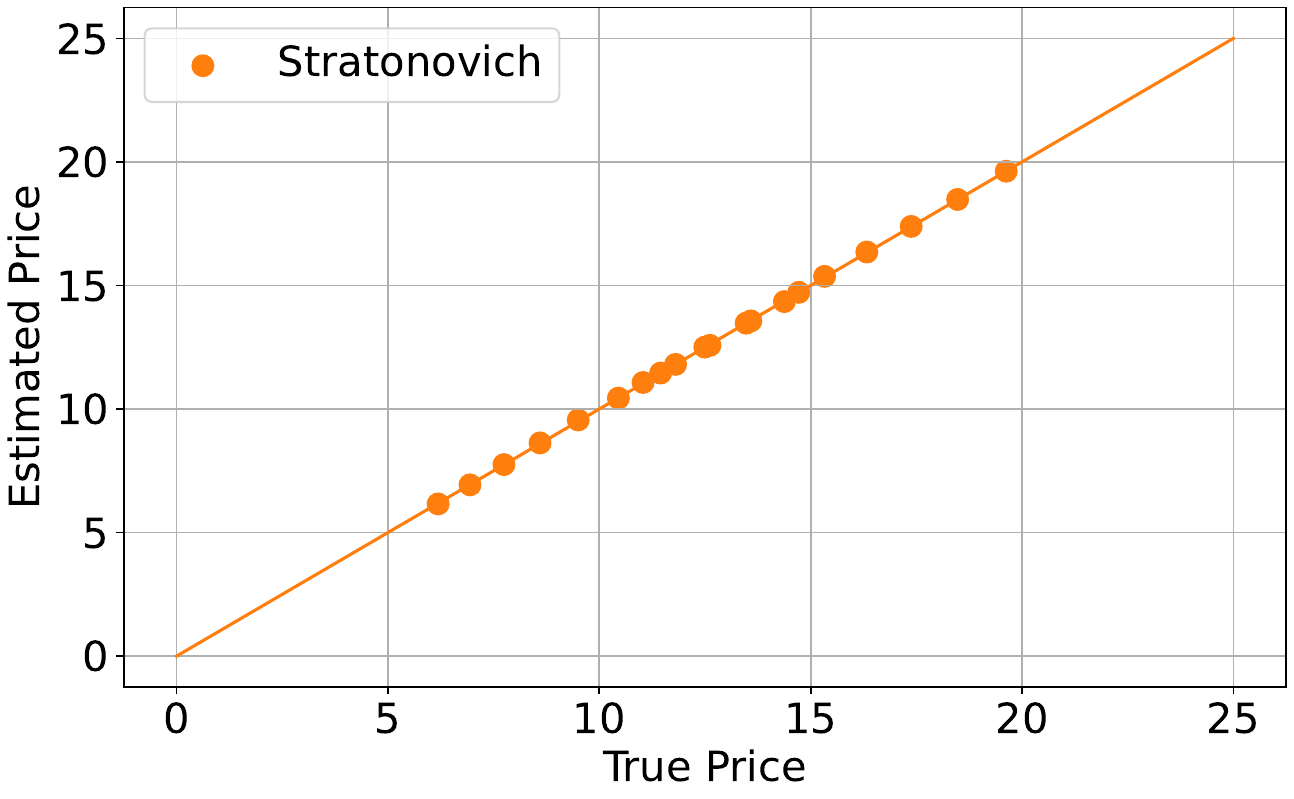}
    }
\end{figure}

\textbf{Results.} Using signature with Lasso regression provides an excellent fit for the prices of stock options. Figure \ref{fig:Scatter_optionpricing} shows the true prices and the estimated prices of the target options for a randomly chosen experiment out of 100, showing that the estimation errors are small for both the It\^o and Stratonovich signatures.

\centerline{[Insert \Cref{fig:Scatter_optionpricing} approximately here.]}

\begin{figure}[htbp]
    \centering
    \caption{Estimation errors for different target stock options.\label{fig:Error_band}}
    \subfigure[Call options.\label{fig:Error_band_call}]{
        \includegraphics[width=0.45\linewidth]{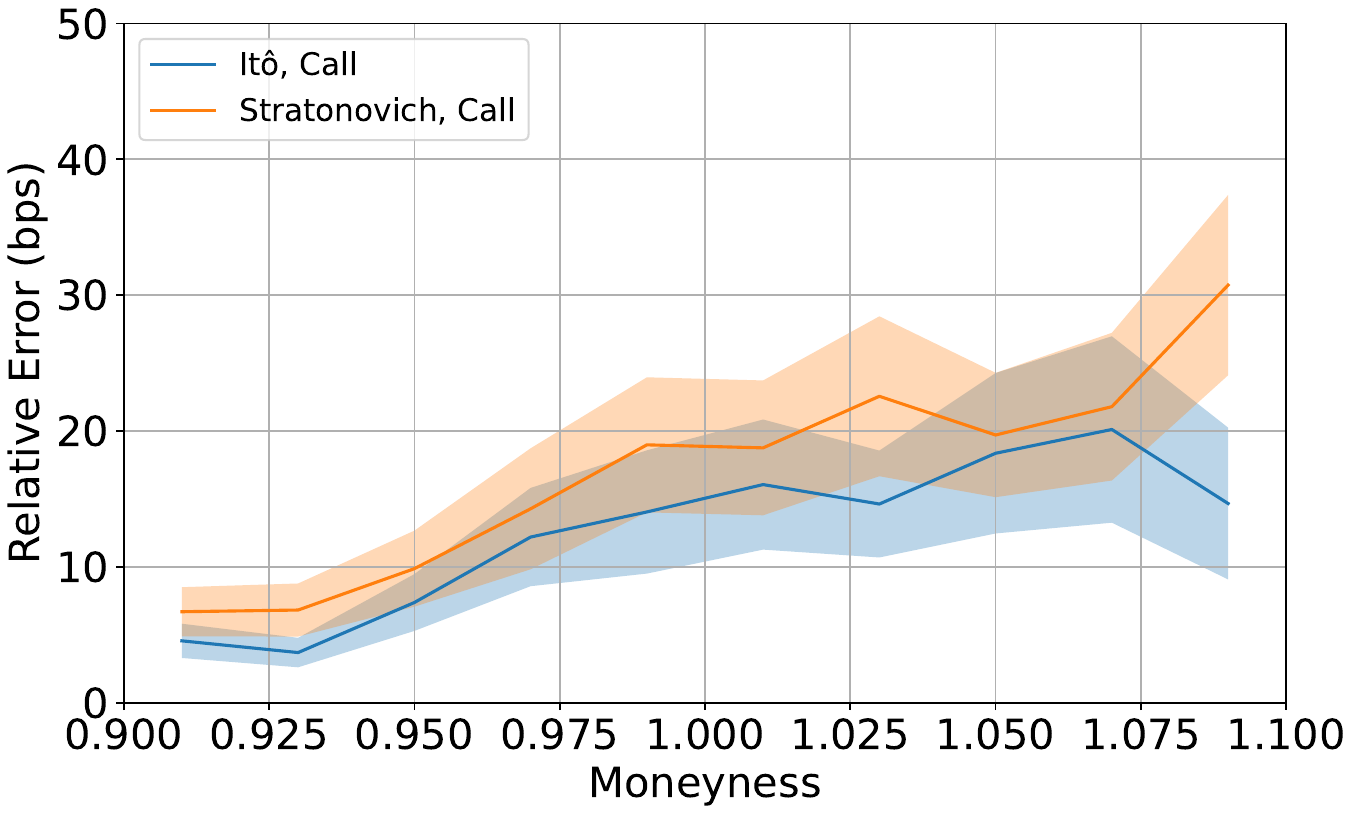}
    }
    % \hfill
    \subfigure[Put options.\label{fig:Error_band_put}]{
        \includegraphics[width=0.45\linewidth]{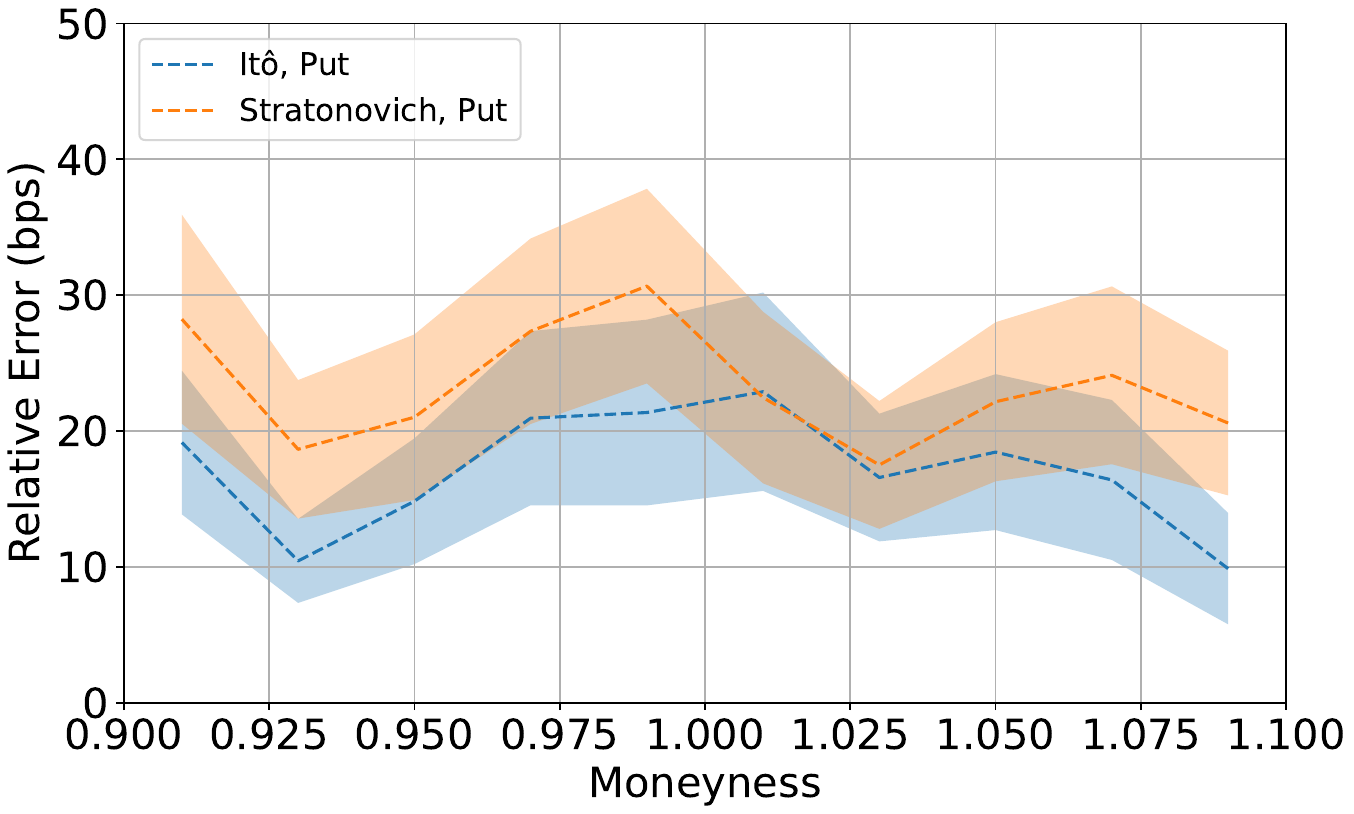}
    }
\end{figure}

In addition, the estimation error of stock option prices when using the It\^o signature is lower compared to using the Stratonovich signature, consistent with our results in \Cref{sec:simulation_mean_revert}. 
Figure \ref{fig:Error_band} shows the average relative errors for the It\^o and the Stratonovich signatures across the 100 experiments. The $x$-axis represents the moneyness (the strike price of the option divided by the initial asset price) of the target options, and the $y$-axis is the average relative error. Note that the estimation error of the It\^o signature is lower because the underlying price process resembles a Brownian motion. 

\centerline{[Insert \Cref{fig:Error_band} approximately here.]}

\subsubsection{Interest Rate Options}

We now turn to interest rate options, whose underlying assets are commonly modeled using mean-reverting processes. 
In particular, consider the interest rate processes $\{r_t\}_{t\geq0}$ by the classical Vasicek model \citep{vasicek1977equilibrium} in the risk-neutral world\footnote{This is also known as a special case of the Hull--White model \citep{hull1990pricing}; see, for example, \cite{veronesi2010fixed}.} 
    \begin{equation*}
        \mathrm{d} r_t = \gamma ( \bar{r} - r_t ) \mathrm{d} t + \sigma \mathrm{d} W_t,
    \end{equation*}
with initial rate $r_0=3\%$, long-term average interest rate $\bar{r}=3\%$, mean-reverting intensity $\gamma = 0.1$, volatility $\sigma = 2\%$, and $W_t$ a standard Brownian motion. The source options are interest rate caplets and floorlets with strikes $r_{\mathrm{strike}}=2.50\%, 2.60\%, \dots, 3.40\%, 3.50\%$ ($m=22$), and their prices $p(A_i)$ are determined using explicit formulas for caplets and floorlets under the Hull--White model (see, for example, \cite{veronesi2010fixed}). The target options are interest rate caplets and floorlets with strikes $r_{\mathrm{strike}}=2.55\%, 2.65\%, \dots, 3.35\%, 3.45\%$ ($n=20$). The payoffs for caplets and floorlets are $\max( r(0.5, 1)-r_{\mathrm{strike}},0 )$ and $\max( r_{\mathrm{strike}}-r(0.5, 1),0 )$, respectively, where $r(0.5, 1)$ is the 0.5-year interest rate at time 0.5. Assume that each of these instruments has a notional value of $\$100$ and a maturity of 0.5 years. Other simulation setups are the same as in Section \ref{subsubsec:stock}, and the experiment is repeated 100 times.

\begin{figure}[htbp]
    \centering
    \caption{Estimated prices versus the true prices for interest rate options.\label{fig:Scatter_optionpricing_interest}}
    \subfigure[It\^o signature.\label{fig:Scatter_optionpricing_I_interest}]{
        \includegraphics[width=0.45\linewidth]{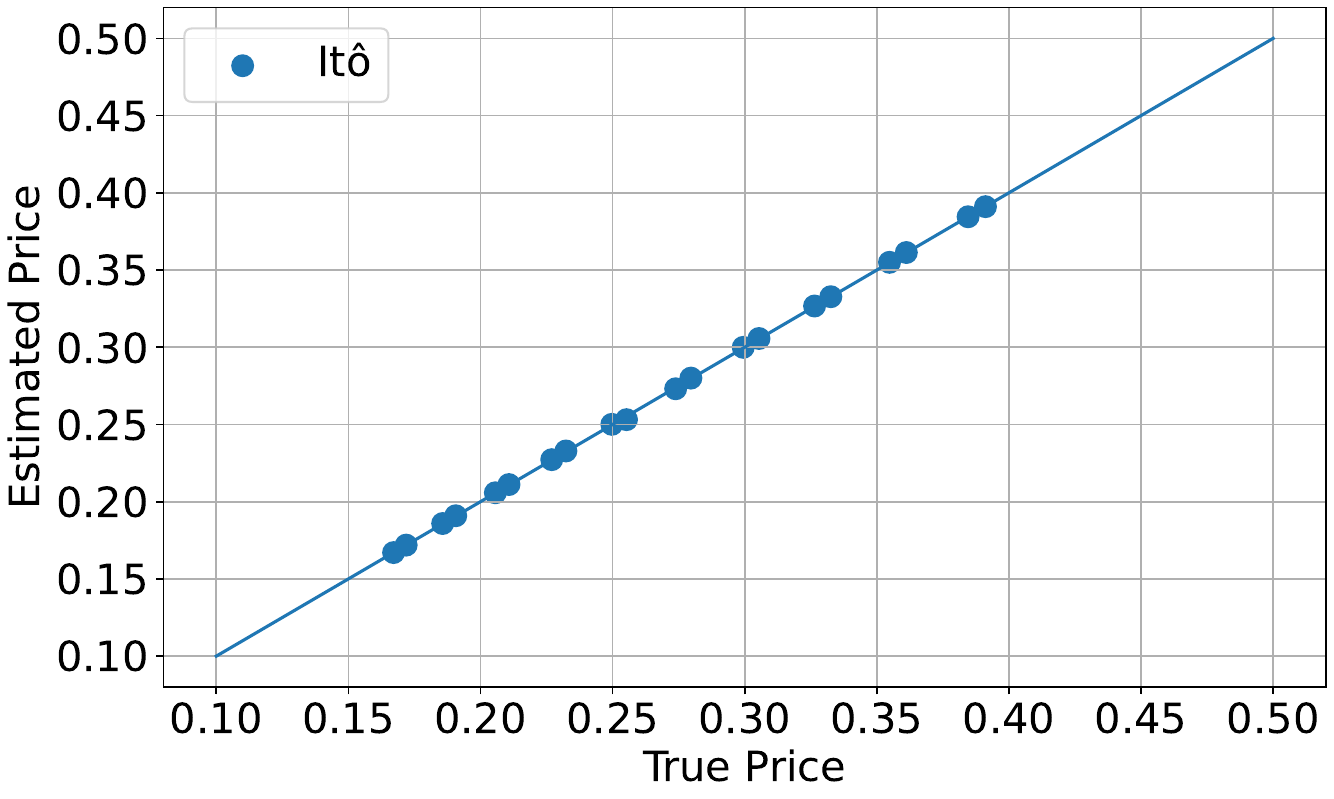}
    }
    % \hfill
    \subfigure[Stratonovich signature.\label{fig:Scatter_optionpricing_S_interest}]{
        \includegraphics[width=0.45\linewidth]{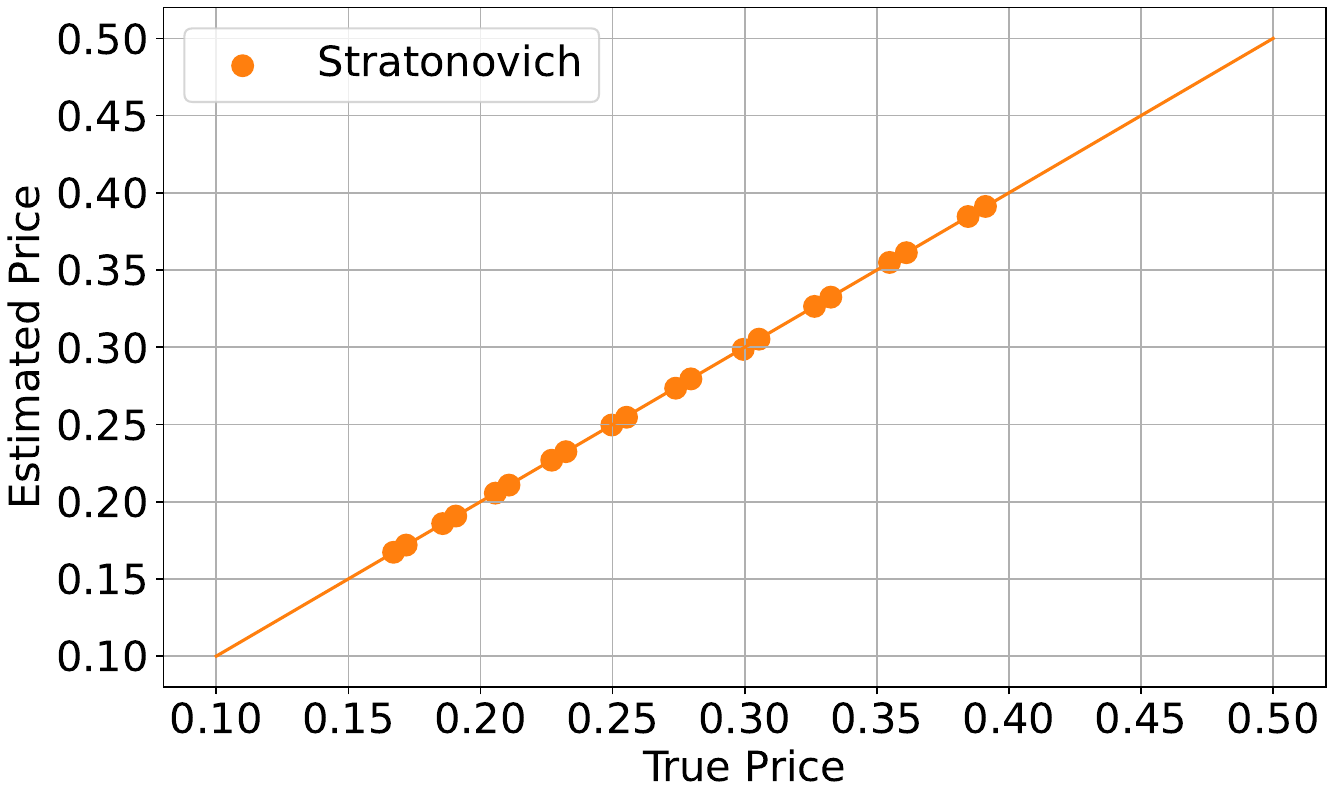}
    }
\end{figure}

\textbf{Results.} Similar to the case of stock options, using signatures with Lasso regression provides an excellent fit for the prices of interest rate options. 
Figure \ref{fig:Scatter_optionpricing_interest} shows the actual and estimated prices of the target options for a randomly chosen experiment out of 100. The actual prices are determined using explicit formulas for caplets and floorlets under the Hull--White model. The prices estimated using both the It\^o and Stratonovich signatures closely align with the actual prices. 

\centerline{[Insert \Cref{fig:Scatter_optionpricing_interest} approximately here.]}

\begin{figure}[htbp]
    \centering
    \caption{Estimation errors for different target interest rate options.\label{fig:Error_band_interest}}
    \subfigure[Call options.\label{fig:Error_band_interest_call}]{
        \includegraphics[width=0.45\linewidth]{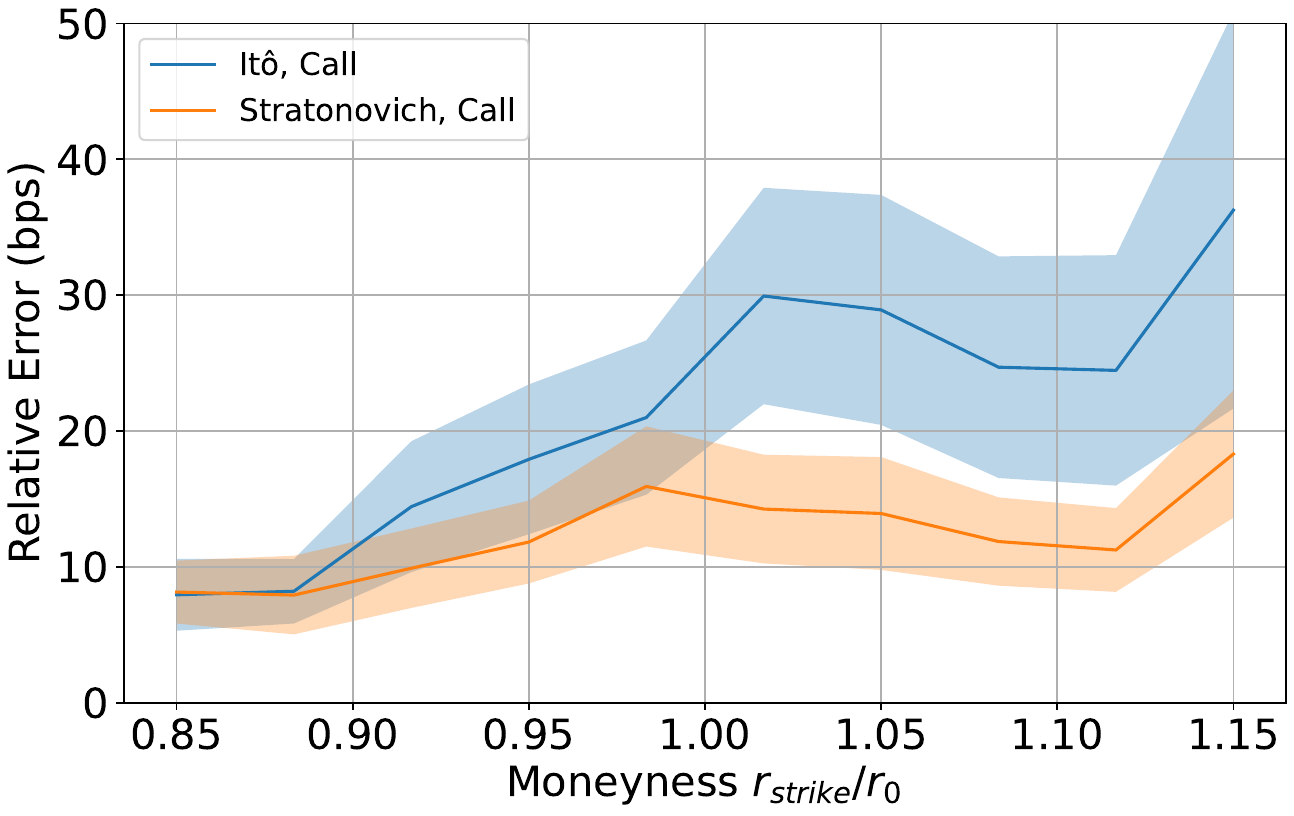}
    }
    % \hfill
    \subfigure[Put options.\label{fig:Error_band_interest_put}]{
        \includegraphics[width=0.45\linewidth]{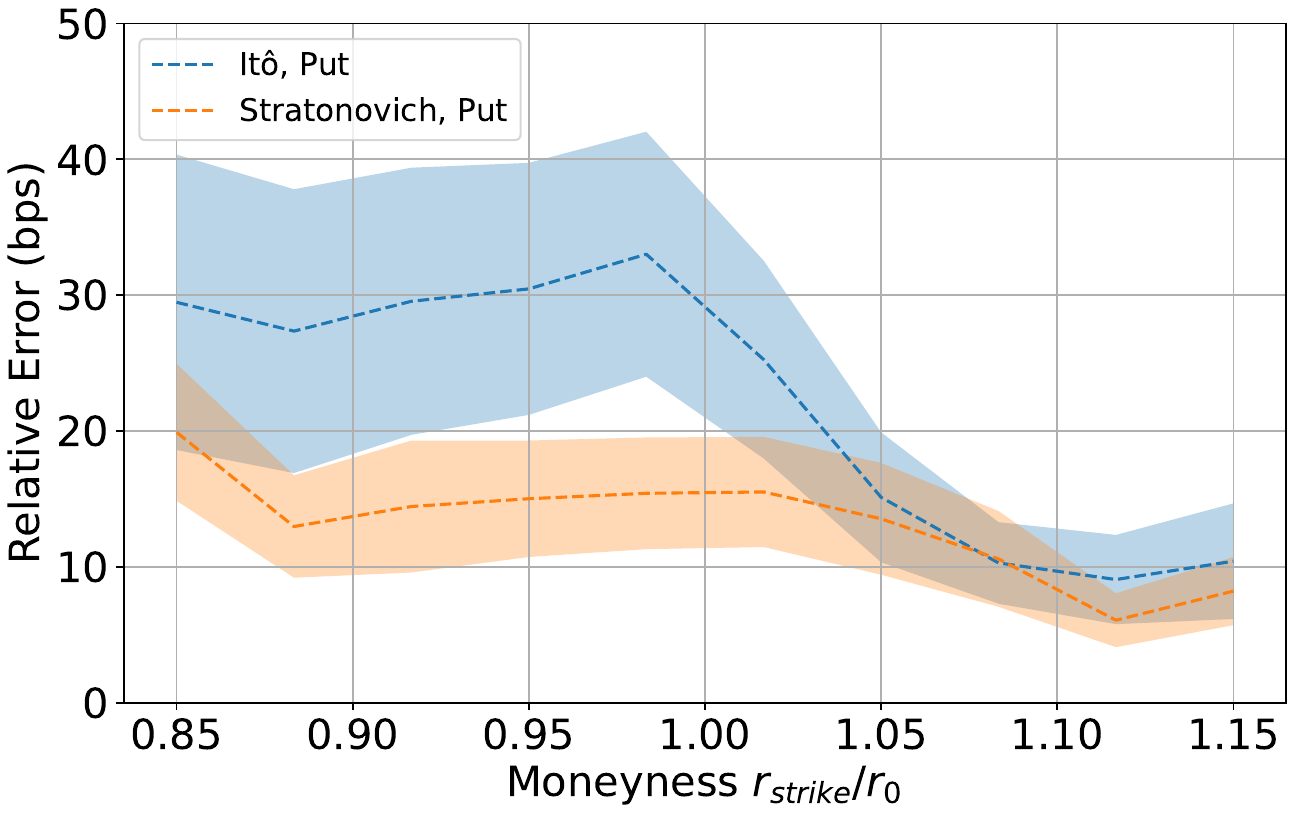}
    }
\end{figure}

Furthermore, in contrast to the case of stock options, the estimation error of interest rate option prices when using the It\^o signature is higher compared to the Stratonovich signature, as shown in Figure \ref{fig:Error_band_interest}. This is also consistent with our results in \Cref{sec:simulation_mean_revert}.  

\centerline{[Insert \Cref{fig:Error_band_interest} approximately here.]}

Overall, our results demonstrate that the Lasso regression with signature is effective in learning nonlinear payoff functions. In addition, the statistical properties of different types of signatures suggest that the It\^o signature is more appropriate if the underlying asset resembles a Brownian motion, and the Stratonovich signature is better if the underlying asset resembles a mean-reverting process.

\section{Conclusion}
\label{sec:conclusion}
This paper studies the statistical consistency of Lasso regression with signatures. 
We first establish a probabilistic uniqueness of the universal nonlinearity, which implies that any feature selection procedure needs to recover this unique linear combination of signature to achieve good predictive performance.

We find that consistency is highly dependent on the definition of the signature, the characteristics of the underlying processes, and the correlation between different dimensions of the underlying process. In particular, the It\^o signature performs better when the underlying process is closer to the Brownian motion and has weaker inter-dimensional correlations, while the Stratonovich signature performs better when the process is sufficiently mean reverting.

The signature method offers an attractive interpretable framework for machine learning and pattern recognition. In fact, the first two orders of signature components correspond to the L\'evy area of sample paths \citep{chevyrev2016primer,levin2016learning}.
In addition, the fact that the target variable can be represented as a linear function of signature components allows for interpretability with respect to the underlying features, and we offer an example in the context of option pricing (\Cref{subsec:optionpricing}).

In general, these results highlight the importance of choosing the appropriate signature for different underlying data, in terms of both learning the right coefficients for interpretation and achieving predictive performance.

Our findings {also} call for further studies on the statistical properties of the signature before its potential in machine learning can be fully realized. 
{ First, in addition to signature, logsignature is also a widely used transform of the path of a stochastic process, which has been shown empirically to improve the training efficiency with simpler and less redundant information of the path \citep{morrill2020generalised,morrill2021neural}. 
However, logsignature does not enjoy the universal nonlinearity \citep{morrill2020generalised,lyons2022signature}, which implies that there is no theoretical basis for using a linear combination of logsignature to approximate a nonlinear function. Therefore, we choose to focus on signature in this study and defer the statistical properties of logsignature to future work.}

{ 
Second, our results highlight the differences in the statistical performance of It\^o and Stratonovich signatures under different probabilistic models of the underlying path. This raises a natural question: is it possible to construct an intermediate signature transform between It\^o and Stratonovich signatures? 
The It\^o and Stratonovich integrals are defined using the left endpoints and midpoints of partition subintervals, respectively. Therefore, one may also consider a class of other stochastic integrals using other points within the subintervals \citep{karatzas1998brownian}. The location of these points may serve as a tuning parameter, which can be selected in practice using techniques such as cross-validation.\footnote{ Specifically, given a tuning parameter $\tau\in[0,1]$, one may define the stochastic integral $\int_0^T A_t \mathrm{d} B_t$ as the limit of $\sum_{j=0}^{n-1}  \left(\tau\cdot A_{t_j} +(1-\tau)\cdot A_{t_{j+1}} \right)  (B_{t_{j+1}} - B_{t_{j}}) $ with $0 = t_0 < t_1 <\dots<t_n=T$ as a partition of $[0,T]$. This reduces to the It\^o integral and Stratonovich integral when $\tau=1$ and $\tau=0.5$, respectively \citep{karatzas1998brownian}. The tuning parameter $\tau$ can be chosen through model selection techniques such as cross-validation.} This approach may balance the statistical advantages of It\^o and Stratonovich signatures. However, further investigation is needed to determine whether the signatures defined by these new types of integrals satisfy universal nonlinearity.
}

{ 
Finally, the theoretical analysis in this study focuses on the statistical consistency of the parameters of each signature component with respect to the number of sample paths, using signatures computed from continuous paths. In practice, the computation of each signature component relies on a discrete sample of the continuous path, which may introduce additional errors. The implications of this discretization on the statistical performance of signature are left for future study.
}

\newpage
% \vspace{3em}
\clearpage
\section*{Tables}
\addcontentsline{toc}{section}{Tables}

% \begin{table}[htbp]
% \centering
% \caption{Outline of main theoretical results.}
% \begin{tabular}{l|c|c|c|c}
% \toprule
% \multicolumn{1}{c|}{Process}    & \multicolumn{2}{c|}{Brownian Motion}                              & \multicolumn{2}{c}{OU Process}          \\ \midrule
% \multicolumn{1}{c|}{Signature}  & \multicolumn{1}{c|}{It\^o }      & \multicolumn{1}{c|}{Stratonovich} & \multicolumn{1}{c|}{It\^o} & Stratonovich \\ \midrule
% Uniqueness of the Universal Nonlinearity (Section \ref{subsec:uniqueness}) & \multicolumn{4}{c}{Theorem \ref{thm:uniqueness}}                                                                               \\ \hhline{~----}
% Correlation Structure  (Section \ref{sec: correlationstructure})  & \multicolumn{1}{c|}{Theorem \ref{prop: structure_coef}} & \multicolumn{3}{c}{Theorem \ref{prop: structure_coef_S_integral}}                                                \\ 
% Consistency, $N=\infty$ (Section \ref{sec: consistency})    & \multicolumn{1}{c|}{Theorem \ref{prop: sufficient_irrepresentable}} & \multicolumn{3}{c}{Theorem \ref{prop: irrepresentable_equiv}}                                                \\
% Consistency, $N<\infty$ (Section \ref{sec: consistency})       & \multicolumn{1}{c|}{Theorem  \ref{th:sampleito}} & \multicolumn{3}{c}{Theorem \ref{th:samplestratonovich}}                                                \\ \bottomrule
% \end{tabular}
% \label{tab:overview}
% \end{table}

\begin{table}[htbp]
\centering
\caption{Methods for computing signature for a discrete time series, $\{\mathbf{X}_{t_j}\}$.} \label{tab:threemethods}
\begin{tabular}{cc}
\toprule
\textbf{Method}          & \textbf{Formula} \\\midrule
It\^o             &   $S(\mathbf{X})_{t_n}^{i_1,\dots,i_k,I} = \sum_{j=0}^{n-1} S(\mathbf{X})_{t_j}^{i_1,\dots,i_{k-1},I} (X_{t_{j+1}}^{i_k} - X_{t_{j}}^{i_k}) $      \\
Stratonovich    &   $S(\mathbf{X})_{t_n}^{i_1,\dots,i_k,S} =  \sum_{j=0}^{n-1} \frac{1}{2} \left( S(\mathbf{X})_{t_j}^{i_1,\dots,i_{k-1},S}+S(\mathbf{X})_{t_{j+1}}^{i_1,\dots,i_{k-1},S} \right)  (X_{t_{j+1}}^{i_k} - X_{t_{j}}^{i_k}) $      \\
Linear &  Linearly interpolate $\{\mathbf{X}_{t_j}\}$ and compute signature using Riemann/Lebesgue integral     \\\bottomrule
\end{tabular}
\end{table}

\clearpage
\section*{Figures}
\addcontentsline{toc}{section}{Figures}

\begin{figure}[htbp]
    \centering
    \caption{Consistency rates for the Brownian motion and the random walk with different values of inter-dimensional correlation $\rho$ and different numbers of true predictors $q$. Solid (dashed) lines correspond to the It\^o (Stratonovich) signature.}
    \label{fig: Experiment_3}
    \subfigure[Brownian motion.]{
        \includegraphics[width=0.45\linewidth]{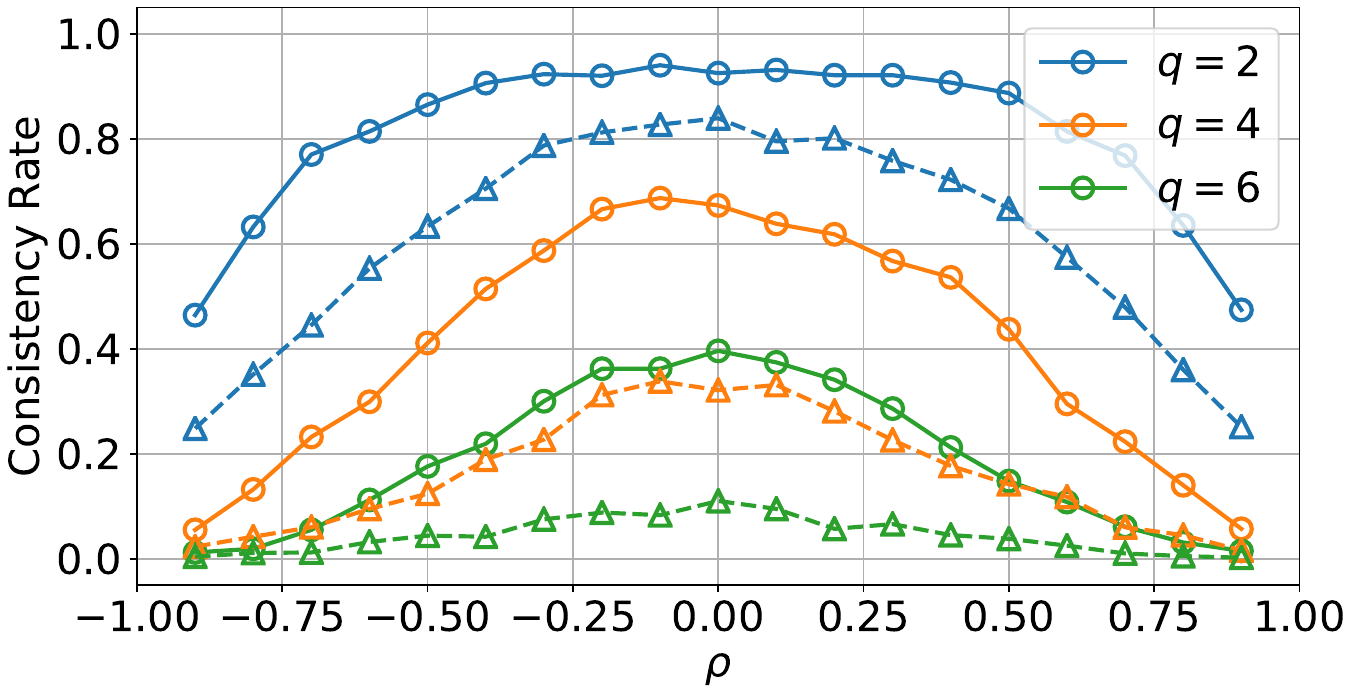}
        \label{fig: Experiment_3_BM}
    }
    % \hfill
    \subfigure[Random walk.]{
        \includegraphics[width=0.45\linewidth]{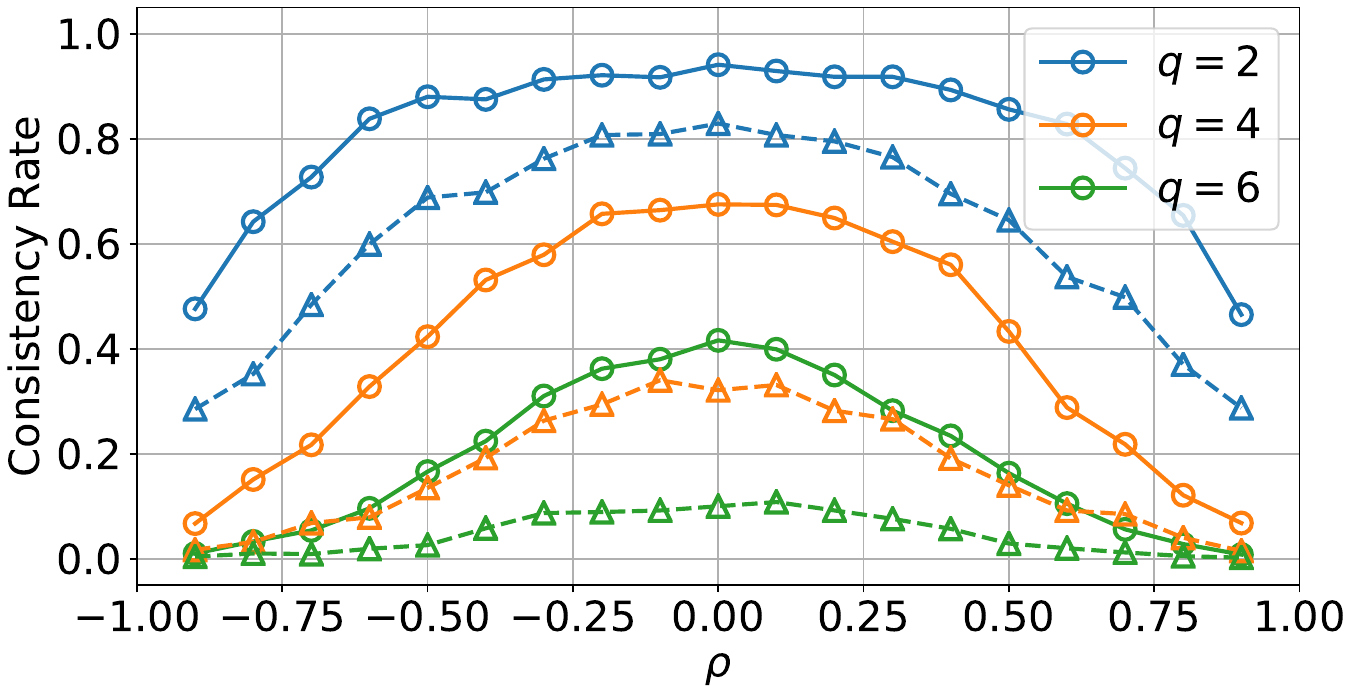}
        \label{fig: Experiment_3_RW}
    }
   
\end{figure}

\begin{figure}[htbp]
    \centering
    \caption{In-sample and out-of-sample $R^2$ for learning option payoffs using different types of predictors.\label{fig:r2total}}
    \subfigure[Call option.\label{fig:call_r2}]{
        \includegraphics[width=0.37\textwidth]{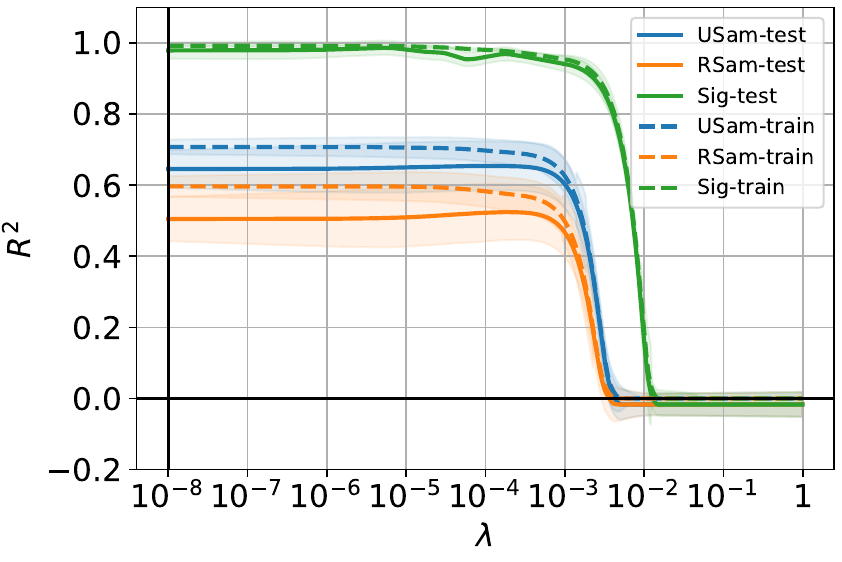}
    }
    % \hfill
    \subfigure[Put option.\label{fig:put_r2}]{
        \includegraphics[width=0.37\textwidth]{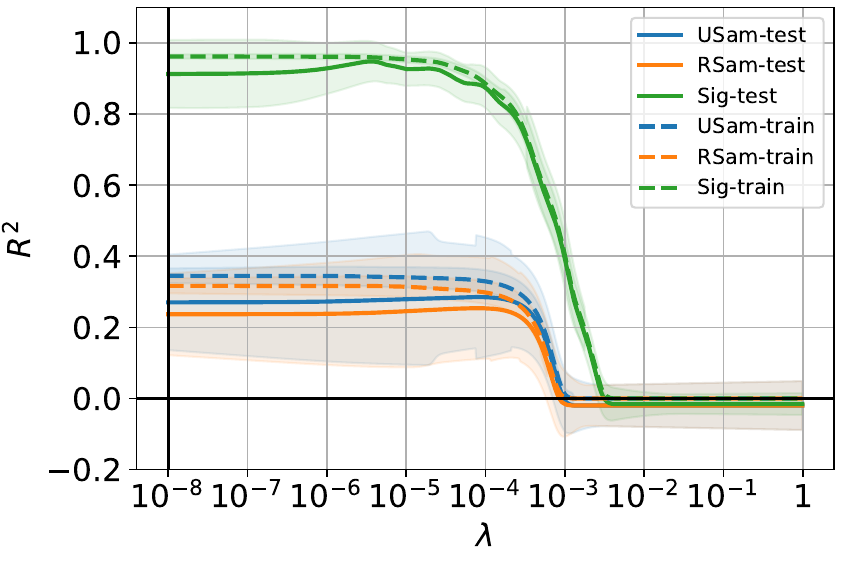}
    }
    % \hfill
    \subfigure[Asian option.\label{fig:asian_r2}]{
        \includegraphics[width=0.37\textwidth]{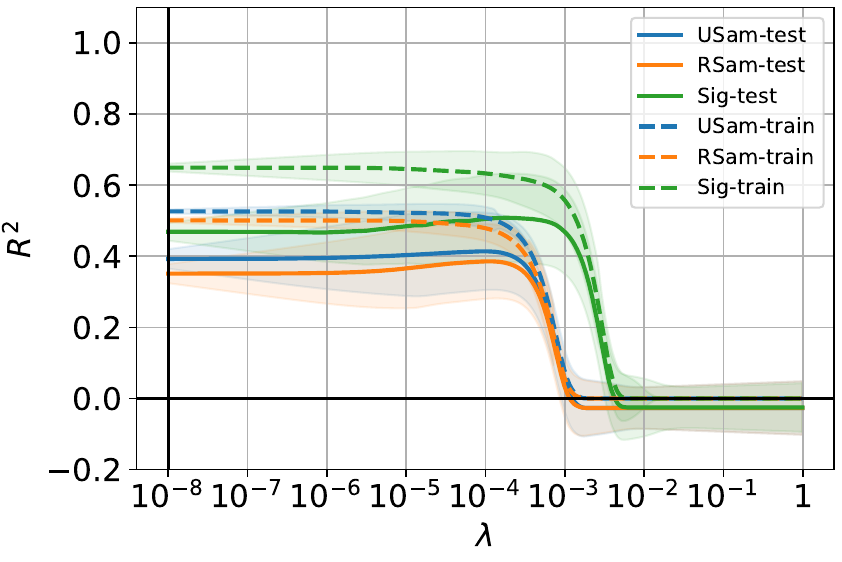}
    }
    % \hfill
    \subfigure[Lookback option.\label{fig:lookback_r2}]{
        \includegraphics[width=0.37\textwidth]{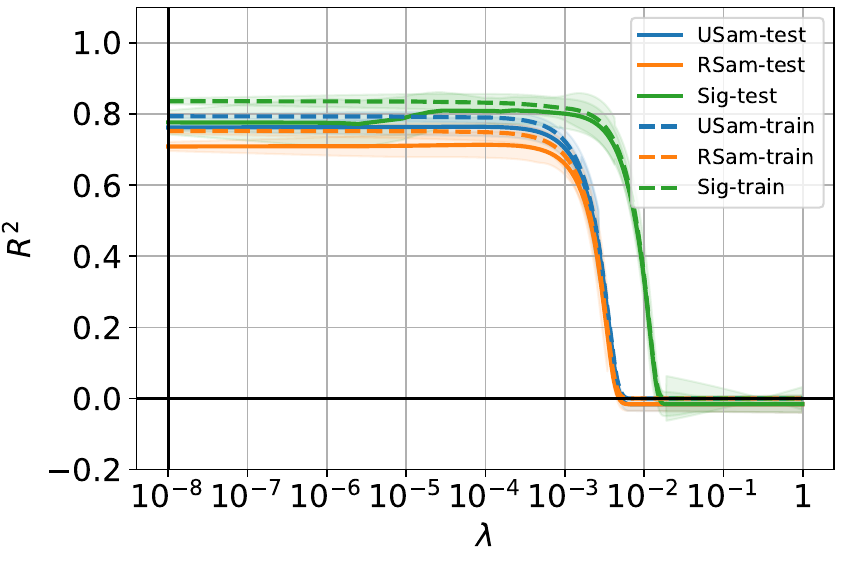}
    }
    % \hfill
    \subfigure[Rainbow option \Rmnum{1}.\label{fig:rainbow1_r2}]{
        \includegraphics[width=0.37\textwidth]{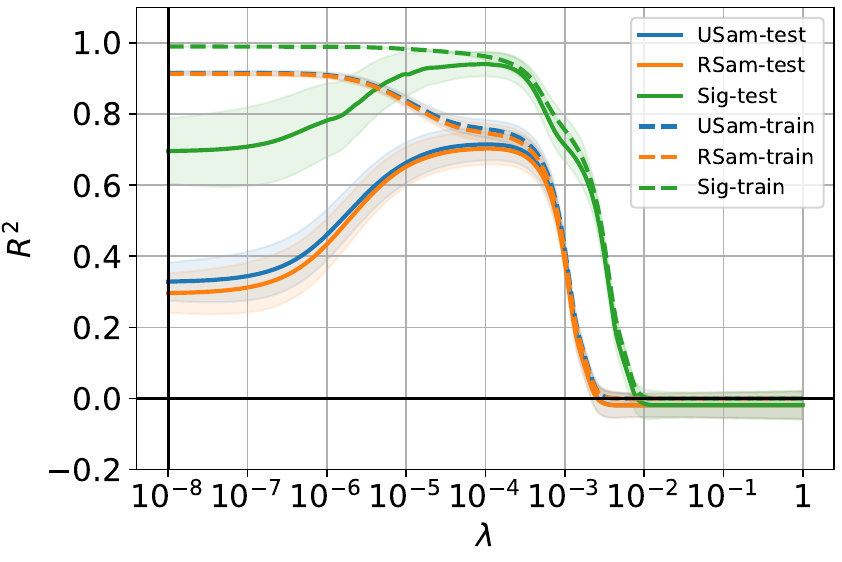}
    }
    % \hfill
    \subfigure[Rainbow option \Rmnum{2}.\label{fig:rainbow2_r2}]{
        \includegraphics[width=0.37\textwidth]{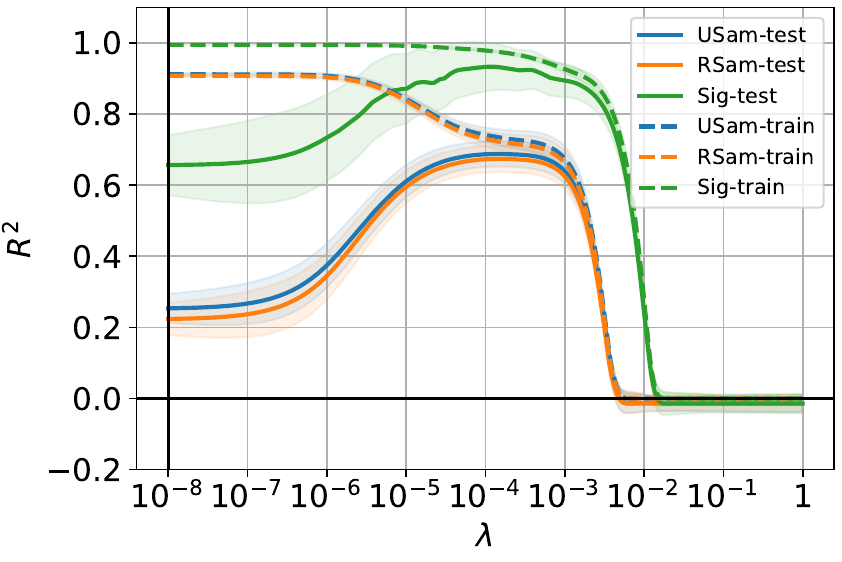}
    }
    % \hfill
    \subfigure[ Rainbow option \Rmnum{3}.\label{fig:rainbow3_r2}]{
        \includegraphics[width=0.37\textwidth]{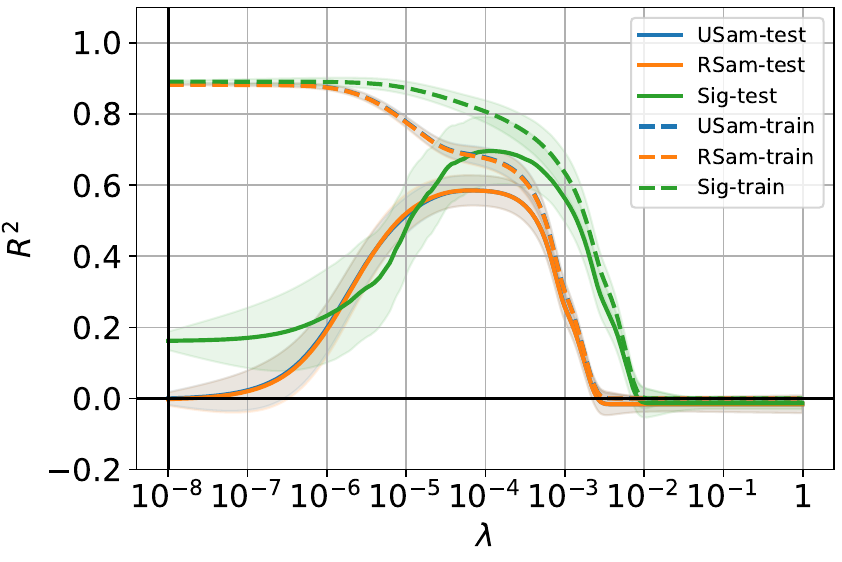}
    }
    % \hfill
    \subfigure[ Rainbow option \Rmnum{4}.\label{fig:rainbow4_r2}]{
        \includegraphics[width=0.37\textwidth]{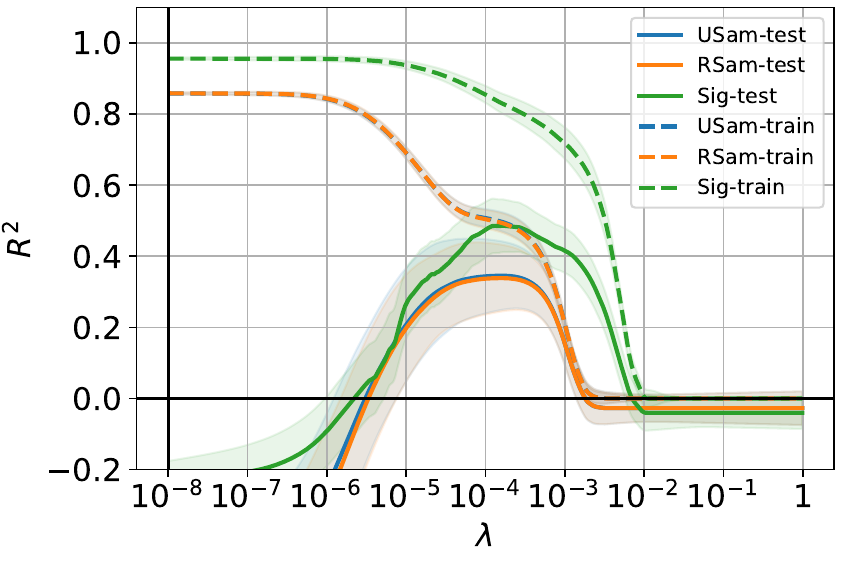}
    }
\end{figure}

\clearpage

\newpage

\counterwithin{figure}{section}
\makeatletter
\renewcommand\p@subfigure{\thefigure}
\makeatother
\counterwithin{equation}{section}
\counterwithin{table}{section}
\counterwithin{theorem}{section}
\counterwithin{proposition}{section}
\counterwithin{definition}{section}
\counterwithin{example}{section}
\counterwithin{lemma}{section}
\counterwithin{remark}{section}
\counterwithin{assumption}{section}
\appendix

\section*{\centering Electronic Companion}
\addcontentsline{toc}{section}{Electronic Companion}

\section{Impact of Time Augmentation}\label{appendix:timeaug}
First, recall that the time-augmented process of a $d$-dimensional continuous-time stochastic process $\mathbf{X}_t = (X_t^1, X_t^2,\dots, X_t^d)^\top \in \mathbb{R}^d$, $0\leq t \leq T$ is a $(d+1)$-dimensional process \citep{chevyrev2016primer,lyons2022signature}
    \begin{equation} \label{equ: time_augmented}
        \tilde{\mathbf{X}}_t = \begin{pmatrix}
      t , {\mathbf{X}}_t^\top 
        \end{pmatrix}^\top = \begin{pmatrix}
       t , X_t^1 , X_t^2 , \dots , X_t^d 
        \end{pmatrix}^\top.
    \end{equation}
The time augmentation does not change the core block-diagonal structure between signature components.

In particular, for a $d$-dimensional Brownian motion $\mathbf{X}$ given by \eqref{equ: MultiBM_setup}, if all signature components with the time dimension are grouped together, with other signature components arranged in recursive order (see Definition \ref{def: recursive order} in Appendix \ref{appendix: calculation_of_corr}), the correlation matrix for It\^o signature of $\tilde{\mathbf{X}}$ with orders truncated to $K$ is given by
\begin{equation*}
\begin{pmatrix}
    \Psi_{0,0} & \Psi_{0,1} & \Psi_{0,2} & \cdots & \Psi_{0,K}  \\
    \Psi_{1,0} & \Omega_{1} & 0 & \cdots & 0\\
    \Psi_{2,0} & 0 & \Omega_{2} &  \cdots & 0 \\
    \vdots & \vdots & \cdots & \ddots & \vdots  \\
    \Psi_{K,0} & 0 & 0 & \cdots & \Omega_K
    \end{pmatrix},
    \end{equation*}
with $\Omega_i$ defined by \eqref{equ: Omega_k} and $\Psi_{0,m}$ the correlation matrix between all signature components with the time dimension and all $m$-th order signature components without the time dimension. 

Similarly, for the Stratonovich signature of a $(d+1)$-dimensional time-augmented Brownian motion given by \eqref{equ: MultiBM_setup} and \eqref{equ: time_augmented}, or the It\^o or Stratonovich signature of a $(d+1)$-dimensional time-augmented OU process given by \eqref{equ: OU_setup} and \eqref{equ: time_augmented}, if we group all signature components with the time dimension together and other signature components together, the correlation matrix for the signature with orders truncated to $K$ can be given by
\begin{equation} \label{equ:structure_time_aug}
\begin{pmatrix}
    \Psi_{0,0} & \Psi_{0,\mathrm{odd}} & \Psi_{0,\mathrm{even}}\\
    \Psi_{\mathrm{odd},0} & \Psi_{\mathrm{odd}} & 0 \\
    \Psi_{\mathrm{even},0} & 0 & \Psi_{\mathrm{even}} 
    \end{pmatrix},
    \end{equation}
where $\Psi_{\mathrm{odd}}$ and $\Psi_{\mathrm{even}}$ are defined by \eqref{equ: def_odd_even_matrix}, $\Psi_{0,0}$ is the correlation matrix between all signature components with the time dimension, and $\Psi_{0,\mathrm{odd}}$ ($\Psi_{0,\mathrm{even}}$) is the correlation matrix between 
all signature components with the time dimension and all odd (even) order signature components without the time dimension. 

\textbf{Simulation.}
Now we perform simulations to study the consistency of signature using Lasso regression for the time-augmented Brownian motion. 
We consider $\tilde{\textbf{X}}$, the time-augmentation of a 2-dimensional Brownian motion with an inter-dimensional correlation of $\rho$. The simulation setups are the same as in \Cref{sec:simulation}.

\begin{figure}[htbp]
    \centering
    \caption{Consistency rates for the time-augmented Brownian motion with different values of inter-dimensional correlation $\rho$ and different numbers of true predictors $q$. Solid (dashed) lines correspond to the It\^o (Stratonovich) signature. \label{fig: Experiment_9}}
    \includegraphics[width=0.48\textwidth]{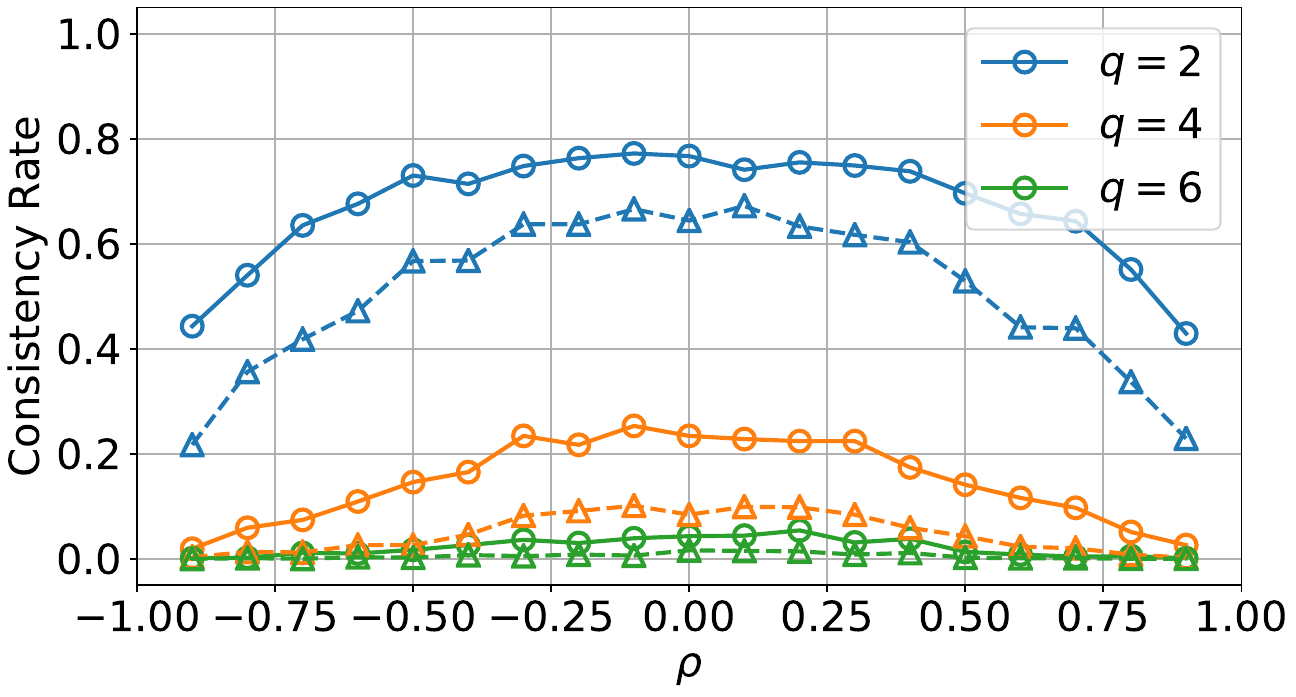}
    \end{figure}

Figure \ref{fig: Experiment_9} shows the consistency rates for different values of inter-dimensional correlation $\rho$, and different numbers of true predictors $q$. The time augmentation generally increases the correlation between signature components and, therefore, leads to a lower consistency rate for Lasso compared to the case without time augmentation (\Cref{fig: Experiment_3_BM}). However, the main relationships of the consistency rate with respect to $\rho$ and $q$ remain the same.

\begin{figure}[htbp]
    \centering
    \caption{ In-sample and out-of-sample $R^2$ for learning option payoffs using different types of predictors with time augmentation.\label{fig:timeaugr2total}}
    \subfigure[Call option.\label{fig:revision_call_r2}]{
        \includegraphics[width=0.37\textwidth]{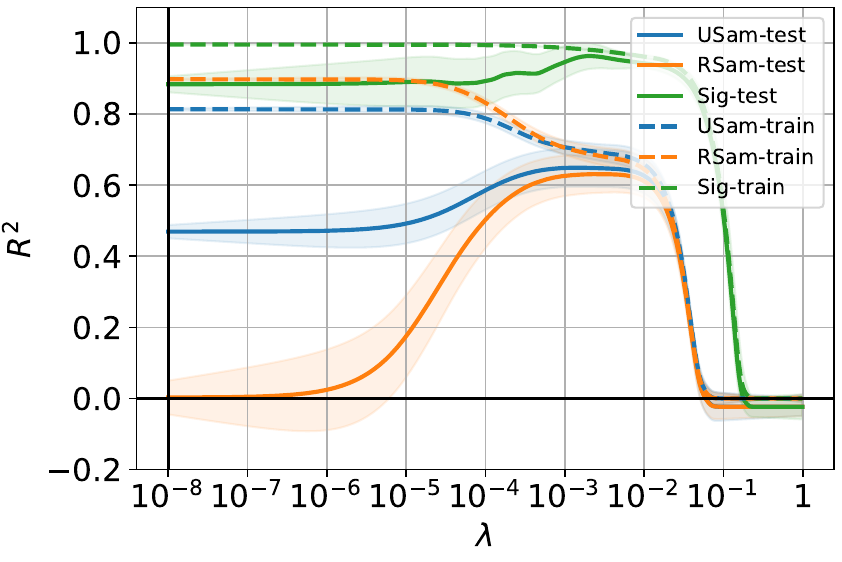}
    }
    \hfill
    \subfigure[Put option.\label{fig:revision_put_r2}]{
        \includegraphics[width=0.37\textwidth]{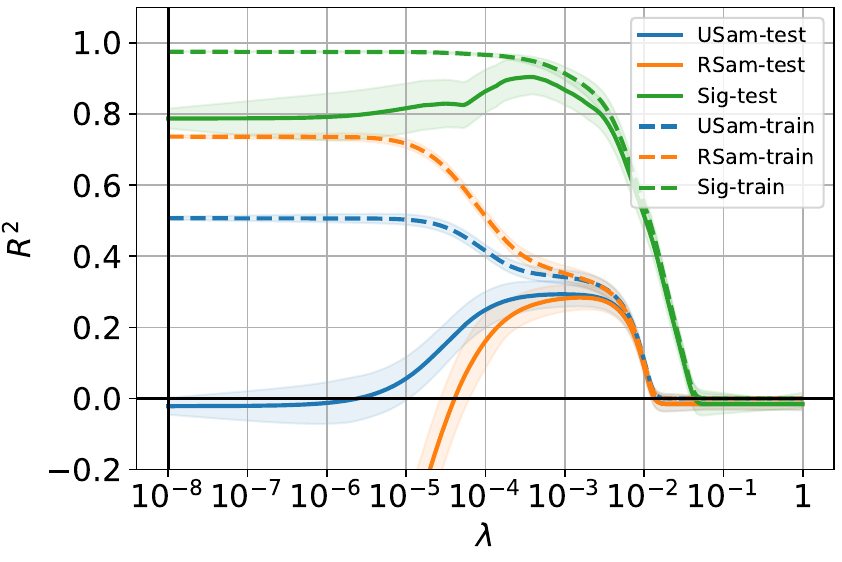}
    }
    \hfill
    \subfigure[Asian option.\label{fig:revision_asian_r2}]{
        \includegraphics[width=0.37\textwidth]{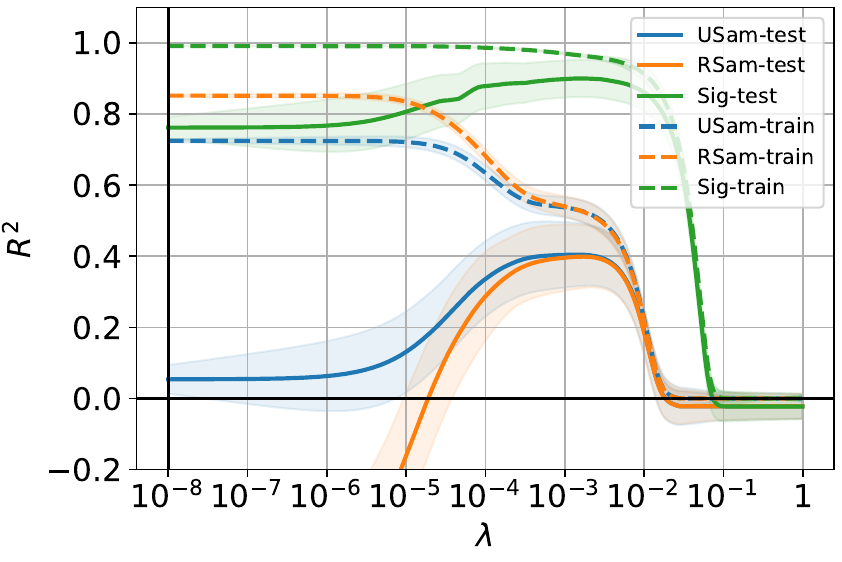}
    }
    \hfill
    \subfigure[Lookback option.\label{fig:revision_lookback_r2}]{
        \includegraphics[width=0.37\textwidth]{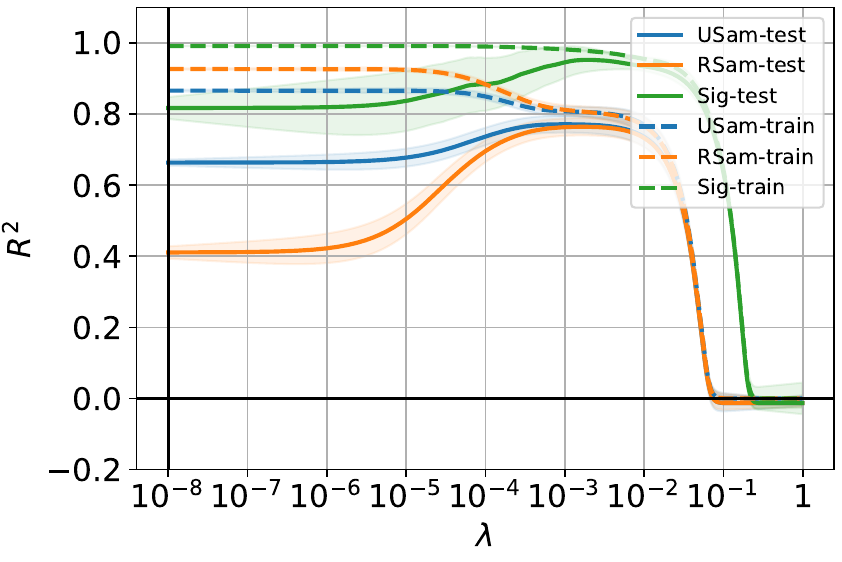}
    }
    \hfill
    \subfigure[Rainbow option \Rmnum{1}.\label{fig:revision_rainbow1_r2}]{
        \includegraphics[width=0.37\textwidth]{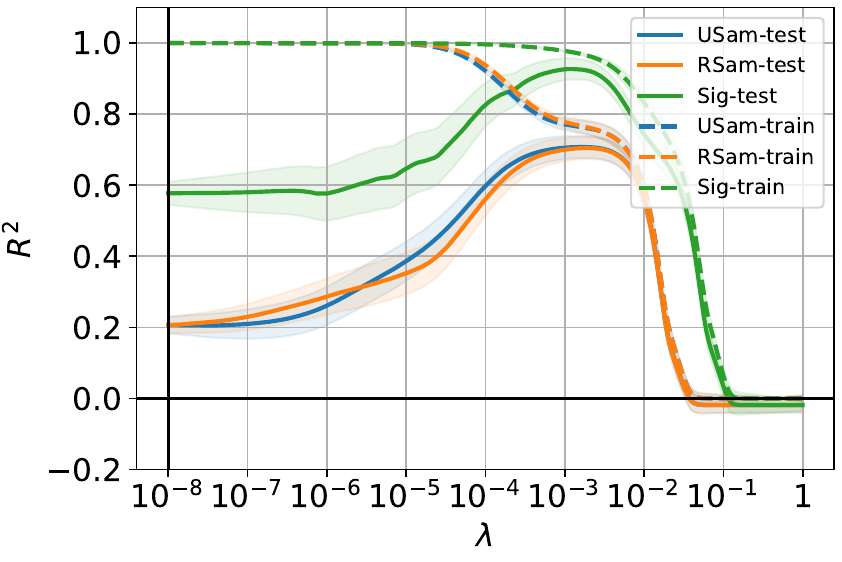}
    }
    \hfill
    \subfigure[Rainbow option \Rmnum{2}.\label{fig:revision_rainbow2_r2}]{
        \includegraphics[width=0.37\textwidth]{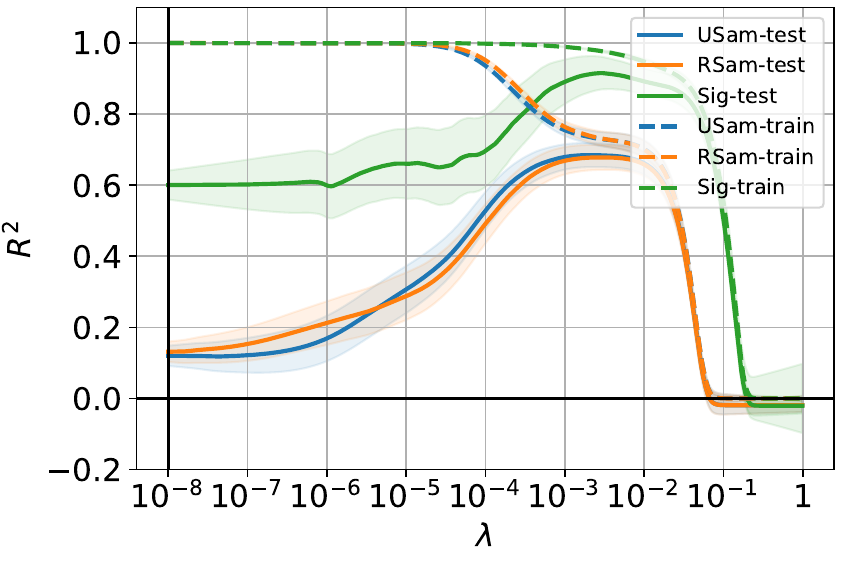}
    }
    \hfill
    \subfigure[Rainbow option \Rmnum{3}.\label{fig:revision_rainbow3_r2}]{
        \includegraphics[width=0.37\textwidth]{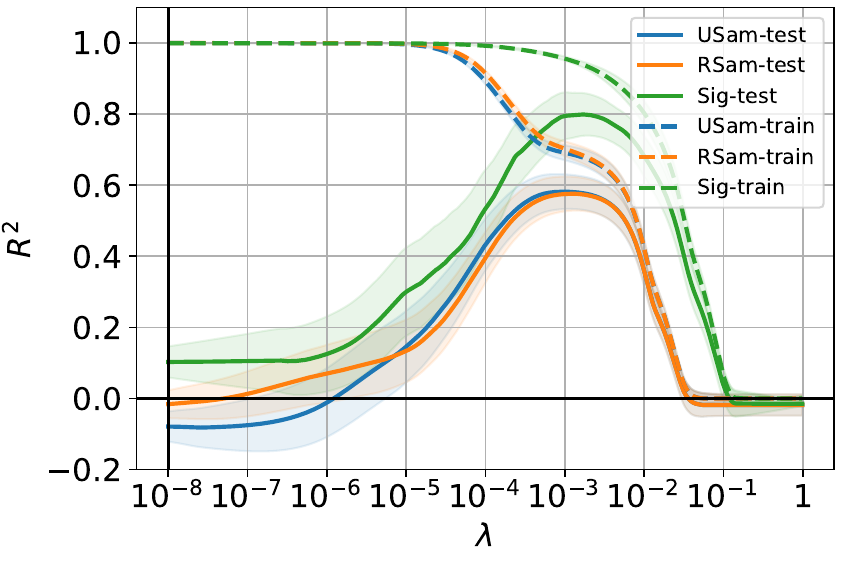}
    }
    \hfill
    \subfigure[Rainbow option \Rmnum{4}.\label{fig:revision_rainbow4_r2}]{
        \includegraphics[width=0.37\textwidth]{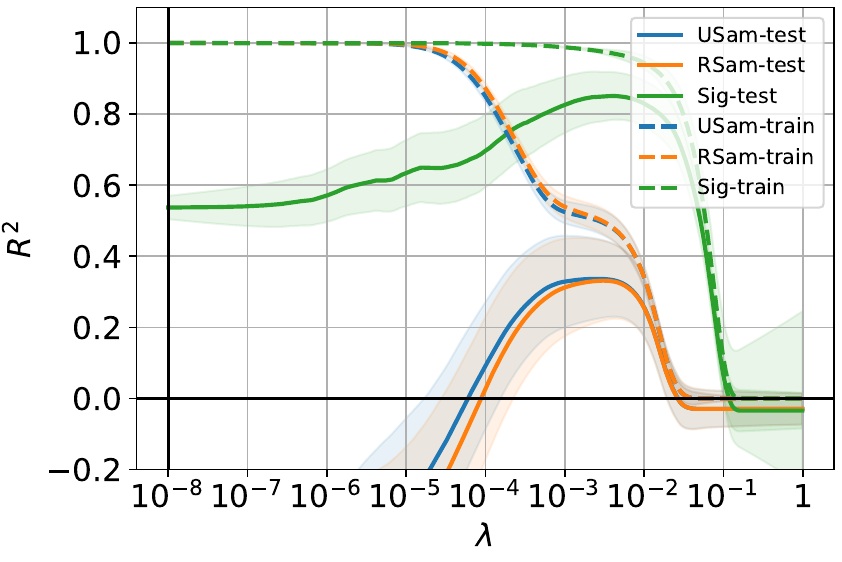}
    }
\end{figure}

{\textbf{Learning option payoffs.} We also demonstrate the ability of signature to learn option payoffs when incorporating time augmentation. Following our framework in Section \ref{subsubsec:olsandsig}, we consider two underlying assets, eight different option payoff functions, and three different types of predictors (Sig, RSam, and USam). The only difference in this section is that we also include the time dimension when calculating these three types of predictors.

Figure \ref{fig:timeaugr2total} shows $R^2$ as a function of the penalization parameter of the Lasso regression $\lambda$, when using different types of predictors with time augmentation. Similar to our observations without time augmentation (Figure \ref{fig:r2total}), both in-sample and out-of-sample $R^2$ values for Lasso regression with signature components as predictors consistently outperform those for Lasso regression with random sampling and equidistant sampling as predictors.

By comparing Figure \ref{fig:timeaugr2total} and Figure \ref{fig:r2total}, we also find that $R^2$ values using signature with time augmentation outperform those without time augmentation, particularly for path-dependent options. This demonstrates that, although the signature of time-augmentation paths has a lower consistency rate due to the inclusion of more predictors, it is more effective in approximating various nonlinear payoff functions, thanks to the universal nonlinearity (Theorem~\ref{th:UN}). 
}

\section{Technical Details and Examples for the Calculation of Correlation Structures} \label{appendix: calculation_of_corr}

This appendix provides details and examples for calculating the correlation structures of signature. Appendices \ref{appendix: calculation_of_corr_BM} and \ref{appendix: calculation_of_corr_OU} discuss the Brownian motion and the OU process, respectively. 

\subsection{Brownian Motion} \label{appendix: calculation_of_corr_BM}

\paragraph{It\^o Signature.} Proposition \ref{prop: moment_signatures} and Theorem \ref{prop: structure_coef} in the main paper give explicit formulas for calculating the correlation structure of the It\^o signature for Brownian motion. The ``recursive order'' mentioned in Theorem~\ref{prop: structure_coef} is defined as follows. 

\begin{definition}[Recursive Order] \label{def: recursive order}
Consider a $d$-dimensional process $\mathbf{X}$. We order the indices of all of its 1st order signature components as
\begin{equation*}
    1 \quad 2 \quad \cdots \quad d.
\end{equation*}
Then, if all $k$-th order signature components are ordered as
\begin{equation*}
    r_1 \quad r_2 \quad \cdots \quad r_{d^k},
\end{equation*}
we define the orders of all $(k+1)$-th order signature components as
\begin{equation*}
    r_1,1 \quad r_2,1 \quad \cdots \quad r_{d^k},1 \quad r_1,2 \quad r_2,2 \quad \cdots \quad r_{d^k},2 \quad \cdots \cdots  \cdots \quad r_1,d \quad r_2,d \quad \cdots \quad r_{d^k},d. 
\end{equation*}
\end{definition}
For example, for a $d=3$-dimensional process, the recursive order of its signature is
\begin{itemize}
    \item 1st order: $1\quad 2 \quad 3$
    \item 2nd order: $1,1 \quad 2,1 \quad 3,1 \quad 1,2 \quad 2,2 \quad 3,2 \quad 1,3 \quad 2,3 \quad 3,3$
    \item 3rd order: $1,1,1 \quad 2,1,1 \quad 3,1,1 \quad 1,2,1 \quad 2,2,1 \quad 3,2,1 \quad 1,3,1 \quad 2,3,1 \quad 3,3,1 $ \\$ ~~~~~~~~~~~~~~~~~~~~~~~~~~1,1,2 \quad 2,1,2 \quad 3,1,2 \quad 1,2,2 \quad 2,2,2 \quad 3,2,2 \quad 1,3,2 \quad 2,3,2 \quad 3,3,2 $ \\$ ~~~~~~~~~~~~~~~~~~~~~~~~~~1,1,3 \quad 2,1,3 \quad 3,1,3 \quad 1,2,3 \quad 2,2,3 \quad 3,2,3 \quad 1,3,3 \quad 2,3,3 \quad 3,3,3 $
    \item ...
\end{itemize}

To provide intuition for Proposition \ref{prop: moment_signatures} and Theorem \ref{prop: structure_coef} in the main paper, the following two examples show the correlation structures of It\^o signatures for 2-dimensional Brownian motions with inter-dimensional correlations $\rho = 0.6$ and $\rho = 0$, respectively.

\begin{figure}[htbp]
    \centering
    \caption{Correlation matrices of signatures for 2-dimensional Brownian motions.\label{fig:corr_rho}}
    \subfigure[It\^o; Inter-dimensional correlation $\rho = 0.6$.\label{fig: corr_I_rho_0.6}]{
        \includegraphics[width=0.45\linewidth]{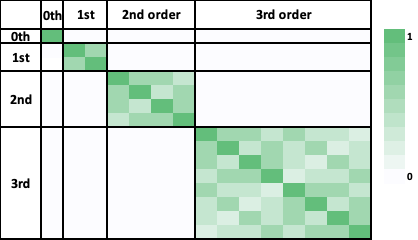}
    }
    \hfill
    \subfigure[It\^o; Inter-dimensional correlation $\rho = 0$.\label{fig: corr_I_rho_0}]{
        \includegraphics[width=0.45\linewidth]{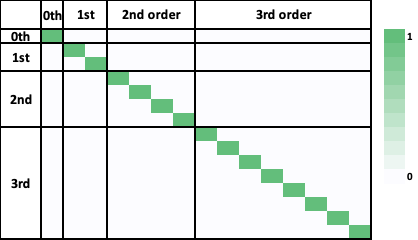}
    }
    \hfill
    \subfigure[Stratonovich; Inter-dimensional correlation $\rho = 0.6$.\label{fig: corr_S_rho_0.6}]{
        \includegraphics[width=0.45\linewidth]{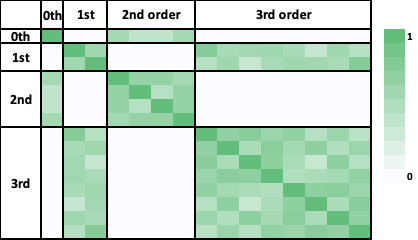}
    }
    \hfill
    \subfigure[Stratonovich; Inter-dimensional correlation $\rho = 0$.\label{fig: corr_S_rho_0}]{
        \includegraphics[width=0.45\linewidth]{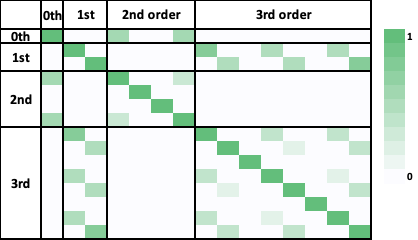}
    }
\end{figure}

\begin{example}\label{exmp: fig: corr_I_rho_0.6}
    Consider a 2-dimensional Brownian motion given by \eqref{equ: MultiBM_setup} with an inter-dimensional correlation of $\rho = 0.6$. Figure \ref{fig: corr_I_rho_0.6} shows the correlation matrix of its It\^o signature calculated using Proposition \ref{prop: moment_signatures}. The figure illustrates Theorem \ref{prop: structure_coef}---the correlation matrix has a block diagonal structure, and each block of the matrix is the Kronecker product of the inter-dimensional correlation matrix $
    \begin{pmatrix}
        1 & 0.6 \\ 0.6 & 1
    \end{pmatrix}
    $. 
\end{example}

\begin{example}\label{exmp: fig: corr_I_rho_0}
    Consider a 2-dimensional Brownian motion given by \eqref{equ: MultiBM_setup} with an inter-dimensional correlation of $\rho = 0$. Figure \ref{fig: corr_I_rho_0} shows the correlation matrix of its It\^o signature calculated using Proposition \ref{prop: moment_signatures}. When $\rho=0$, the block diagonal correlation matrix reduces to an identity matrix, indicating that all of its It\^o signature components 
    are mutually uncorrelated. 
\end{example}

\paragraph{Stratonovich Signature.} Proposition \ref{prop: moment_signatures_S_integral} and Theorem \ref{prop: structure_coef_S_integral} in the main paper provide formulas for calculating the correlation structure of the Stratonovich signature for a Brownian motion. The following proposition gives the concrete recursive formulas for calculating $\mathbb{E} \left[S(\mathbf{X})_{t}^{i_1,\dots,i_{2n},S} S(\mathbf{X})_{t}^{j_1,\dots,j_{2m},S}\right]$ and $\mathbb{E} \left[S(\mathbf{X})_{t}^{i_1,\dots,i_{2n-1},S} S(\mathbf{X})_{t}^{j_1,\dots,j_{2m-1},S}\right]$, which extends Proposition \ref{prop: moment_signatures_S_integral} in the main paper.

\begin{proposition} \label{prop: moment_signatures_S_integral_TEMP}
Let $\mathbf{X}$ be a $d$-dimensional Brownian motion given by \eqref{equ: MultiBM_setup}. For any $l,t\geq 0$ and $m, n \in \mathbb{N}^+$, define $ f_{2n,2m}(l,t) := \mathbb{E} \left[S(\mathbf{X})_{l}^{i_1,\dots,i_{2n},S} S(\mathbf{X})_{t}^{j_1,\dots,j_{2m},S} \right]$, we have
\begin{align}
    f_{2n,2m}(l,t) &= g_{2n,2m}(l,t) + \frac{1}{2} \rho_{j_{2m-1} j_{2m}} \sigma_{j_{2m-1}}\sigma_{j_{2m}} \int_0^t f_{2n,2m-2}(l,s) \mathrm{d}s, \label{equ: propTEMP_1}\\
    g_{2n,2m}(l,t) &= \rho_{i_{2n}j_{2m}}\sigma_{i_{2n}}\sigma_{j_{2m}} \int_0^{l \wedge t} f_{2n-1,2m-1}(s,s) \mathrm{d} s \nonumber\\
    & \qquad\qquad + \frac{1}{2} \rho_{i_{2n-1} i_{2n}} \sigma_{i_{2n-1}}\sigma_{i_{2n}} \int_0^l g_{2n-2,2m}(s,t) \mathrm{d}s,\label{equ: propTEMP_2}
\end{align}
with initial conditions
\begin{align}
    f_{0,0}(l,t) &= 1, \label{equ: propTEMP_ini1} \\
    g_{0,2m}(l,t) &= 0.\label{equ: propTEMP_ini2}
\end{align}
In addition, define $ f_{2n-1,2m-1}(l,t) := \mathbb{E} \left[S(\mathbf{X})_{l}^{i_1,\dots,i_{2n-1},S} S(\mathbf{X})_{t}^{j_1,\dots,j_{2m-1},S} \right]$, we have
\begin{align}
    f_{2n-1,2m-1}(l,t) &= g_{2n-1,2m-1}(l,t) + \frac{1}{2} \rho_{j_{2m-2} j_{2m-1}} \sigma_{j_{2m-2}}\sigma_{j_{2m-1}} \int_0^t f_{2n-1,2m-3}(l,s) \mathrm{d}s, \label{equ: propTEMP_3}\\
    g_{2n-1,2m-1}(l,t) &= \rho_{i_{2n-1}j_{2m-1}}\sigma_{i_{2n-1}}\sigma_{j_{2m-1}} \int_0^{l \wedge t} f_{2n-2,2m-2}(s,s) \mathrm{d} s \nonumber\\
    &\qquad\qquad + \frac{1}{2} \rho_{i_{2n-2} i_{2n-1}} \sigma_{i_{2n-2}}\sigma_{i_{2n-1}} \int_0^l g_{2n-3,2m-1}(s,t) \mathrm{d}s,\label{equ: propTEMP_4}
\end{align}
with initial conditions
\begin{align}
    f_{1,1}(l,t) &= \rho_{i_1j_1} \sigma_{i_1} \sigma_{j_1} (l \wedge t), \label{equ: propTEMP_ini3}\\
    g_{1,2m-1}(l,t) &= \rho_{i_1 j_{2m-1}} \frac{1}{2^{m-1}} \frac{(l \wedge t)^{m-1}}{(m-1)!} \sigma_{i_1} \prod_{k=1}^{2m-1} \sigma_{j_k} \prod_{k=1}^{m-1} \rho_{j_{2k-1}j_{2k}} .\label{equ: propTEMP_ini4}
\end{align}
Here, $x \wedge y$ represents the smaller value between $x$ and $y$.
\end{proposition}

The following two examples show the correlation structures of Stratonovich signatures for 2-dimensional Brownian motions with inter-dimensional correlations $\rho = 0.6$ and $\rho = 0$, respectively, calculated using Proposition \ref{prop: moment_signatures_S_integral} and Theorem \ref{prop: structure_coef_S_integral} in the main paper and Proposition \ref{prop: moment_signatures_S_integral_TEMP}.

\begin{example} \label{exmp: fig: corr_S_rho_0.6}
    Consider a 2-dimensional Brownian motion given by \eqref{equ: MultiBM_setup} with an inter-dimensional correlation of $\rho = 0.6$. Figure \ref{fig: corr_S_rho_0.6} shows the correlation matrix of its Stratonovich signature calculated using Propositions \ref{prop: moment_signatures_S_integral} and \ref{prop: moment_signatures_S_integral_TEMP}. 
    The figure illustrates that the correlation matrix has an odd--even alternating structure. 
\end{example}

\begin{example} \label{exmp: fig: corr_S_rho_0}
    Consider a 2-dimensional Brownian motion given by \eqref{equ: MultiBM_setup} with an inter-dimensional correlation of $\rho = 0$. Figure \ref{fig: corr_S_rho_0} shows the correlation matrix of its Stratonovich signature calculated using Propositions \ref{prop: moment_signatures_S_integral} and \ref{prop: moment_signatures_S_integral_TEMP}. 
    The figure demonstrates that the correlation matrix has an odd--even alternating structure, even though different dimensions of the Brownian motion are mutually independent ($\rho=0$). This is different from the result for It\^o signature shown in Example \ref{exmp: fig: corr_I_rho_0}, where all It\^o signature are mutually uncorrelated. 

    In this case, assume that one includes all Stratonovich signature components of orders up to $K=4$ in the Lasso regression given by \eqref{equ: lasso}, and the true model given by \eqref{equ: linearregression} has beta coefficients $    \beta_0 =0$, $    \beta_1 >0$,  $\beta_2 > 0$,  $\beta_{1,1} > 0$, $\beta_{1,2} >0 $, $\beta_{2,1} > 0$, $\beta_{2,2} < 0$, and $\beta_{i_1,i_2,i_3} = \beta_{i_1,i_2,i_3,i_4}= 0$. Let $\Delta^2$ be the correlation matrix between all predictors given by Theorem \ref{prop: structure_coef_S_integral}. 
Then, by Proposition \ref{prop: moment_signatures_S_integral},
\begin{align*}
    {\Delta}_{A^{*c},A^*}^{2} ({\Delta}_{A^*,A^*}^{2})^{-1} \mathrm{sign}( {\boldsymbol \beta}_{A^*} ) &= (0, 0.77,  0.5 , 0, 0.5, 0.5, 0, 0.5, 0.77, 1.01, 0.73, 0.47, 0, \\&~~~~0.47, 0, 0.58, 0.73, 0.73, -0.58, 0, 0.47, 0, 0.47, 0.73, -1.01)^\top,
\end{align*}
which does not satisfy the irrepresentable conditions I and II defined in Definition \ref{def: irrepresentable_model} because $|-1.01|>1$.  
\end{example}

\subsection{OU Process} \label{appendix: calculation_of_corr_OU}

Deriving explicit formulas for calculating the exact correlation between signature components of OU processes (both It\^o and Stratonovich) is complicated. Here we provide an example to show the general approach for calculating the correlation. The proof of this example is given in Appendix \ref{appendix: proofs}, and one can use a similar routine to compute the correlation for other setups of OU processes. 

\begin{example}\label{exmp: 1D_OU}
Consider a 1-dimensional OU process $\mathbf{X}_t = Y_t$ with a mean reversion speed $\kappa>0$, which is driven by
\begin{equation} \label{equ:Exmp_OU}
    \mathrm{d} Y_t = - \kappa Y_t \mathrm{d} t + \mathrm{d} W_t, \quad Y_0 = 0.
\end{equation}
The correlation coefficients between its 0-th order and 2nd order of signature are
    \begin{align*}
         \frac{ \mathbb{E} \left[S(\mathbf{X})_T^{0,I} S(\mathbf{X})_T^{1,1,I}\right] }{ \sqrt{ \mathbb{E}\left[S(\mathbf{X})_T^{0,I}\right]^2 \mathbb{E}\left[S(\mathbf{X})_T^{1,1,I}\right]^2 } } &= \frac{ -2\kappa T - e^{-2\kappa T} + 1 }{ 
 \sqrt{  4\kappa T e^{-2\kappa T} + 3 e^{-4\kappa T} - 6 e^{-2\kappa T} - 4 \kappa T + 3 + 4\kappa^2 T^2 } } , \label{equ: example_OU_Ito_corr} \\
          \frac{ \mathbb{E} \left[S(\mathbf{X})_T^{0,S} S(\mathbf{X})_T^{1,1,S}\right] }{ \sqrt{ \mathbb{E}\left[S(\mathbf{X})_T^{0,S}\right]^2 \mathbb{E}\left[S(\mathbf{X})_T^{1,1,S}\right]^2 } } &= \frac{\sqrt{3}}{3}, %\label{equ: example_OU_S_corr} 
    \end{align*}
    for It\^o and Stratonovich signature, respectively. The proof is provided in Appendix \ref{appendix: proofs}.

Figure \ref{fig: corr_vs_kappa} shows the absolute values of correlation coefficients between the 0-th order and 2nd order signature components calculated using the formulas above under different values of $\kappa$ with $T=1$. Notably, the correlation for It\^o signature increases with respect to $\kappa$, while the correlation for Stratonovich signature remains fixed at $\sqrt{3}/{3}$.

\begin{figure}[htbp]
    \centering
    \caption{Absolute values of correlation coefficients between signature components of the 1-dimensional OU process. Solid (dashed) lines correspond to the It\^o (Stratonovich) signature. \label{fig: example_OU}}
    \subfigure[Correlation between the 0-th and the 2nd order signature components.\label{fig: corr_vs_kappa}]{
        \includegraphics[width=0.45\linewidth]{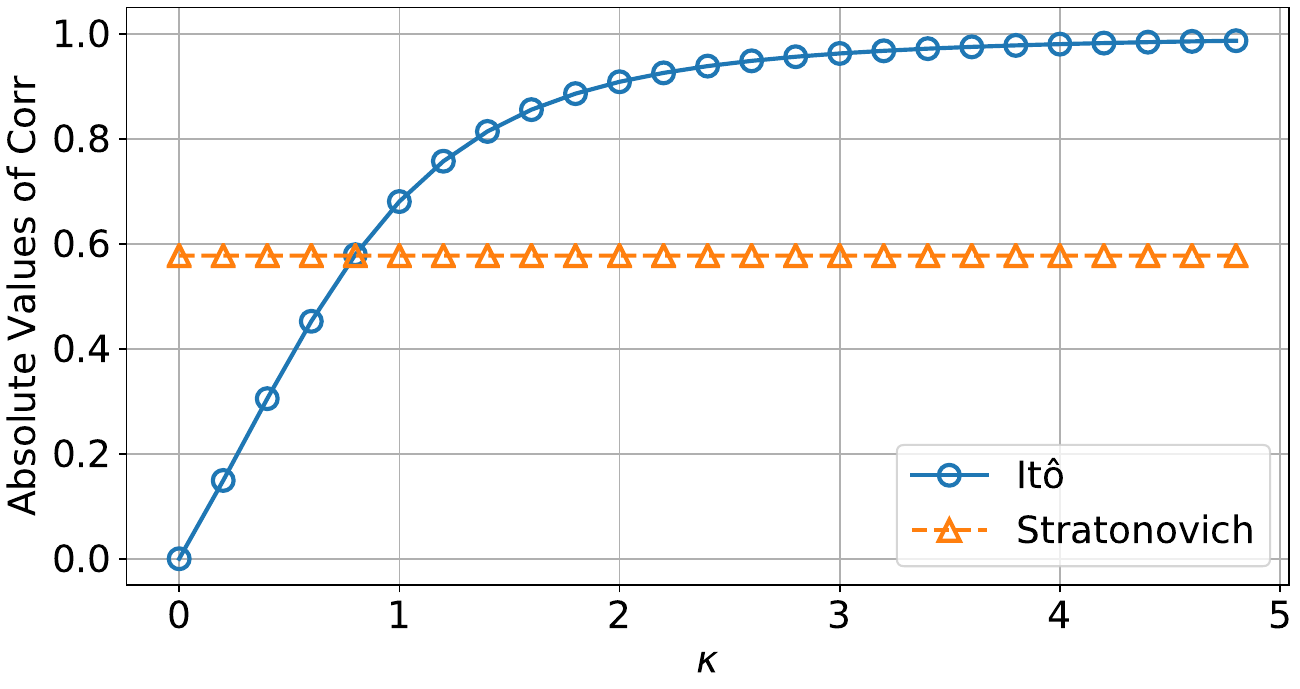}
    }
    \hfill
    \subfigure[Correlation between the first four order signature components. \label{fig: Experiment_15_abs}]{
        \includegraphics[width=0.45\linewidth]{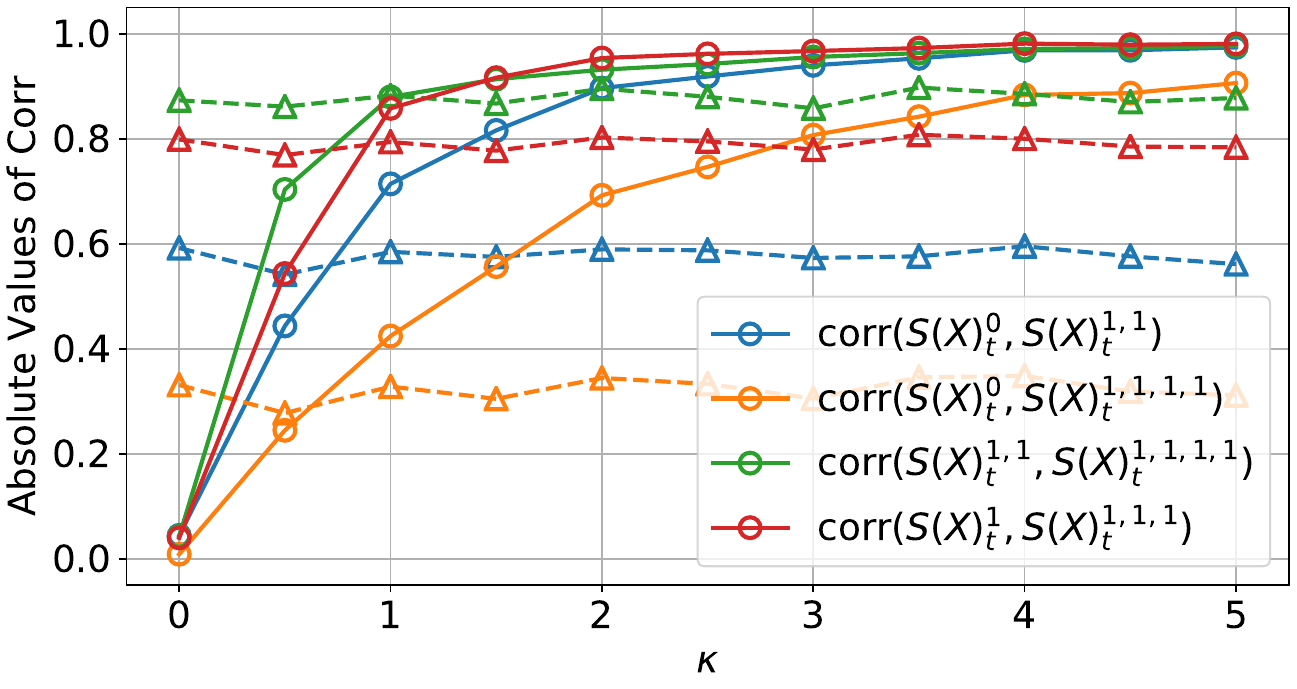}
    }
\end{figure}

We further perform simulations to estimate the correlation coefficients for higher-order signature components of the OU process. We generate 10,000 sample paths of the OU process using the methods discussed in Appendix \ref{appendix: discrete}. For each path, we calculate the corresponding signature components and then estimate the sample correlation matrix based on the 10,000 simulated samples. Figure \ref{fig: Experiment_15_abs} shows the simulation results for the absolute values of correlation coefficients between the first four order signature components under different values of $\kappa$. Consistent with the observation in Figure \ref{fig: corr_vs_kappa}, the correlations for It\^o signature increase with respect to $\kappa$, while the correlations for Stratonovich signature remain relatively stable. Notably, the correlations for It\^o signature are zero when $\kappa=0$, which reduces to the results for a Brownian motion. In addition, when $\kappa$ is sufficiently large, the absolute values of correlation coefficients for It\^o signature exceed those for Stratonovich signature.

\end{example}

Recall that the irrepresentable condition, as defined in Definition \ref{def: irrepresentable_model}, illustrates that a higher correlation generally leads to poorer consistency. Therefore, based on Example \ref{exmp: 1D_OU}, we can expect that the Lasso is more consistent when using It\^o signature for small values of $\kappa$ (weaker mean reversion), and more consistent when using Stratonovich signature for large values of $\kappa$ (stronger mean reversion). This provides a theoretical explanation for our observations in Section \ref{sec:simulation_mean_revert} of the main paper---When processes are sufficiently rough or mean reverting \citep{el2018microstructural,gatheral2018volatility}, using Lasso with Stratonovich signature will likely lead to higher statistical consistency compared to It\^o signature.

\section{Technical Details for Consistency of Signature}
\subsection{Tightness of the Sufficient Condition for Consistency}
\label{appendix:tight_sufficient}

In this appendix, we investigate the irrepresentable condition for It\^o signature of a multi-dimensional Brownian motion with constant inter-dimensional correlation. This analysis not only provides further insights into the irrepresentable condition but also demonstrates the tightness of the sufficient condition presented in Theorem \ref{prop: sufficient_irrepresentable} in our main paper.

The following proposition characterizes the irrepresentable condition for a Brownian motion with constant inter-dimensional correlation when using It\^o signature. For mathematical simplicity, we assume that only the first order signature components are included in the regression model. 
\begin{proposition} \label{prop: sufficient_irrepresentable_tight}
For a multi-dimensional Brownian motion given by \eqref{equ: MultiBM_setup} with equal inter-dimensional correlation $\rho = \rho_{ij}$, assume that only its first order It\^o signature components are included in \eqref{equ: linearregression}, and that all true beta coefficients are positive. Then, the irrepresentable conditions I and II hold if $ \rho  \in( -\frac{1}{2 \# A_1^* }, 1)$, and do not hold if $\rho \in ( -\frac{1}{\# A_1^* } , -\frac{1}{2 \# A_1^* } ]$. 
\end{proposition}

\begin{remark}
    Proposition \ref{prop: sufficient_irrepresentable_tight} only discusses the results for $\rho \in (-\frac{1}{\# A_1^* } ,1 )$. If $\rho \leq -\frac{1}{\# A_1^* } $, then the inter-dimensional correlation matrix for the Brownian motion is not positive definite. 
\end{remark}
 
Proposition \ref{prop: sufficient_irrepresentable_tight} demonstrates that the sufficient condition \eqref{equ:bound_irc} is tight when the inter-dimensional correlation $\rho$ is constant and negative. Meanwhile, for $\rho>0$, the irrepresentable conditions always hold but may not satisfy \eqref{equ:bound_irc}.

\subsection{Consistency of Lasso with General Predictors in Finite Sample}
\label{appendix:consistency_general_feature}
In this appendix, we present additional results on the consistency of Lasso with general predictors (not necessarily signature components) in finite sample. 

Consider a linear regression model with $N$ samples and $p$ predictors $X_1,\dots,X_p$ given by 
\begin{equation} \label{equ: generallinearregression}
    y = X \boldsymbol\beta + \varepsilon,
\end{equation}
where $\varepsilon \in \mathbb{R}^N$ is a vector of independent and normally distributed white noise with mean zero and variance $\sigma^2$, $X \in \mathbb{R}^{N \times p}$ is the random design matrix with each row represents a random sample of $(X_1,\dots,X_p)^\top$, $y \in \mathbb{R}^N$ is the target to predict, and $\boldsymbol\beta \in \mathbb{R}^p$ is the vector of beta coefficients. Assume that $X$ has full column rank. Given a tuning parameter $\lambda>0$, we adopt the Lasso estimator given by
\begin{align}
    \hat{{\boldsymbol \beta}}^N(\lambda) = \arg\min_{\hat{\boldsymbol \beta}}\left\{ \left\|  y - \tilde{X} \hat{\boldsymbol\beta} \right\|_2^2  + \lambda \left\|\hat{\boldsymbol \beta} \right\|_1   \right\} \label{equ: generallasso}
\end{align}
to identify the true predictors, where $\tilde{X}$ represents the standardized version of $X$ across $N$ samples by the $l_2$-norm, whose $(n,j)$-entry is defined by
\begin{equation*}
    \tilde{X}_{n,j} = \frac{ X_{n,j} }{  \sqrt{ \sum_{m=1}^N (X_{m,j})^2 \Big/ N } }, \quad n=1,2,\dots,N;~ j=1,2,\dots,p.
\end{equation*}
Therefore, the sample covariance matrix calculated using $\tilde{X}$ is the same as the sample correlation matrix of $X$.

Denote by $\hat{\Delta}$ and $\Delta$ the sample correlation matrix and the population correlation matrix of all predictors in the Lasso regression, respectively. Because the number of samples, $N$, is finite, $\hat{\Delta}$ may deviate from $\Delta$. Therefore, when studying the consistency of Lasso, $\hat{\Delta}$ may not satisfy the irrepresentable condition even if $\Delta$ does. Nevertheless, in this appendix, we show that $\hat{\Delta}$ will satisfy the irrepresentable condition with high probability. 

Our analysis aligns with existing works in high-dimensional statistics. For example, \cite{wainwright2009sharp} assumes that the predictors are normally distributed, while \cite{cai2022sparse} and \cite{sample2023lasso} assume a sub-Gaussian distribution. However, these results cannot be directly applied in our setup, because the predictors in our paper are signature components, which are neither Gaussian nor sub-Gaussian. 

We first introduce some notations. Denote the set of true predictors by $A^*$, false predictors by $A^{*c}$, the number of true predictors by $q=\#A^*$, the population covariance matrix of all predictors in the Lasso regression by $\Sigma$, the population correlation matrix of all predictors in the Lasso regression by $\Delta$, the correlation matrix between predictors in sets $A$ and $B$ by $\Delta_{AB}$, and the volatility of components of $\varepsilon$ in \eqref{equ: generallinearregression} by $\sigma$. We also let $\tilde{\boldsymbol \beta}$ be the vector containing all standardized beta coefficients of the true model whose $j$-th component is given by
\begin{equation*}
    \tilde{\beta}_j = \beta_j \cdot \sqrt{\frac{1}{N} \sum_{m=1}^N (X_{m,j})^2  }.
\end{equation*}

The following result shows the consistency of Lasso under the assumption that all predictors have finite fourth moments. 
\begin{theorem}\label{th:samplelasso3}
    For the Lasso regression given by \eqref{equ: generallinearregression} and \eqref{equ: generallasso}, assume that the following two conditions hold:
    \begin{enumerate}[(i)]
\item The irrepresentable condition II in Definition \ref{def: irrepresentable_model} holds for the population correlation matrix, i.e., there exists some $\gamma \in (0,1]$ such that $\left\|\Delta_{A^{*c}A^{*}}\Delta_{A^{*}A^{*}}^{-1} \right\|_{\infty} \leq 1-\gamma $;
\item The predictors have finite fourth moments, i.e., there exists $K<\infty$ such that $\mathbb{E}[X_i^4]\leq K$ for all $i=1,\dots,p$.
%\item The design matrix is full column rank.
\end{enumerate}
In addition, we assume that the sequence of regularization parameters $\{\lambda_N\}$ satisfies $\lambda_N > \frac{4\sigma}{\gamma}\sqrt{\frac{2\ln p}{N}}$. Then, the following properties hold with probability greater than 
\begin{equation*}
    \left(1-\frac{8p^4\sigma_{\max}^4(\sigma_{\min}^4+K)}{N\xi^2\sigma_{\min}^4}\right)\left(1-4e^{-cN\lambda_N^2}\right) 
\end{equation*}
 for some constant $c>0$.
 \begin{enumerate}[(a)]
     \item The Lasso has a unique solution $\hat{\boldsymbol\beta}^N(\lambda_N) \in \mathbb{R}^p$ with its support contained within the true support, and satisfies
 $$\left\|\hat{\boldsymbol\beta}^N(\lambda_N) - \tilde{\boldsymbol\beta}\right\|_\infty \leq \lambda_N \left[\frac{\zeta(2+2\alpha\zeta+\gamma)}{2+2\alpha\zeta} + \frac{4\sigma}{\sqrt{\frac{1}{2}C_{\min}}}\right] =: h(\lambda_N);$$
\item If in addition $\min_{i \in A^*} |\tilde{ \beta}_i| > h(\lambda_N)$, then $\mathrm{sign}(\hat{\boldsymbol\beta}^N(\lambda_N))=\mathrm{sign}(\tilde{\boldsymbol\beta})$.
 \end{enumerate}
 Here, $\sigma_{\min} = \min_{1 \leq i \leq p}\sqrt{\Sigma_{ii}}$, $\sigma_{\max} = \max_{1 \leq i \leq p}\sqrt{\Sigma_{ii}}$, $\alpha = \left\|\Delta_{A^{*c}A^{*}}\right\|_{\infty}$, $\zeta = \left\|\Delta_{A^{*}A^{*}}^{-1}\right\|_{\infty}$, 
$C_{\min} = \Lambda_{\min}(\Delta_{A^{*}A^{*}}) = \frac{1}{\left\|\Delta_{A^{*}A^{*}}^{-1}\right\|_2}  > 0$, and $\xi = \min\left\{g_\Sigma^{-1}\left(\frac{\gamma}{\zeta(2+2\alpha\zeta+\gamma)}\right),g_\Sigma^{-1}\left(\frac{C_{\min}}{2\sqrt{p}}\right)\right\} >0$, where the definition of $g_{\Sigma}(\cdot)$ is given by \eqref{equ:def_g_x}.
\end{theorem}

A detailed proof of Theorem \ref{th:samplelasso3} can be found in Appendix \ref{appendix: proofs}.

\section{Additional Details for Simulation} \label{appendix: discrete}
This appendix provides technical details, computational cost, more numerical experiments, and robustness checks for the simulations conducted in this paper.

\subsection{More Technical Details}
Throughout our simulations in the paper, we set the time index $0 = t_0 < t_1 < \dots < t_n = T$ with $t_{k+1} - t_k = \Delta t = T/n$ for any $k \in \{0,1,\dots,n-1\}$ and $n=100$.

\textbf{Simulation of Processes.} We simulate the $i$-th dimension of the Brownian motion $W_t^i$, and OU process $Y_t^i$, by discretizing the stochastic differential equations of the processes using the Euler--Maruyama schemes given by
\begin{itemize}
    \item Brownian motion: $W_{t_{k+1}}^i = W_{t_{k}}^i + \sqrt{\Delta t} \varepsilon_k^i$, $W_0^i = 0$;
    \item OU process: $Y_{t_{k+1}}^i = Y_{t_{k}}^i - \kappa_i Y_{t_{k}}^i \Delta t + \sqrt{\Delta t} \varepsilon_k^i$, $Y_0^i = 0$,
\end{itemize}
with $\varepsilon_k^i$ randomly drawn from the standard normal distribution. %The number of steps is set to $N=100$. 

The $i$-th dimension of the random walk and AR(1) model, both denoted by $Z_t^i$, are simulated using the following formulas.
\begin{itemize}
    \item Random walk: $Z_{t_{k+1}}^i = Z_{t_{k}}^i + e_k^i$, $Z_0^i = 0$;
    \item AR(1) model: $Z_{t_{k+1}}^i = \phi_i Z_{t_{k}}^i + \varepsilon_k^i$, $Z_0^i = 0$,
\end{itemize}
with $e_k^i$ randomly drawn from% the following distribution
\begin{equation*}
    \mathbb{P} (e_k^i = +1) = \mathbb{P} (e_k^i = -1) = 0.5,
\end{equation*}
and $\varepsilon_k^i$  randomly drawn from the standard normal distribution. %The number of steps is set to $N=100$. 

After simulating each dimension of the processes, we simulate the inter-dimensional correlation between different dimensions of the processes using the Cholesky decomposition. 
Finally, we generate $\mathbf{X}$ using \eqref{equ: MultiBM_setup} or \eqref{equ: OU_setup}. 

In all the simulations, we set the length of the processes $T=1$, and the initial values of the processes to zero. These choices have no impact on the results because the signature of a path $\mathbf{X}$ is invariant under a time reparametrization and a shift of the starting point of $\mathbf{X}$ \citep{chevyrev2016primer}.

\textbf{Calculation of Integrals.} The calculation of It\^o and Stratonovich signatures requires the calculation of It\^o and Stratonovich integrals. By definition, they are computed using the following schemes.
\begin{itemize}
    \item It\^o integral: $\int_0^T A_t \mathrm{d} B_t \approx \sum_{k=0}^{n-1} A_{t_k} (B_{t_{k+1}}-B_{t_{k}}) $;
    \item Stratonovich integral: $\int_0^T A_t \circ \mathrm{d} B_t \approx \sum_{k=0}^{n-1} \frac{1}{2} (A_{t_k}+A_{t_{k+1}}) (B_{t_{k+1}}-B_{t_{k}}) $.
\end{itemize} 

\subsection{Computational Details}\label{appendix:comp_detail}
\begin{itemize}
\item The simulations are implemented using Python 3.7.
\item The simulations are run on a laptop with an Intel(R) Core(TM) i7-9750H CPU @ 2.60GHz.
\item The random seed is set to 0 for reproducibility.
\item The Lasso regressions are performed using the \texttt{sklearn.linear\_model.lars\_path} package.
\item Each individual experiment, including generating 100 paths, calculating their signatures, and performing the Lasso regression, can be completed within one second.
\end{itemize}

\subsection{Impact of the Dimension of the Process and the Number of Samples} \label{appendix:choiceofd}

Most simulations in Section \ref{sec:simulation} of our main paper consider the case of $d=2$ (dimension of the process) and $N=100$ (number of samples).

\begin{figure}[htbp]
    \centering
    \caption{Consistency rates for the Brownian motion and the OU process with different numbers of dimensions $d$ and different values of inter-dimensional correlation $\rho$. Solid (dashed) lines correspond to the It\^o (Stratonovich) signature. \label{fig:Figure_rebuttal_1}}
    \subfigure[Brownian motion.\label{fig:Figure_rebuttal_1a}]{
        \includegraphics[width=0.45\linewidth]{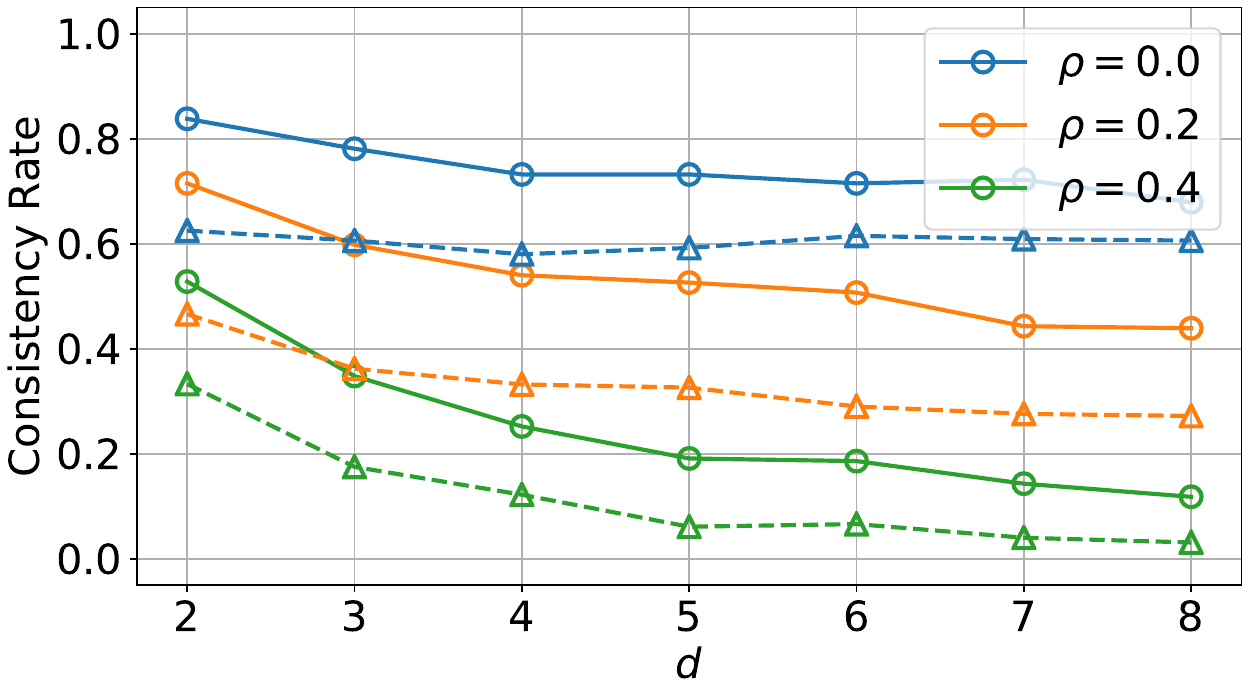}
    }
    \hfill
    \subfigure[OU process.\label{fig:Figure_rebuttal_1b}]{
        \includegraphics[width=0.45\linewidth]{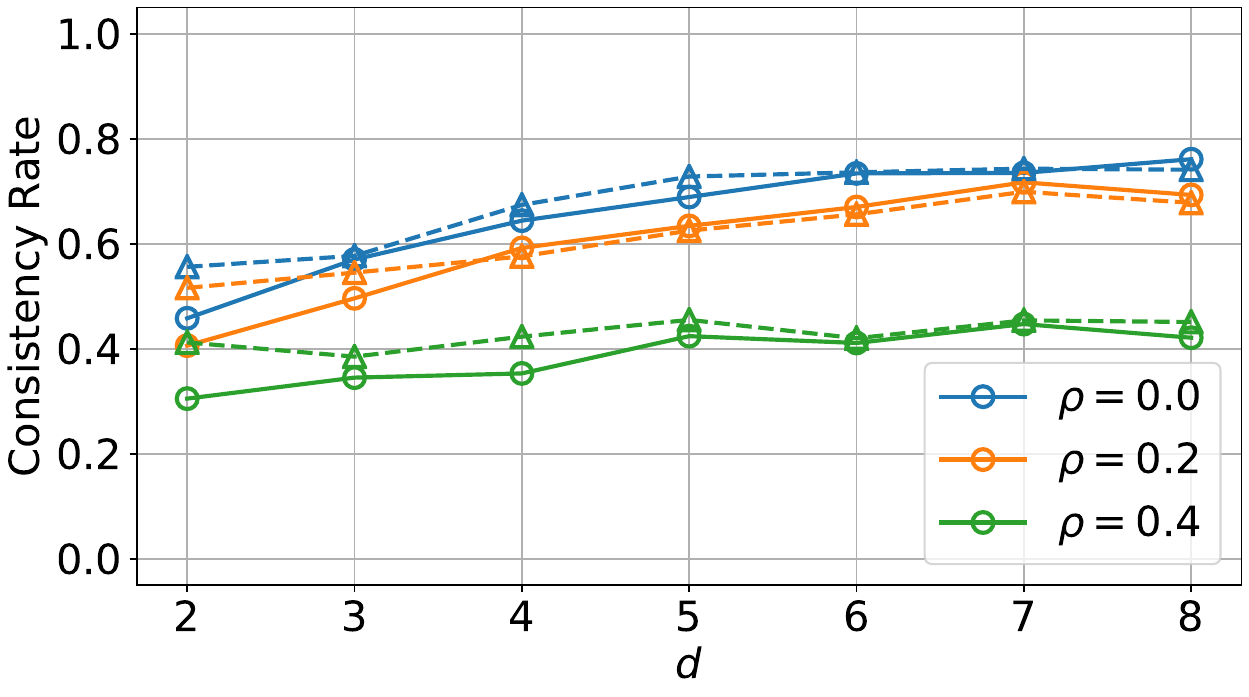}
    }
\end{figure}

Figure \ref{fig:Figure_rebuttal_1} shows how the consistency of Lasso varies with the dimension of the process $d$, with Figure \ref{fig:Figure_rebuttal_1a} for the Brownian motion and Figure \ref{fig:Figure_rebuttal_1b} for the OU process with $\kappa = 2$. We set the number of true predictors to be three. Other simulation setups remain the same as in Section \ref{sec:simulation_consistency} of the main paper. 

First, for the Brownian motion, the consistency rate decreases with $d$. This can be attributed to the fact that the inter-dimensional correlation of the process leads to stronger correlations between signature components as more dimensions are included. Second, for the OU process, the consistency rate increases with $d$ because the inter-dimensional correlation of the process is weaker than the correlation between the increments of the OU process itself.

\begin{figure}[htbp]
    \centering
    \caption{Consistency rates for the Brownian motion and the OU process with different numbers of samples $N$ and different values of inter-dimensional correlation $\rho$. Solid (dashed) lines correspond to the It\^o (Stratonovich) signature. \label{fig:Figure_rebuttal_3}}
    \subfigure[Brownian motion.\label{fig:Figure_rebuttal_3a}]{
        \includegraphics[width=0.45\linewidth]{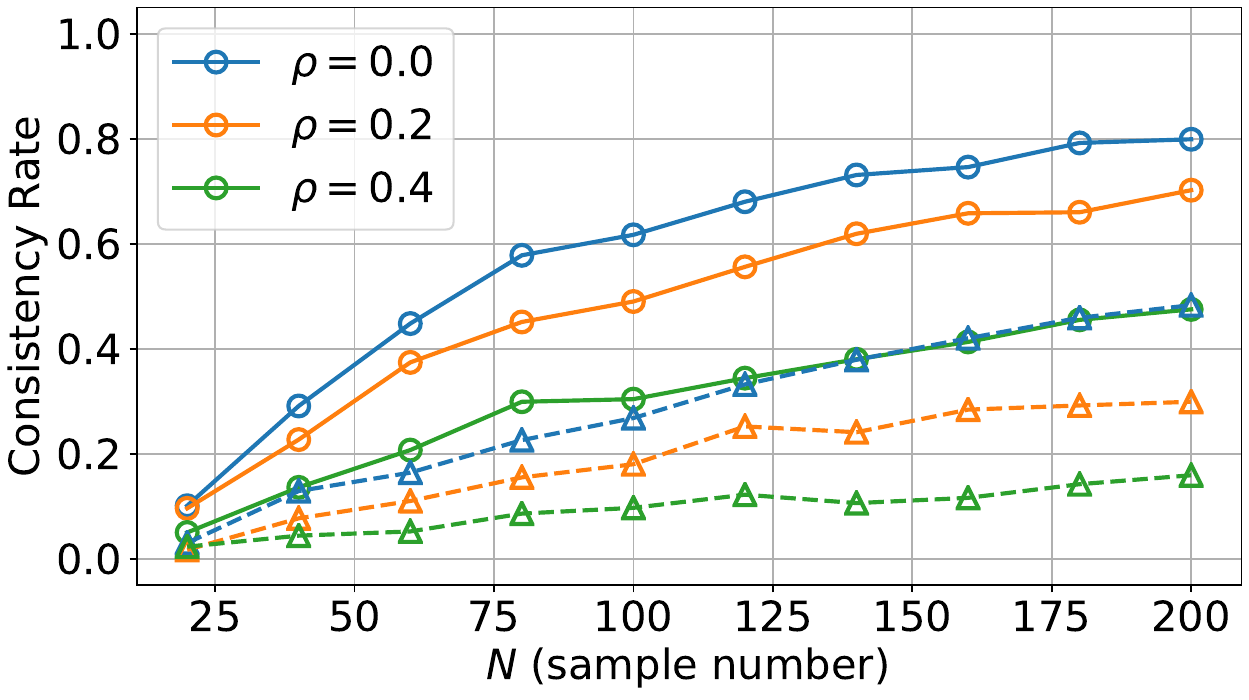}
    }
    \hfill
    \subfigure[OU process.\label{fig:Figure_rebuttal_3b}]{
        \includegraphics[width=0.45\linewidth]{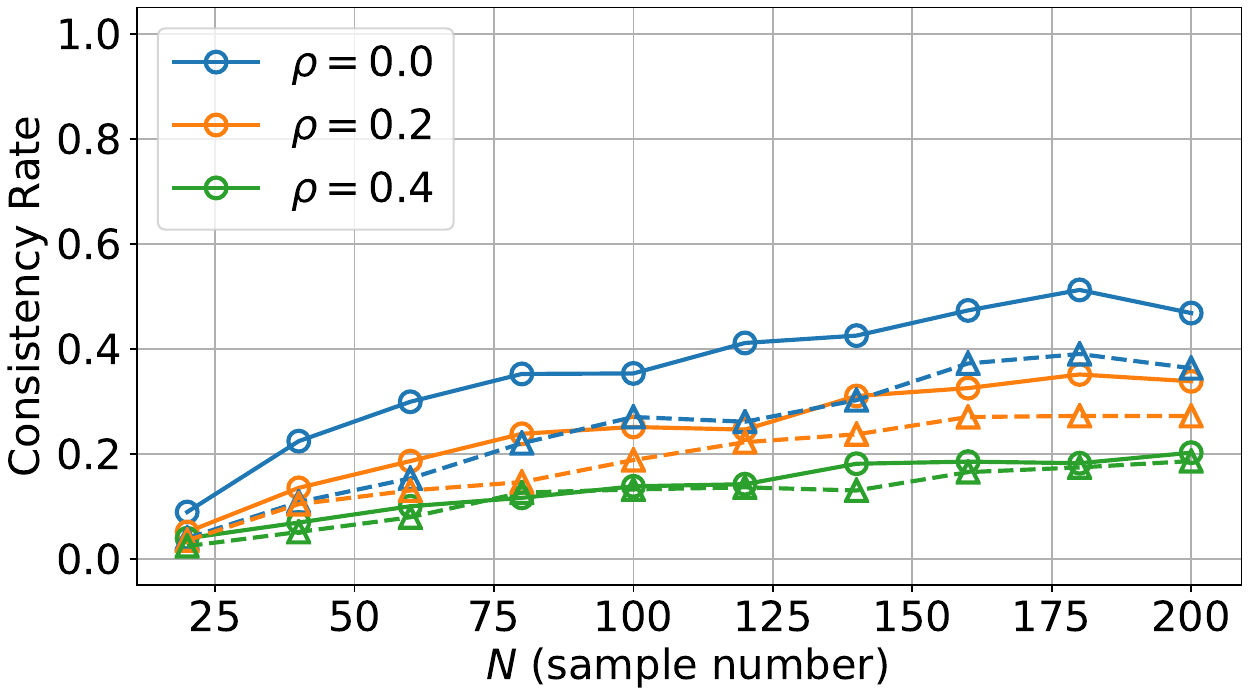}
    }
\end{figure}

Figure \ref{fig:Figure_rebuttal_3} shows the relationship between the consistency rate and the number of samples. In general, we find that the consistency rate increases as the number of samples increases.

\subsection{The ARIMA Process} \label{appendix:ARIMA}

This appendix examines the consistency of signature for the ARIMA($p,I,q$) model, where $p$ is the lag of AR, $I$ is the degree of differencing, and $q$ is the lag of MA.

\begin{figure}[htbp]
    \centering
    \caption{Consistency rates for the ARIMA($p,I,q$) with different lags of AR, $p$, lags of MA, $q$, and degrees of differencing, $I$. Solid (dashed) lines correspond to the It\^o (Stratonovich) signature. \label{fig:Figure_rebuttal_2}}
    \subfigure[Consistency rates for different $p$ and $I$.\label{fig:Figure_rebuttal_2a}]{
        \includegraphics[width=0.45\linewidth]{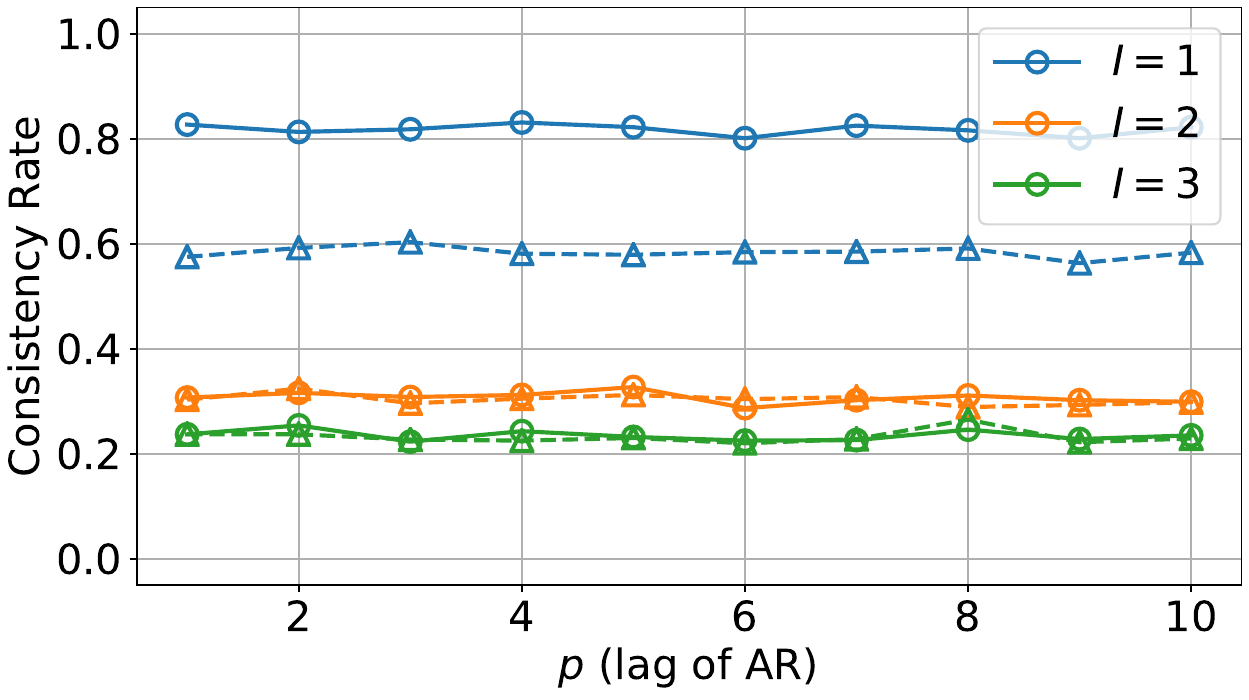}
    }
    \hfill
    \subfigure[Consistency rates for different $q$ and $I$.\label{fig:Figure_rebuttal_2b}]{
        \includegraphics[width=0.45\linewidth]{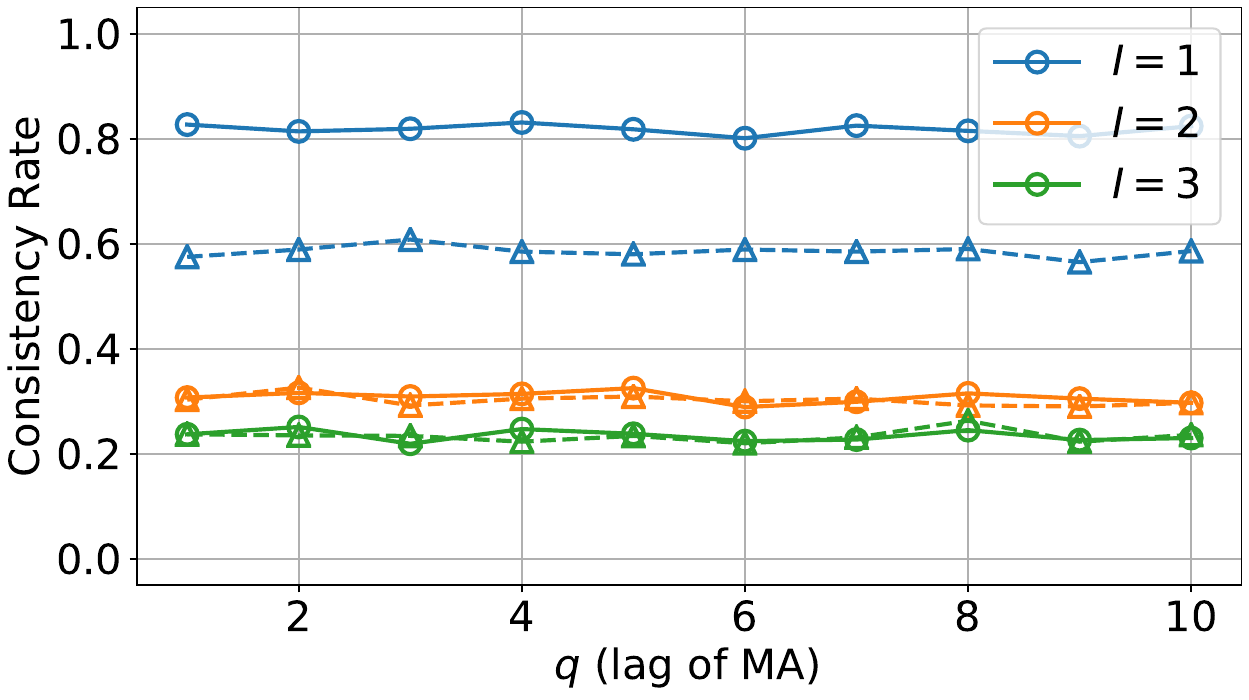}
    }
\end{figure}

Figure \ref{fig:Figure_rebuttal_2} shows how the consistency rate varies with $p$, $q$, and $I$. We find that the consistency rate does not exhibit any apparent dependence on $p$ and $q$, but does highly rely on $I$. Specifically, the consistency rate generally decreases as $I$ increases due to the stronger correlation between the increments of the ARIMA processes introduced by $I$. 

\subsection{Robustness Checks} \label{subsec:robust}
\begin{figure}[htbp]
    \centering
    \caption{Consistency rates for the Brownian motion and the random walk with different values of inter-dimensional correlation $\rho$ and different numbers of true predictors $q$. Solid (dashed) lines correspond to the It\^o (Stratonovich) signature. Shaded regions are confidence intervals of the experiments.  \label{fig: Experiment_3_withbar}}
    \subfigure[Brownian motion.\label{fig: Experiment_3_BM_withbar}]{
        \includegraphics[width=0.45\linewidth]{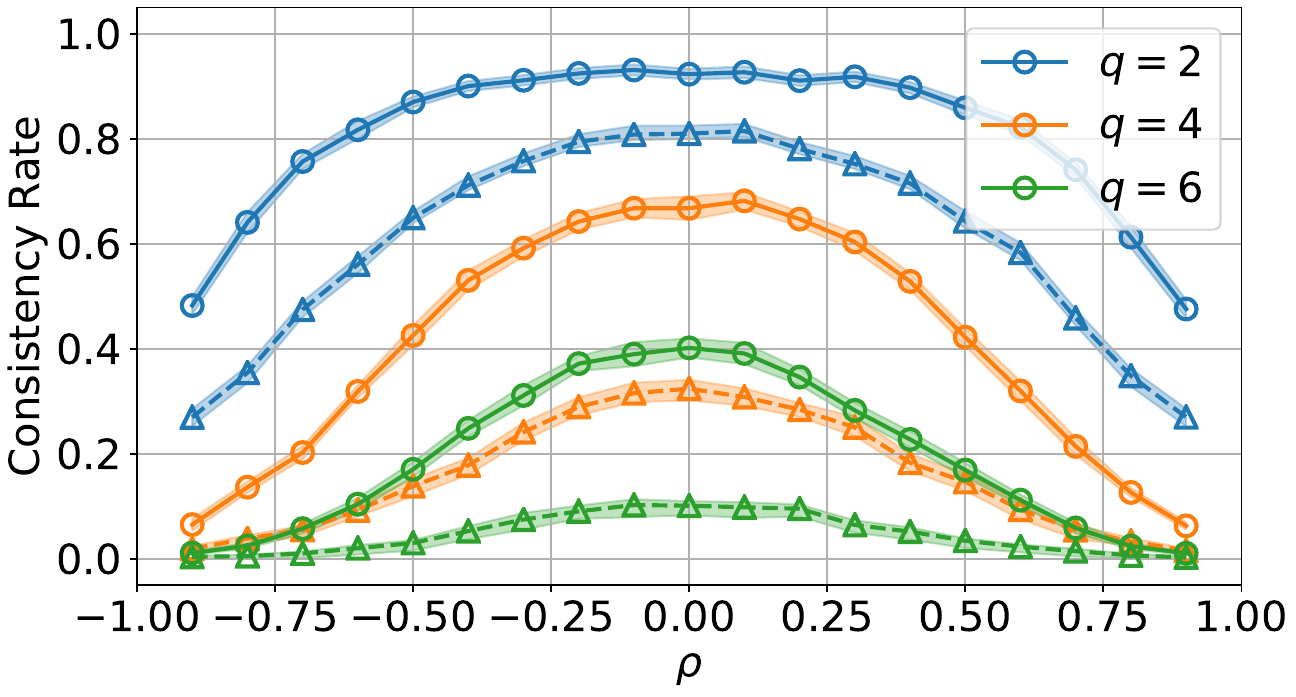}
    }
    \hfill
    \subfigure[Random walk.\label{fig: Experiment_3_RW_withbar}]{
        \includegraphics[width=0.45\linewidth]{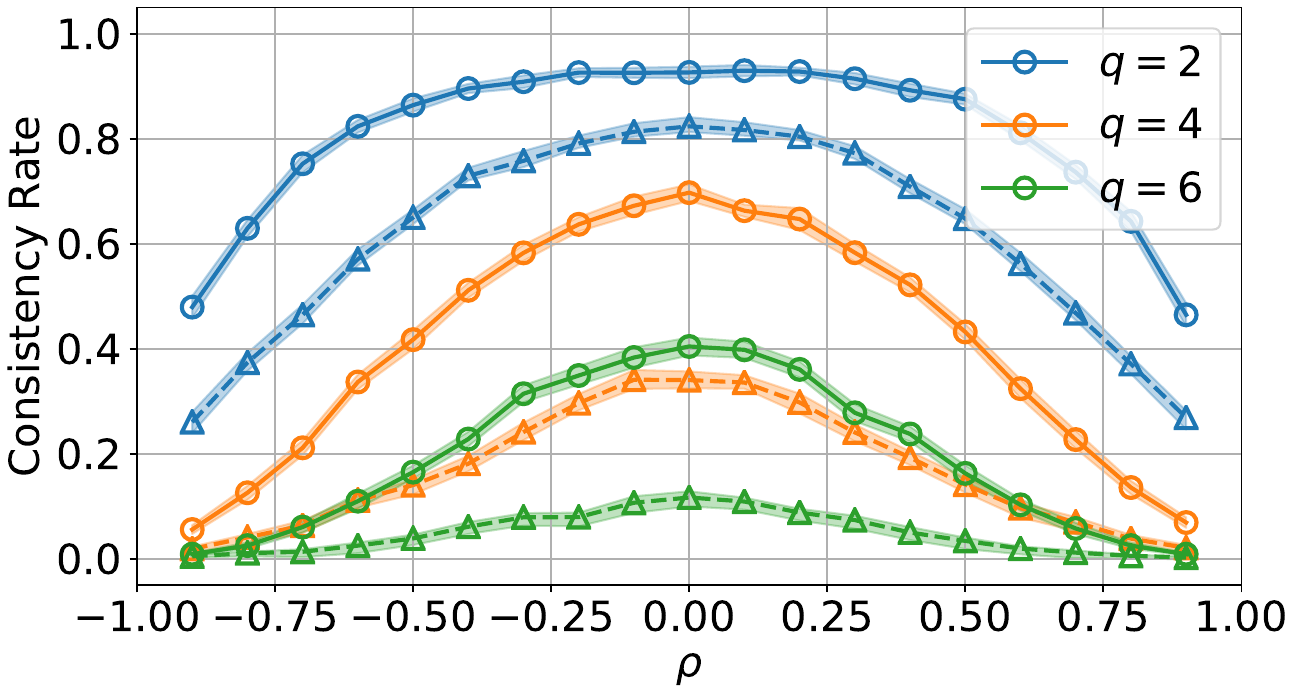}
    }
\end{figure}

\begin{figure}[htbp]
    \centering
    \caption{Consistency rates for the OU process and the AR(1) model with different parameters ($\kappa$ and $1-\phi$) and different numbers of true predictors $q$. Solid (dashed) lines correspond to the It\^o (Stratonovich) signature. Shaded regions are confidence intervals of the experiments. \label{fig: Experiment_12and13_withbar}}
    \subfigure[OU process.\label{fig: Experiment_12_withbar}]{
        \includegraphics[width=0.45\linewidth]{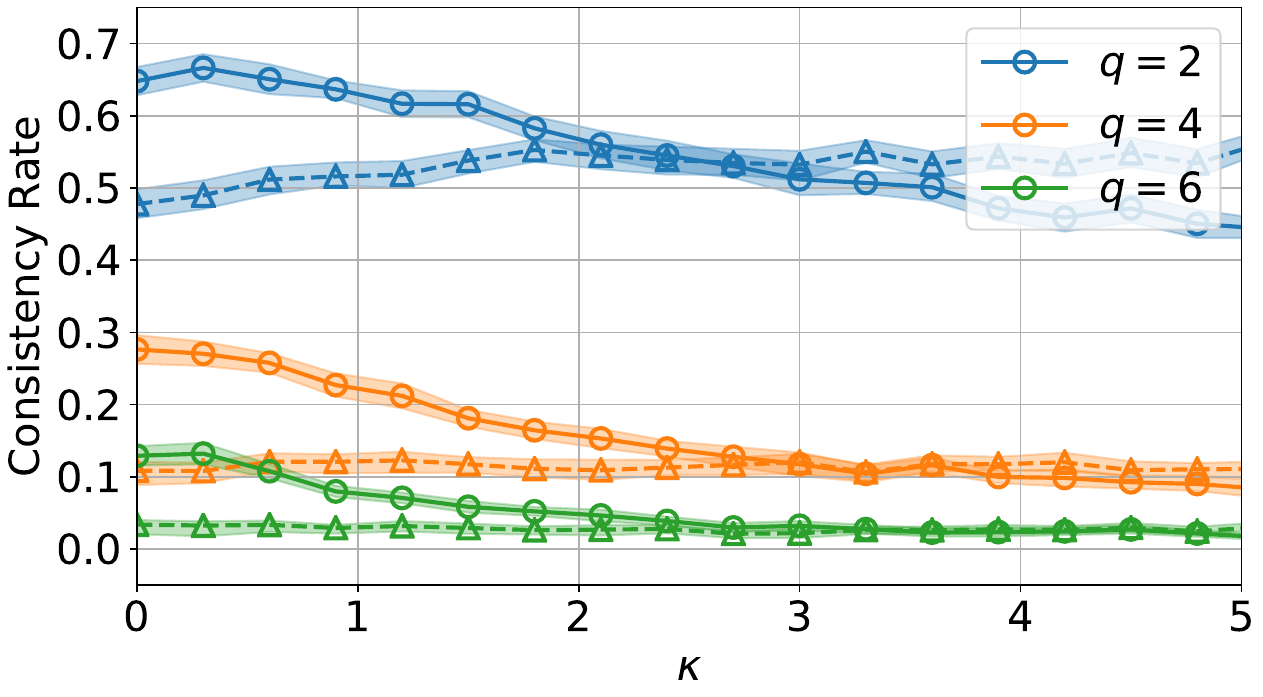}
    }
    \hfill
    \subfigure[AR(1) model.\label{fig: Experiment_AR_withbar}]{
        \includegraphics[width=0.45\linewidth]{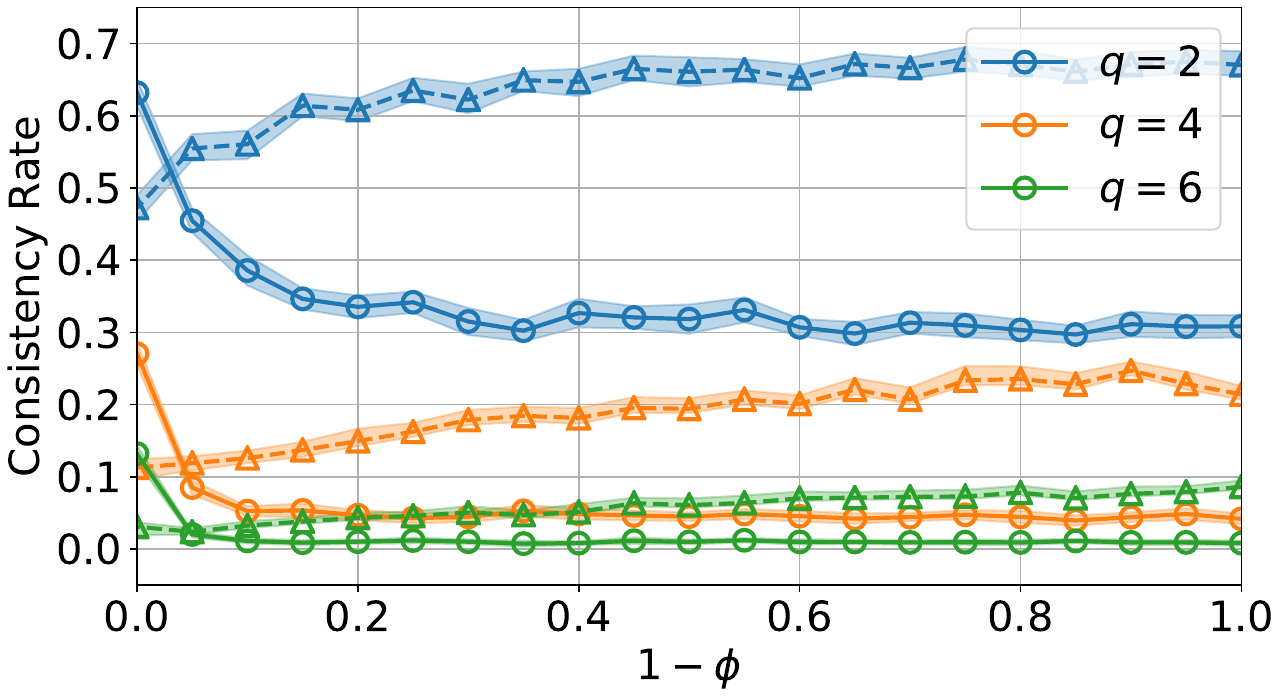}
    }
\end{figure}

To show the robustness of our simulations shown in Section \ref{sec:simulation} of the main paper, we present Figures \ref{fig: Experiment_3_withbar} and \ref{fig: Experiment_12and13_withbar}, which include confidence intervals (shaded regions) for the estimated consistency rates of the Brownian motion/random walk and OU process/AR(1) model, respectively.

In Figures \ref{fig: Experiment_3_withbar} and \ref{fig: Experiment_12and13_withbar}, we estimate the consistency rate by repeating the procedure described in Section \ref{sec:simulation_consistency} 100 times, and this process is repeated 30 times to obtain the confidence interval for the estimation. Thus, these confidence intervals are based on 30 estimations of the consistency rate, with each estimation calculated using 100 experiments.

We observe that the confidence intervals of the consistency rates shown in Figures \ref{fig: Experiment_3_withbar} and \ref{fig: Experiment_12and13_withbar} are narrow. Moreover, the observations made in Section \ref{sec:simulation} are consistent with the results presented here, further confirming the robustness of our findings.

\section{Lemmas and Proofs} \label{appendix: proofs}
This appendix provides the proofs of all theoretical results in this article and lemmas used in the proofs. 

\subsection{Lemmas}
\begin{lemma}\label{lemmafund}
Assume that $\hat{\Sigma}$ and $\Sigma$ are $p\times p$ positive definite matrices with diagonal entries $\{\hat{\sigma}_i^2\}_{i=1}^p$ and $\{{\sigma}_i^2\}_{i=1}^p$, respectively, with $\sigma_{\min} = \min_{1 \leq i \leq p}\sigma_{i}$ and $\sigma_{\max} = \max_{1 \leq i \leq p} \sigma_{i}$. Let $\hat{\Delta}$ and $\Delta$ be $p\times p$ matrices with $(i,j)$-entries $\hat{\Delta}_{ij} = \hat{\Sigma}_{ij} / (\hat{\sigma}_i \hat{\sigma}_{j})$ and  ${\Delta}_{ij} = {\Sigma}_{ij} / (\sigma_i \sigma_{j})$ for $i,j=1,2,\dots,p$, respectively. For $\epsilon < \sigma_{\min}^2$, if $\|\hat{\Sigma} - \Sigma \|_\infty \leq \epsilon $, we have $\|\hat{\Delta} - \Delta \|_\infty \leq g_{\Sigma}(\epsilon)$, where $g_{\Sigma}(\cdot)$ is given by \eqref{equ:def_g_x}. 
\end{lemma}

\proof{Proof of Lemma \ref{lemmafund}.}

For any $i,j=1,2,\dots,p$, 
\begin{equation*}
    |\hat{\Sigma}_{ij} - \Sigma_{ij}| \leq \|\hat{\Sigma} - \Sigma\|_\infty \leq \epsilon,
\end{equation*}
which implies that
\begin{equation*}
    \hat{\Sigma}_{ij} \in \left( \Sigma_{ij}-\epsilon ,\Sigma_{ij}+ \epsilon \right).
\end{equation*}
Hence, for $\epsilon < \sigma_{\min}^2$,
\begin{align*}
    \hat{\sigma}_i \in \left( \sqrt{\sigma_i^2- \epsilon },\sqrt{\sigma_i^2+\epsilon }\right).
\end{align*}
Now we estimate the difference between $\Delta_{ij}$ and $\hat{\Delta}_{ij}$. If $\Delta_{ij}>0$,
%Suppose $\Delta_{ij}>0$, then we have:
\begin{align}
    \Delta_{ij} - \hat{\Delta}_{ij} &= \frac{\Sigma_{ij}}{\sigma_i \sigma_j} - \frac{\hat{\Sigma}_{ij}}{\hat{\sigma}_i \hat{\sigma}_j} \leq \frac{\Sigma_{ij}}{\sigma_i \sigma_j} - \frac{\Sigma_{ij}- \epsilon}{\sqrt{\sigma_i^2+ \epsilon } \cdot \sqrt{\sigma_j^2+ \epsilon }} = \frac{\Sigma_{ij} \sqrt{\sigma_i^2+ \epsilon}  \sqrt{\sigma_j^2+ \epsilon } - \left(\Sigma_{ij}- \epsilon \right) \sigma_i \sigma_j}{\sigma_i \sigma_j \sqrt{\sigma_i^2+ \epsilon} \sqrt{\sigma_j^2+\epsilon }}\notag\\
    & \leq  \frac{\Sigma_{ij}\left(\sqrt{\sigma_i^2+\epsilon }  \sqrt{\sigma_j^2+ \epsilon }-\sigma_i\sigma_j\right) + \epsilon \sigma_i\sigma_j}{\sigma_i^2 \sigma_j^2 } =\frac{\Sigma_{ij}\cdot \frac{ \epsilon  (\sigma_i^2+\sigma_j^2) +  \epsilon^2 }{\sqrt{\sigma_i^2+ \epsilon }  \sqrt{\sigma_j^2+ \epsilon }+\sigma_i\sigma_j} +  \epsilon \sigma_i\sigma_j}{\sigma_i^2 \sigma_j^2 } \notag\\
    & \leq \frac{\Sigma_{ij}\cdot \frac{ \epsilon  (\sigma_i^2+\sigma_j^2) +  \epsilon^2 }{2\sigma_i\sigma_j} +  \epsilon \sigma_i\sigma_j}{\sigma_i^2 \sigma_j^2 }  = \Delta_{ij} \cdot  \frac{ \epsilon  (\sigma_i^2+\sigma_j^2) + \epsilon^2 }{2\sigma_i^2 \sigma_j^2} + \frac{ \epsilon }{\sigma_i\sigma_j} \leq\Delta_{ij} \cdot \frac{2 \epsilon  \sigma_{\min}^2 + \epsilon^2 }{2\sigma_{\min}^4} + \frac{\epsilon }{\sigma_{\min}^2}.\label{eq:deltadiff1}
\end{align}
Meanwhile,
\begin{align}
     \hat{\Delta}_{ij} - \Delta_{ij} =&  \frac{\hat{\Sigma}_{ij}}{\hat{\sigma}_i \hat{\sigma}_j} - \frac{\Sigma_{ij}}{\sigma_i \sigma_j}  \leq  \frac{\Sigma_{ij}+ \epsilon }{\sqrt{\sigma_i^2- \epsilon } \cdot \sqrt{\sigma_j^2-\epsilon}}- \frac{\Sigma_{ij}}{\sigma_i \sigma_j} 
    = \frac{\left(\Sigma_{ij}+ \epsilon \right) \sigma_i \sigma_j -\Sigma_{ij} \sqrt{\sigma_i^2- \epsilon}  \sqrt{\sigma_j^2- \epsilon}}{\sigma_i \sigma_j \sqrt{\sigma_i^2- \epsilon } \sqrt{\sigma_j^2- \epsilon}}\notag\\
    =& \frac{\Sigma_{ij} \cdot \left(\sigma_i\sigma_j-\sqrt{\sigma_i^2- \epsilon }  \sqrt{\sigma_j^2- \epsilon }\right) + \epsilon \sigma_i \sigma_j}{\sigma_i \sigma_j \sqrt{\sigma_i^2-\epsilon } \sqrt{\sigma_j^2- \epsilon}}\notag \\
    =& \Delta_{ij}\cdot \frac{ \epsilon (\sigma_i^2+\sigma_j^2)-\epsilon^2 }{\sqrt{\sigma_i^2- \epsilon } \sqrt{\sigma_j^2- \epsilon}\left(\sigma_i\sigma_j+\sqrt{\sigma_i^2-\epsilon }  \sqrt{\sigma_j^2- \epsilon }\right) } + \frac{ \epsilon}{\sqrt{\sigma_i^2- \epsilon } \sqrt{\sigma_j^2- \epsilon }}\notag \\
    \leq& \Delta_{ij}\cdot \frac{ \epsilon (\sigma_i^2+\sigma_j^2) }{\sqrt{\sigma_i^2- \epsilon } \sqrt{\sigma_j^2- \epsilon}\left(\sigma_i\sigma_j+\sqrt{\sigma_i^2-\epsilon }  \sqrt{\sigma_j^2- \epsilon }\right) } + \frac{ \epsilon}{\sqrt{\sigma_i^2- \epsilon } \sqrt{\sigma_j^2- \epsilon }} \notag\\
    \leq& \Delta_{ij}\cdot \frac{2 \epsilon \sigma_{\min}^2}{\sqrt{\sigma_{\min}^2-\epsilon} \sqrt{\sigma_{\min}^2- \epsilon }\left(\sigma_{\min}^2+\sqrt{\sigma_{\min}^2- \epsilon }  \sqrt{\sigma_{\min}^2-\epsilon }\right) } + \frac{ \epsilon }{\sqrt{\sigma_{\min}^2- \epsilon} \sqrt{\sigma_{\min}^2- \epsilon }} \notag\\
    =& \Delta_{ij}\cdot \frac{2 \epsilon \sigma_{\min}^2}{(\sigma_{\min}^2- \epsilon ) \left(2\sigma_{\min}^2- \epsilon \right) } + \frac{ \epsilon }{\sigma_{\min}^2- \epsilon }.\label{eq:deltadiff2}
\end{align}
Combining \eqref{eq:deltadiff1} and \eqref{eq:deltadiff2}, we see
\begin{align*}
     |\hat{\Delta}_{ij} - \Delta_{ij}| \leq \Delta_{ij}\cdot \frac{2 \epsilon \sigma_{\min}^2}{(\sigma_{\min}^2- \epsilon ) \left(2\sigma_{\min}^2- \epsilon \right) } + \frac{ \epsilon }{\sigma_{\min}^2- \epsilon }.
\end{align*}
For the case of $\Delta_{ij}\leq 0$, one can similarly establish
\begin{align*}
     |\hat{\Delta}_{ij} - \Delta_{ij}| \leq |\Delta_{ij}|\cdot \frac{2 \epsilon \sigma_{\min}^2}{(\sigma_{\min}^2- \epsilon ) \left(2\sigma_{\min}^2- \epsilon \right) } + \frac{ \epsilon }{\sigma_{\min}^2- \epsilon}.
\end{align*}
Hence,
\begin{align*}
     \sum_{1 \leq j \leq p} |\hat{\Delta}_{ij} - \Delta_{ij}| = \sum_{1 \leq j \leq p, j\neq i} |\hat{\Delta}_{ij} - \Delta_{ij}| &\leq  \frac{2 \epsilon \sigma_{\min}^2}{(\sigma_{\min}^2- \epsilon ) \left(2\sigma_{\min}^2- \epsilon \right) } \cdot \sum_{1 \leq j \leq p, j\neq i}|\Delta_{ij}|+ \frac{(p-1) \epsilon }{\sigma_{\min}^2- \epsilon }\\
     &\leq \frac{2\epsilon\sigma_{\min}^2 (p-1)\rho}{(\sigma_{\min}^2- \epsilon) \left(2\sigma_{\min}^2- \epsilon \right) } + \frac{(p-1) \epsilon }{\sigma_{\min}^2- \epsilon }.
\end{align*}
Finally,
\begin{align*}
\|\hat{\Delta} - \Delta \|_\infty \leq \frac{2\epsilon\sigma_{\min}^2 (p-1)\rho}{(\sigma_{\min}^2- \epsilon) \left(2\sigma_{\min}^2- \epsilon \right) } + \frac{(p-1) \epsilon }{\sigma_{\min}^2- \epsilon } =g_{\Sigma}(\epsilon). \tag*{\Halmos}
\end{align*}
\endproof
\begin{lemma}\label{lemma1}
    Let $A$ and $B$ be invertible $p \times p$ matrices
    satisfying $\|I-A^{-1}B\| <1$, where $I$ is an $p\times p$ identity matrix. Then, $$\|A^{-1}-B^{-1}\|\leq \frac{\|A^{-1}\|^2\|A-B\|}{1-\|A^{-1}\|\|A-B\|}.$$
    Here, $\| \cdot \|$ is any specific sub-multiplicative matrix norm. 
\end{lemma}
\proof{Proof of Lemma \ref{lemma1}.}
    Since $\|I-A^{-1}B\| <1$, we have $B^{-1}A = (A^{-1}B)^{-1} = \sum_{n=0}^\infty (I-A^{-1}B)^n.$ Thus, $B^{-1} = \sum_{n=0}^\infty (I-A^{-1}B)^n A^{-1}$. Hence,
\begin{align*}
    \|A^{-1}-B^{-1}\| &= \left\| I-\sum_{n=0}^\infty (I-A^{-1}B)^n A^{-1}  \right\|= \left\| \sum_{n=1}^\infty (I-A^{-1}B)^n A^{-1} \right\|\\
    &\leq \|A^{-1}\| \cdot \sum_{n=1}^\infty \|I-A^{-1}B\|^n= \|A^{-1}\| \cdot \frac{\|I-A^{-1}B\|}{1-\|I-A^{-1}B\|}=\|A^{-1}\| \cdot \frac{\|A^{-1}(A-B)\|}{1-\|A^{-1}(A-B)\|}\\
    &\leq \|A^{-1}\| \cdot \frac{\|A^{-1}\| \cdot \|A-B\|}{1-\|A^{-1}\| \cdot \|A-B\|}=\frac{\|A^{-1}\|^2\|A-B\|}{1-\|A^{-1}\|\|A-B\|}.\tag*{\Halmos}
\end{align*}
% This completes the proof.
\endproof
\begin{lemma}\label{le:signaturebound}
    Let $\mathbf{X}$ be a $d$-dimensional Brownian motion given by \eqref{equ: MultiBM_setup} or an OU process given by \eqref{equ: OU_setup}. For any $k=1,2,\dots$, there exists a constant $ \lambda_k<\infty$ such that for all $0\leq t\leq T$ and $i_1,\dots,i_k \in \{1,2,\dots,d\}$,  
    \begin{equation}\mathbb{E} \left( S(\mathbf{X})_{t}^{i_1,\dots,i_k, I} \right)^4 \leq \lambda_k,\label{eq:itobound}\end{equation}
    and
    \begin{equation}\mathbb{E} \left( S(\mathbf{X})_{t}^{i_1,\dots,i_k, S} \right)^4 \leq \lambda_k.\label{eq:Sbound}\end{equation}
\end{lemma}
\proof{Proof of Lemma \ref{le:signaturebound}.}
The proofs for OU process and Brownian motion are similar, and we will focus on the case of the Brownian motion. We first prove \eqref{eq:itobound} by induction. Let $\sigma_{\max} = \max_{j=1,\dots,d} \sigma_j$. When $k=1$, 
\begin{align*}
    \mathbb{E}\left( S(\mathbf{X})_{t}^{i_1, I}\right)^4 = \mathbb{E}( X_{t}^{i_1})^4 = 3\sigma_{i_1}^4t^2 \leq 3 \sigma_{\max}^4 T^2 =: \lambda_1 < \infty.
\end{align*}
Now, for $n>1$, assume that \eqref{eq:itobound} holds for all $k<n$. Then, for $k=n$, the quadratic variation of the It\^o signature component satisfies
\begin{align*}
    \mathbb{E} \left( \left[S(\mathbf{X})^{i_1,\dots,i_n, I}\right]_t \right)^2 &= \mathbb{E} \left( \int_0^t \left(S(\mathbf{X})_{s}^{i_1,\dots,i_{n-1}, I}\right)^2 \sigma_{i_n}^2 \mathrm{d}s \right)^2\\
    &= \sigma_{i_n}^4 \cdot \mathbb{E} \int_0^t \int_0^t \left(S(\mathbf{X})_{w}^{i_1,\dots,i_{n-1}, I}\right)^2 \left(S(\mathbf{X})_{s}^{i_1,\dots,i_{n-1}, I}\right)^2 \mathrm{d}w\mathrm{d}s \\
    &= \sigma_{i_n}^4 \cdot \int_0^t \int_0^t \mathbb{E}\left(  \left(S(\mathbf{X})_{w}^{i_1,\dots,i_{n-1}, I}\right)^2 \left(S(\mathbf{X})_{s}^{i_1,\dots,i_{n-1}, I}\right)^2 \right) \mathrm{d}w\mathrm{d}s \\
    &\leq \sigma_{i_n}^4 \cdot \int_0^t \int_0^t \sqrt{\mathbb{E}\left(S(\mathbf{X})_{w}^{i_1,\dots,i_{n-1}, I}\right)^4 \cdot \mathbb{E}\left(S(\mathbf{X})_{s}^{i_1,\dots,i_{n-1}, I}\right)^4} \mathrm{d}w\mathrm{d}s \\
    &\leq \sigma_{i_n}^4 \cdot \int_0^t \int_0^t \sqrt{\lambda_{n-1} \cdot \lambda_{n-1}} \mathrm{d}w\mathrm{d}s = \lambda_{n-1}\sigma_{i_n}^4 t^2.
\end{align*}
Thus, by the Burkholder--Davis--Gundy inequality, there exists a constant $c<\infty$ such that for all $0\leq t\leq T$ and $i_1,\dots,i_n \in \{1,2,\dots,d\}$,  
\begin{align*}
    \mathbb{E} \left(S(\mathbf{X})_{t}^{i_1,\dots,i_n, I}\right)^4 \leq c \cdot \mathbb{E} \left( \left[S(\mathbf{X})^{i_1,\dots,i_n, I}\right]_t \right)^2  \leq c\lambda_{n-1}\sigma_{i_n}^4 t^2 \leq c\lambda_{n-1}\sigma_{\max}^4 T^2 =: \lambda_{n} < \infty.
\end{align*}
This implies that \eqref{eq:itobound} holds when $k=n$, which completes the proof of \eqref{eq:itobound}.

Now we prove \eqref{eq:Sbound}. By the relationship between the Stratonovich integral and the It\^o integral, we have
    \begin{align*}
        S(\mathbf{X})_t^{i_1,\dots,i_{k},S} &= \int_0^t S(\mathbf{X})_s^{i_1,\dots,i_{k-1},S} 
        \circ \mathrm{d} X_s^{i_k} \\
        &= \int_0^t S(\mathbf{X})_s^{i_1,\dots,i_{k-1},S} 
         \mathrm{d} X_s^{i_k} + \frac{1}{2} \left[ S(\mathbf{X})^{i_1,\dots,i_{k-1},S}, X^{i_k} \right]_t,
    \end{align*}
    where $[A,B]_t$ represents the quadratic covariation between processes $A$ and $B$ from time 0 to $t$. Furthermore, by properties of the quadratic covariation, 
    \begin{align*}
        \left[ S(\mathbf{X})^{i_1,\dots,i_{k-1},S}, X^{i_k} \right]_t &= \int_0^t S(\mathbf{X})_s^{i_1,\dots,i_{k-2},S} 
         \mathrm{d} \left[ X^{i_{k-1}}, X^{i_k} \right]_s \\
         &= \rho_{i_{k-1}i_{k}} \sigma_{i_{k-1}} \sigma_{i_{k}} \int_0^t S(\mathbf{X})_s^{i_1,\dots,i_{k-2},S} 
         \mathrm{d} s.
    \end{align*}
   Therefore,
   \begin{equation}\label{eq:SandIto}
       S(\mathbf{X})_t^{i_1,\dots,i_{k},S} =  \int_0^t S(\mathbf{X})_s^{i_1,\dots,i_{k-1},S} 
         \mathrm{d} X_s^{i_k} + \frac{1}{2} \rho_{i_{k-1}i_{k}} \sigma_{i_{k-1}} \sigma_{i_{k}} \int_0^t S(\mathbf{X})_s^{i_1,\dots,i_{k-2},S} 
         \mathrm{d} s.
   \end{equation}
We prove \eqref{eq:Sbound} by induction. Let $\sigma_{\max} = \max_{j=1,\dots,d} \sigma_j$.  
When $k=1$, we have
\begin{align*}
    \mathbb{E}\left( S(\mathbf{X})_{t}^{i_1, S}\right)^4 = \mathbb{E}\left( S(\mathbf{X})_{t}^{i_1, I}\right)^4 = \mathbb{E}( X_{t}^{i_1})^4 = 3\sigma_{i_1}^4t^2  \leq 3 \sigma_{\max}^4 T^2 =: \lambda_1 < \infty.
\end{align*}
When $k=2$, by \eqref{eq:itobound} and \eqref{eq:SandIto}, there exists a constant $C$ such that
\begin{align*}
    \mathbb{E}\left( S(\mathbf{X})_{t}^{i_1,i_2, S}\right)^4 = \mathbb{E}\left( S(\mathbf{X})_{t}^{i_1,i_2, I} + \frac{1}{2}\rho_{i_1 i_2} \sigma_{i_1}  \sigma_{i_2} t \right)^4 &\leq 8\mathbb{E}\left( S(\mathbf{X})_{t}^{i_1,i_2, I} \right)^4 + \frac{1}{2}\rho_{i_1 i_2}^4 \sigma_{i_1}^4  \sigma_{i_2}^4 t^4 \\
    & \leq  8 C + \frac{1}{2}\sigma_{\max}^8 T^4 =: \lambda_2 < \infty.
\end{align*}
For $n>2$, assume that \eqref{eq:Sbound} holds for all $k<n$. Thus, for $k=n$, we have
\begin{align*}
    \mathbb{E} \left(\left[\int_0^t S(\mathbf{X})_s^{i_1,\dots,i_{n-1},S} 
         \mathrm{d} X_s^{i_n}\right]_t\right)^2 &= \mathbb{E}\left( \int_0^t \left(S(\mathbf{X})_s^{i_1,\dots,i_{n-1},S}\right)^2 \sigma_{i_n}^2
         \mathrm{d} s\right)^2 \\
        &= \sigma_{i_n}^4 \cdot \mathbb{E} \int_0^t\int_0^t \left(S(\mathbf{X})_w^{i_1,\dots,i_{n-1},S}\right)^2\left(S(\mathbf{X})_s^{i_1,\dots,i_{n-1},S}\right)^2 \mathrm{d} w \mathrm{d} s\\
        &= \sigma_{i_n}^4 \cdot  \int_0^t\int_0^t \mathbb{E} \left( \left(S(\mathbf{X})_w^{i_1,\dots,i_{n-1},S}\right)^2\left(S(\mathbf{X})_s^{i_1,\dots,i_{n-1},S}\right)^2 \right) \mathrm{d} w \mathrm{d} s\\
        &\leq \sigma_{i_n}^4 \cdot \int_0^t \int_0^t \sqrt{\mathbb{E}\left(S(\mathbf{X})_{w}^{i_1,\dots,i_{n-1}, S}\right)^4 \cdot \mathbb{E}\left(S(\mathbf{X})_{s}^{i_1,\dots,i_{n-1}, S}\right)^4} \mathrm{d}w\mathrm{d}s \\
        &\leq \sigma_{i_n}^4 \cdot \int_0^t \int_0^t \sqrt{\lambda_{n-1} \cdot \lambda_{n-1}} \mathrm{d}w\mathrm{d}s = \lambda_{n-1}\sigma_{i_n}^4 t^2.
\end{align*}
Hence, by the Burkholder--Davis--Gundy inequality, there exists a constant $c<\infty$ such that for all $0\leq t\leq T$ and $i_1,\dots,i_n \in \{1,2,\dots,d\}$, 
\begin{align*}
    \mathbb{E} \left( \int_0^t S(\mathbf{X})_s^{i_1,\dots,i_{n-1},S} 
         \mathrm{d} X_s^{i_n}\right)^4 \leq c \cdot \mathbb{E} \left( \left[\int_0^t S(\mathbf{X})_s^{i_1,\dots,i_{n-1},S} 
         \mathrm{d} X_s^{i_n}\right]_t \right)^2  \leq c\lambda_{n-1}\sigma_{i_n}^4 t^2 .
\end{align*}
In addition, we have
\begin{align*}
    &\mathbb{E}\left(\int_0^t S(\mathbf{X})_s^{i_1,\dots,i_{n-2},S} \mathrm{d} s\right)^4 \\= &\mathbb{E} \int_0^t\int_0^t\int_0^t\int_0^t S(\mathbf{X})_w^{i_1,\dots,i_{n-2},S}S(\mathbf{X})_s^{i_1,\dots,i_{n-2},S}S(\mathbf{X})_u^{i_1,\dots,i_{n-2},S}S(\mathbf{X})_v^{i_1,\dots,i_{n-2},S} \mathrm{d} w\mathrm{d} s\mathrm{d} u\mathrm{d} v\\
    =& \int_0^t\int_0^t\int_0^t\int_0^t\mathbb{E}  \left(  S(\mathbf{X})_w^{i_1,\dots,i_{n-2},S}S(\mathbf{X})_s^{i_1,\dots,i_{n-2},S}S(\mathbf{X})_u^{i_1,\dots,i_{n-2},S}S(\mathbf{X})_v^{i_1,\dots,i_{n-2},S} \right) \mathrm{d} w\mathrm{d} s\mathrm{d} u\mathrm{d} v\\
    \leq& \int_0^t\int_0^t\int_0^t\int_0^t \frac{1}{4} \Big( \mathbb{E}\left(S(\mathbf{X})_w^{i_1,\dots,i_{n-2},S}\right)^4+\mathbb{E}\left(S(\mathbf{X})_s^{i_1,\dots,i_{n-2},S}\right)^4\\
    & \qquad\qquad\qquad\qquad\qquad\qquad\qquad  +\mathbb{E}\left(S(\mathbf{X})_u^{i_1,\dots,i_{n-2},S}\right)^4+\mathbb{E}\left(S(\mathbf{X})_v^{i_1,\dots,i_{n-2},S}\right)^4  \Big)\mathrm{d} w\mathrm{d} s\mathrm{d} u\mathrm{d} v\\
    \leq& \int_0^t\int_0^t\int_0^t\int_0^t \frac{1}{4}\cdot 4\lambda_{n-2} \mathrm{d} w\mathrm{d} s\mathrm{d} u\mathrm{d} v = \lambda_{n-2} t^4 .
\end{align*}
Therefore, by \eqref{eq:SandIto},
\begin{align*}
    \mathbb{E} \left(S(\mathbf{X})_{t}^{i_1,\dots,i_n, S}\right)^4 &\leq 8\mathbb{E} \left( \int_0^t S(\mathbf{X})_s^{i_1,\dots,i_{n-1},S} 
         \mathrm{d} X_s^{i_n}\right)^4 + \frac{1}{2}\rho_{i_{n-1}i_{n}}^4\sigma_{i_{n-1}}^4 \sigma_{i_{n}}^4 \mathbb{E}\left(\int_0^t S(\mathbf{X})_s^{i_1,\dots,i_{n-2},S} \mathrm{d} s\right)^4\\
        & \leq 8c\lambda_{n-1}\sigma_{i_n}^4 t^2 + \frac{1}{2}\rho_{i_{n-1}i_{n}}^4\sigma_{i_{n-1}}^4 \sigma_{i_{n}}^4\lambda_{n-2} t^4 \\
        & \leq 8c\lambda_{n-1}\sigma_{\max}^4 T^2 + \frac{1}{2}\sigma_{\max}^8 \lambda_{n-2} T^4 =:\lambda_n <\infty .
\end{align*}
This implies that \eqref{eq:Sbound} holds when $k=n$, which completes the proof.
\Halmos\endproof

\begin{lemma}\label{th:samplelasso1}
For the Lasso regression given by \eqref{equ: generallinearregression} and \eqref{equ: generallasso}, assume that conditions (i) and (ii) in Theorem \ref{th:samplelasso3} hold. Then, $$\mathbb{P} \left(\Lambda_{\min}(\hat{\Delta}_{A^{*}A^{*}})\geq \frac{1}{2}C_{\min} \right)\geq 1-\frac{4p^4\sigma_{\max}^4(\sigma_{\min}^4+K)}{N\xi^2\sigma_{\min}^4}$$ holds with $\xi=g_\Sigma^{-1}\left(\frac{C_{\min}}{2\sqrt{p}}\right)>0$, and the definition of $g_{\Sigma}(\cdot)$ and other notations the same as in Theorem \ref{th:samplelasso3}. 
\end{lemma}

\proof{Proof of Lemma \ref{th:samplelasso1}.}
Condition (ii) implies that
\begin{align*}
    \mathbb{E}\left[\frac{X_i^4}{\Sigma_{ii}^2}\right] \leq \mathbb{E}\left[\frac{X_i^4}{\sigma_{\min}^4}\right] \leq \frac{K}{\sigma_{\min}^4}.
\end{align*}
Hence, by \citet[Lemma 2]{ravikumar2011high}, for any $i,j \in \{1,\dots,p\}$, 
\begin{align*}
    \mathbb{P}\left( \left|\hat{\Sigma}_{ij}-\Sigma_{ij} \right|>\frac{\xi}{p}\right) \leq \frac{4p^2\sigma_{\max}^4(1+\frac{K}{\sigma_{\min}^4})}{N\xi^2}.
\end{align*}
Thus,
\begin{align*}
    \mathbb{P}\left(\sum_{j=1}^p \left|\hat{\Sigma}_{ij}-\Sigma_{ij} \right|\leq \xi\right)  \geq 1-\frac{4p^3\sigma_{\max}^4(1+\frac{K}{\sigma_{\min}^4})}{N\xi^2},
\end{align*}
which further implies that
\begin{equation} \label{equ:prob_2norm_bound}
    \mathbb{P}\left(\|\hat{\Sigma}-\Sigma\|_\infty \leq \xi\right) \geq 1-\frac{4p^4\sigma_{\max}^4(1+\frac{K}{\sigma_{\min}^4})}{N\xi^2} = 1-\frac{4p^4\sigma_{\max}^4(\sigma_{\min}^4+K)}{N\xi^2\sigma_{\min}^4}.
\end{equation}
Therefore, by Lemma \ref{lemmafund}, we have 
\begin{align*}
    \mathbb{P}\left(\|\hat{\Delta}-\Delta\|_2 \leq \frac{C_{\min}}{2}\right) &\geq \mathbb{P}\left(\|\hat{\Delta}-\Delta\|_\infty \leq \frac{C_{\min}}{2\sqrt{p}}\right) \geq \mathbb{P}\left(\|\hat{\Sigma}-\Sigma\|_\infty \leq g_\Sigma^{-1}\left(\frac{C_{\min}}{2\sqrt{p}}\right)\right)\\
    &\geq \mathbb{P}\left(\|\hat{\Sigma}-\Sigma\|_\infty \leq \xi\right)\geq
    1-\frac{4p^4\sigma_{\max}^4(\sigma_{\min}^4+K)}{N\xi^2\sigma_{\min}^4}.
\end{align*}
Now, whenever $\|\hat{\Delta}-\Delta\|_2 \leq \frac{C_{\min}}{2}$ holds, we have
\begin{align*}
\|I-\Delta_{A^*A^*}^{-1}\hat{\Delta}_{A^*A^*}\|_2 &\leq \|\Delta_{A^*A^*}^{-1}\|_2 \cdot \|\Delta_{A^*A^*}-\hat{\Delta}_{A^*A^*}\|_2 \\&= \frac{1}{C_{\min}} \cdot \|\Delta_{A^*A^*}-\hat{\Delta}_{A^*A^*}\|_2 \leq \frac{1}{C_{\min}} \cdot \|\hat{\Delta}-\Delta\|_2 \leq \frac{1}{2}<1,
\end{align*}
which implies that
\begin{align*}
\|\hat{\Delta}_{A^*A^*}^{-1}\|_2 \leq \|\hat{\Delta}_{A^*A^*}^{-1} - \Delta_{A^*A^*}^{-1}\|_2 + \|\Delta_{A^*A^*}^{-1}\|_2&= \|\hat{\Delta}_{A^*A^*}^{-1} - \Delta_{A^*A^*}^{-1}\|_2 + \frac{1}{C_{\min}} \\ &\leq \frac{\frac{1}{C_{\min}^2}\cdot \frac{C_{\min}}{2}}{1-\frac{1}{C_{\min}}\cdot \frac{C_{\min}}{2}} + \frac{1}{C_{\min}} = \frac{2}{C_{\min}},
\end{align*}
where the second inequality holds because of Lemma \ref{lemma1}. Therefore, $\|\hat{\Delta}-\Delta\|_2 \leq \frac{C_{\min}}{2}$ implies
\begin{align*}
\Lambda_{\min}(\hat{\Delta}_{A^*A^*}) = \frac{1}{\|\hat{\Delta}_{A^*A^*}^{-1}\|_2} \geq \frac{1}{2} C_{\min}.
\end{align*}
Thus,
\begin{align*}
    \mathbb{P}\left( \Lambda_{\min}(\hat{\Delta}_{A^*A^*})\geq \frac{1}{2} C_{\min}\right)
 \geq 1-\frac{4p^4\sigma_{\max}^4(\sigma_{\min}^4+K)}{N\xi^2\sigma_{\min}^4} ,
\end{align*}
which completes the proof.
\Halmos\endproof

\begin{lemma}\label{th:samplelasso2}
For the Lasso regression given by \eqref{equ: generallinearregression} and \eqref{equ: generallasso}, assume that conditions (i) and (ii) in Theorem \ref{th:samplelasso3} hold. Then, $$\mathbb{P}\left(\left\|\Hat{\Delta}_{A^{*c}A^{*}}\hat{\Delta}_{A^{*}A^{*}}^{-1} \right\|_{\infty} \leq 1-\frac{\gamma}{2} \right)\geq 1-\frac{4p^4\sigma_{\max}^4(\sigma_{\min}^4+K)}{N\xi^2\sigma_{\min}^4}$$ 
holds with $\xi = g_\Sigma^{-1}\left(\frac{\gamma}{\zeta(2+2\alpha\zeta+\gamma)}\right)>0$, and the definition of $g_{\Sigma}(\cdot)$ and other notations the same as in Theorem \ref{th:samplelasso3}. 
\end{lemma}

\proof{Proof of Lemma \ref{th:samplelasso2}.}
By Lemma \ref{lemmafund}, 
\begin{align*}
    \mathbb{P}\left(\|\hat{\Delta}-\Delta\|_\infty \leq \frac{\gamma}{\zeta(2+2\alpha\zeta+\gamma)}\right) &\geq \mathbb{P}\left(\|\hat{\Sigma}-\Sigma\|_\infty \leq g_\Sigma^{-1}\left(\frac{\gamma}{\zeta(2+2\alpha\zeta+\gamma)}\right)\right)\\
    &\geq \mathbb{P}\left(\|\hat{\Sigma}-\Sigma\|_\infty \leq \xi\right)\geq
    1-\frac{4p^4\sigma_{\max}^4(\sigma_{\min}^4+K)}{N\xi^2\sigma_{\min}^4},
\end{align*}
where the last inequality holds by \eqref{equ:prob_2norm_bound}.
Whenever $\|\hat{\Delta}-\Delta\|_\infty \leq \frac{\gamma}{\zeta(2+2\alpha\zeta+\gamma)} $ holds, we have
\begin{align*}
    \|\hat{\Delta}_{A^{*c}A^{*}} - \Delta_{A^{*c}A^{*}}\|_\infty &\leq \|\hat{\Delta}-\Delta\|_\infty \leq  \frac{\gamma}{\zeta(2+2\alpha\zeta+\gamma)},\\
    \|\hat{\Delta}_{A^{*}A^{*}} - \Delta_{A^{*}A^{*}}\|_\infty &\leq \|\hat{\Delta}-\Delta\|_\infty \leq  \frac{\gamma}{\zeta(2+2\alpha\zeta+\gamma)},\\
    \|\hat{\Delta}_{A^{*c}A^{*}}\|_\infty &\leq
    \|\Delta_{A^{*c}A^{*}}\|_\infty + \|\hat{\Delta}_{A^{*c}A^{*}} - \Delta_{A^{*c}A^{*}}\|_\infty  \leq \alpha + \frac{\gamma}{\zeta(2+2\alpha\zeta+\gamma)},
\end{align*}
and 
\begin{align*}
    \|I-\Delta_{A^{*}A^{*}}^{-1}\hat{\Delta}_{A^{*}A^{*}}\|_\infty \leq \|\Delta_{A^{*}A^{*}}^{-1}\|_\infty \cdot \|\Delta_{A^{*}A^{*}}-\hat{\Delta}_{A^{*}A^{*}}\|_\infty &= \zeta \cdot \|\Delta_{A^{*}A^{*}}-\hat{\Delta}_{A^{*}A^{*}}\|_\infty \\&\leq \zeta \cdot \|\hat{\Delta}-\Delta\|_\infty \leq \frac{\gamma}{2+2\alpha\zeta+\gamma} < 1.
\end{align*}
Therefore, applying Lemma \ref{lemma1} yields
\begin{equation*}
\|\hat{\Delta}_{A^{*}A^{*}}^{-1} - \Delta_{A^{*}A^{*}}^{-1}\|_\infty \leq \frac{\|\Delta_{A^{*}A^{*}}^{-1}\|_\infty^2 \cdot \|\hat{\Delta}_{A^{*}A^{*}} - \Delta_{A^{*}A^{*}}\|_\infty}{ 1-\|\Delta_{A^{*}A^{*}}^{-1}\|_\infty \cdot \|\hat{\Delta}_{A^{*}A^{*}} - \Delta_{A^{*}A^{*}}\|_\infty} \leq \frac{\zeta^2 \cdot \frac{\gamma}{\zeta(2+2\alpha\zeta+\gamma)}}{1-\zeta \cdot \frac{\gamma}{\zeta(2+2\alpha\zeta+\gamma)}} = \frac{\gamma\zeta}{2+2\alpha\zeta},
\end{equation*}
which further implies that\begin{equation*}%\label{eq:inversedelta}
\|\hat{\Delta}_{A^{*}A^{*}}^{-1}\|_\infty \leq \| \Delta_{A^{*}A^{*}}^{-1}\|_\infty + \|\hat{\Delta}_{A^{*}A^{*}}^{-1} - \Delta_{A^{*}A^{*}}^{-1}\|_\infty \leq \zeta + \frac{\gamma\zeta}{2+2\alpha\zeta} = \frac{\zeta(2+2\alpha\zeta+\gamma)}{2+2\alpha\zeta}.
\end{equation*}
Hence, $\|\hat{\Delta}-\Delta\|_\infty \leq \frac{\gamma}{\zeta(2+2\alpha\zeta+\gamma)} $ implies that
\begin{align*}
\left\|\hat{\Delta}_{A^{*c}A^{*}}\hat{\Delta}_{A^{*}A^{*}}^{-1} \right\|_{\infty} \leq &\left\|\hat{\Delta}_{A^{*c}A^{*}}\hat{\Delta}_{A^{*}A^{*}}^{-1} - \hat{\Delta}_{A^{*c}A^{*}}\Delta_{A^{*}A^{*}}^{-1}\right\|_{\infty}+ \\& \left\|\hat{\Delta}_{A^{*c}A^{*}}\Delta_{A^{*}A^{*}}^{-1} - \Delta_{A^{*c}A^{*}}\Delta_{A^{*}A^{*}}^{-1}\right\|_{\infty} + \left\| \Delta_{A^{*c}A^{*}}\Delta_{A^{*}A^{*}}^{-1}\right\|_{\infty} \\\leq &\|\hat{\Delta}_{A^{*c}A^{*}}\|_\infty \cdot \|\hat{\Delta}_{A^{*}A^{*}}^{-1} - \Delta_{A^{*}A^{*}}^{-1}\|_\infty + \\&
\|\Delta_{A^{*}A^{*}}^{-1}\|_\infty \cdot \|\hat{\Delta}_{A^{*c}A^{*}} - \Delta_{A^{*c}A^{*}}\|_\infty + 1-\gamma \\\leq &\left(\alpha + \frac{\gamma}{\zeta(2+2\alpha\zeta+\gamma)}\right) \cdot \frac{\gamma\zeta}{2+2\alpha\zeta} + \zeta \cdot \frac{\gamma}{\zeta(2+2\alpha\zeta+\gamma)} +  1-\gamma 
\\= &1-\frac{\gamma}{2}.
\end{align*}
Therefore,
\begin{align*}
    \mathbb{P}\left(\left\|\hat{\Delta}_{A^{*c}A^{*}}\hat{\Delta}_{A^{*}A^{*}}^{-1} \right\|_{\infty} \leq 1-\frac{\gamma}{2} \right) \geq 1-\frac{4p^4\sigma_{\max}^4(\sigma_{\min}^4+K)}{N\xi^2\sigma_{\min}^4}.\tag*{\Halmos}
\end{align*}
\endproof

\subsection{Proofs}
\proof{Proof of Theorem \ref{thm:uniqueness}.}
For any $\theta>0$,
\begin{align}
    \mathbb{P}(\left|L_a - L_b\right|>\eta) \geq& \mathbb{P}\left(\left|\sum_{i=1}^p c_iS_i\right| > \eta , \|S\|_2 < \theta\sqrt{p\|\Sigma\|_2} \right)\nonumber \\= &\mathbb{P}\left(\left|\sum_{i=1}^p c_iS_i\right| > \eta \Bigl| \|S\|_2 < \theta\sqrt{p\|\Sigma\|_2} \right) \cdot \mathbb{P} \left(\|S\|_2 < \theta\sqrt{p\|\Sigma\|_2} \right). \label{equ:proof_thm_2_cond}
\end{align}
By Markov's inequality, 
\begin{align}
    \mathbb{P}\left(\|S\|_2 \leq \theta\sqrt{p\|\Sigma\|_2} \right) &\geq 1- 
    \mathbb{P}\left(\|S\|_2 > \theta\sqrt{p\|\Sigma\|_2} \right) \geq 1-\frac{\mathbb{E}\|S\|_2}{ \theta \sqrt{p\|\Sigma\|_2} } \nonumber\\
    & \geq 1- \frac{\sqrt{\mathbb{E}\|S\|_2^2}}{\theta \sqrt{p\|\Sigma\|_2}} = 1- \frac{\sqrt{\mathrm{tr}(\Sigma)}}{\theta \sqrt{p\|\Sigma\|_2}} \geq 1-\frac{\sqrt{p\|\Sigma\|_2 }}{\theta \sqrt{p\|\Sigma\|_2}} = 
    1-\frac{1}{\theta}.  \label{signaturebound}
\end{align}
In addition, applying Markov's inequality to $X = \theta\|C\|_\infty p \sqrt{\|\Sigma\|_2} - \left|\sum_{i=1}^p c_iS_i\right|$ for a sufficiently small $\eta>0$ yields
\begin{align*}
    \mathbb{P}\left(\left|\sum_{i=1}^p c_iS_i\right| \leq \eta \Bigl| \|S\|_2 \leq \theta\sqrt{p\|\Sigma\|_2} \right) = \mathbb{P} \left(X \geq \theta \|C\|_\infty p\sqrt{\|\Sigma\|_2} - \eta \Bigl| \|S\|_2 \leq \theta\sqrt{p\|\Sigma\|_2} \right) \\ \quad\leq \frac{\mathbb{E}\left[X\Bigl| \|S\|_2 \leq \theta\sqrt{p\|\Sigma\|_2}\right]}{\theta \|C\|_\infty p \sqrt{\|\Sigma\|_2} - \eta} 
    = \frac{\theta \|C\|_\infty p \sqrt{\|\Sigma\|_2} -  \mathbb{E} \left[\left|\sum_{i=1}^p c_iS_i\right| \Bigl| \|S\|_2 \leq \theta\sqrt{p\|\Sigma\|_2}\right]}{\theta \|C\|_\infty p \sqrt{\|\Sigma\|_2} - \eta}.
\end{align*}
Hence,
\begin{align}
    \mathbb{P}\left(\left|\sum_{i=1}^p c_iS_i\right| > \eta \Bigl| \|S\|_2 \leq \theta\sqrt{p\|\Sigma\|_2} \right) \geq \frac{  \mathbb{E} \left[\left|\sum_{i=1}^p c_iS_i\right| \Bigl| \|S\|_2 \leq \theta\sqrt{p\|\Sigma\|_2}\right] - \eta}{\theta \|C\|_\infty p \sqrt{\|\Sigma\|_2} - \eta} . \label{equ:proof_thm_2_equ} 
\end{align}
Under the condition of $\|S\|_2 \leq \theta\sqrt{p\|\Sigma\|_2 }$, 
\begin{equation*}%\label{sigcombupperbound}
    \left|\sum_{i=1}^p c_iS_i\right| \leq \|C\|_\infty \cdot \|S\|_1 \leq \|C\|_\infty \cdot \sqrt{p}\|S\|_2 \leq \theta \|C\|_\infty p \sqrt{\|\Sigma\|_2},
\end{equation*}
and by multiplying both sides of the above inequality by $\left|\sum_{i=1}^p c_iS_i\right|$  and taking expectations, we obtain
\begin{equation}\label{sigcombElow}
    \mathbb{E} \left[\left|\sum_{i=1}^p c_iS_i\right| \Bigl| \|S\|_2 \leq \theta\sqrt{p\|\Sigma\|_2}\right] \geq \frac{\mathbb{E} \left[ {\left(\sum_{i=1}^p c_iS_i\right)}^2 \Bigl| \|S\|_2 \leq \theta\sqrt{p\|\Sigma\|_2} \right]}{\theta \|C\|_\infty p \sqrt{\|\Sigma\|_2}}.
\end{equation} 
Thus, by combining \eqref{equ:proof_thm_2_cond}, \eqref{signaturebound}, \eqref{equ:proof_thm_2_equ}, and \eqref{sigcombElow}, we obtain
\begin{align*}
    \mathbb{P}(\left|L_a-L_b\right|>\eta) \geq \left(1-\frac{1}{\theta} \right) \cdot \frac{  \frac{\mathbb{E} \left[ {\left(\sum_{i=1}^p c_iS_i\right)}^2 \Bigl| \|S\|_2 \leq \theta\sqrt{p\|\Sigma\|_2} \right]}{\theta \|C\|_\infty p \sqrt{\|\Sigma\|_2}} - \eta}{\theta \|C\|_\infty p \sqrt{\|\Sigma\|_2} - \eta} .
\end{align*}
When $\theta$ is sufficiently large, $\mathbb{E} \left[ {\left(\sum_{i=1}^p c_iS_i\right)}^2 \Bigl| \|S\|_2 \leq \theta\sqrt{p\|\Sigma\|_2} \right] >0$ because the distribution of $S$ is non-degenerate. Therefore, \eqref{equ:unique1} holds. Furthermore, \eqref{equ:unique} is a direct result of the triangle inequality, which completes the proof.
\Halmos\endproof

\proof{Proof of Proposition \ref{prop: moment_signatures}.}
First, we have
\begin{equation} \label{equ: proof_prop1_exp}
\mathbb{E} \left[S(\mathbf{X})_{t}^{i_1,\dots,i_n,I} \right] = \mathbb{E}\left[ \int_{0}^t S(\mathbf{X})_{s}^{i_1,\dots,i_{n-1}} \mathrm{d} X_s^{i_n} \right] = 0
\end{equation}

Next we prove $\mathbb{E}\left[   S(\mathbf{X})_{t}^{i_1,\dots,i_n,I} S(\mathbf{X})_{t}^{j_1,\dots,j_m,I}  \right] = 0$ for $m\neq n$ by induction. Without loss of generality, we assume that $m>n$. When $n=1$, for any $m>1$, we have
\begin{align*}
    \mathbb{E}\left[   S(\mathbf{X})_{t}^{i_1,I} S(\mathbf{X})_{t}^{j_1,\dots,j_m,I}  \right] &= \mathbb{E}\left[  \left( \int_{0}^t \mathrm{d} X_s^{i_1}  \right) \left( \int_{0}^t S(\mathbf{X})_s^{j_1,\dots,j_{m-1}} \mathrm{d} X_s^{j_m}  \right)   \right]   \\
    &=  \int_{0}^t \mathbb{E}\left[S(\mathbf{X})_s^{j_1,\dots,j_{m-1},I} \right] \rho_{i_1 j_m} \sigma_{i_1} \sigma_{j_m} \mathrm{d} s = 0,
\end{align*}
where the second equality uses the It\^o isometry and the third equality uses \eqref{equ: proof_prop1_exp}. Now assume $\mathbb{E}\left[   S(\mathbf{X})_{t}^{i_1,\dots,i_n,I} S(\mathbf{X})_{t}^{j_1,\dots,j_m,I}  \right] = 0$ for any $m>n$. Then, 
\begin{align*}
    &\mathbb{E}\left[   S(\mathbf{X})_{t}^{i_1,\dots,i_{n+1},I} S(\mathbf{X})_{t}^{j_1,\dots,j_{m+1},I}  \right] \\
    =& \mathbb{E}\left[  \left( \int_{0}^t S(\mathbf{X})_{t}^{i_1,\dots,i_n,I} \mathrm{d} X_s^{i_{n+1}}  \right) \left( \int_{0}^t S(\mathbf{X})_s^{j_1,\dots,j_{m},I} \mathrm{d} X_s^{j_{m+1}}  \right)   \right]   \\
    =&  \int_{0}^t \mathbb{E}\left[S(\mathbf{X})_{t}^{i_1,\dots,i_k,I} S(\mathbf{X})_s^{j_1,\dots,j_{m},I} \right] \rho_{i_{n+1} j_{m+1}} \sigma_{i_{n+1}} \sigma_{j_{m+1}} \mathrm{d} s   = 0.
\end{align*}
This proves $\mathbb{E}\left[   S(\mathbf{X})_{t}^{i_1,\dots,i_n,I} S(\mathbf{X})_{t}^{j_1,\dots,j_m,I}  \right] = 0$.

We finally prove $\mathbb{E}\left[   S(\mathbf{X})_{t}^{i_1,\dots,i_n,I} S(\mathbf{X})_{t}^{j_1,\dots,j_n,I}  \right] = \frac{t^n}{n!} \prod_{k=1}^n \rho_{i_k j_k} \sigma_{i_k} \sigma_{j_k}$ by induction. When $n=1$, 
\begin{align*}
    &\mathbb{E}\left[   S(\mathbf{X})_{t}^{i_1,I} S(\mathbf{X})_{t}^{j_1,I}  \right] \\
    =& \mathbb{E}\left[  \left( \int_{0}^t \mathrm{d} X_s^{i_1}  \right) \left( \int_{0}^t  \mathrm{d} X_s^{j_1}  \right)   \right] = \int_{0}^t \rho_{i_1 j_1} \sigma_{i_1} \sigma_{j_1} \mathrm{d} s  = t \rho_{i_1 j_1} \sigma_{i_1} \sigma_{j_1} . 
\end{align*}
Now, assume $\mathbb{E}\left[   S(\mathbf{X})_{t}^{i_1,\dots,i_n,I} S(\mathbf{X})_{t}^{j_1,\dots,j_n,I}  \right] = \frac{t^n}{n!} \prod_{k=1}^n \rho_{i_k j_k} \sigma_{i_k} \sigma_{j_k}$, then 
\begin{align*}
    &\mathbb{E}\left[   S(\mathbf{X})_{t}^{i_1,\dots,i_{n+1},I} S(\mathbf{X})_{t}^{j_1,\dots,j_{n+1},I}  \right] \\
    =& \mathbb{E}\left[  \left( \int_{0}^t S(\mathbf{X})_{s}^{i_1,\dots,i_n,I} \mathrm{d} X_s^{i_{n+1}}  \right) \left( \int_{0}^t S(\mathbf{X})_s^{j_1,\dots,j_{n},I} \mathrm{d} X_s^{j_{n+1}}  \right)   \right]   \\
    =&  \int_{0}^t \mathbb{E} \left[S(\mathbf{X})_{s}^{i_1,\dots,i_n,I} S(\mathbf{X})_s^{j_1,\dots,j_{n},I} \right] \rho_{i_{n+1} j_{n+1}} \sigma_{i_{n+1}} \sigma_{j_{n+1}} \mathrm{d} s  \\
    =& \int_{0}^t \left( \frac{s^n}{n!} \prod_{k=1}^n \rho_{i_k j_k} \sigma_{i_k} \sigma_{j_k} \right) \rho_{i_{n+1} j_{n+1}} \sigma_{i_{n+1}} \sigma_{j_{n+1}}  \mathrm{d} s = \frac{t^{n+1}}{(n+1)!} \prod_{k=1}^{n+1} \rho_{i_k j_k} \sigma_{i_k} \sigma_{j_k}. 
\end{align*}
Therefore, $\mathbb{E}\left[   S(\mathbf{X})_{t}^{i_1,\dots,i_n,I} S(\mathbf{X})_{t}^{j_1,\dots,j_n,I}  \right] = \frac{t^n}{n!} \prod_{k=1}^n \rho_{i_k j_k} \sigma_{i_k} \sigma_{j_k}$.  
\Halmos\endproof

\proof{Proof of Theorem \ref{prop: structure_coef}.}
    By Proposition \ref{prop: moment_signatures}, for any $n$, 
    \begin{equation*}
        \frac{\mathbb{E}\left[   S(\mathbf{X})_{t}^{i_1,\dots,i_n,I} S(\mathbf{X})_{t}^{j_1,\dots,j_n,I}  \right]  }{ \sqrt{ \mathbb{E}\left[   S(\mathbf{X})_{t}^{i_1,\dots,i_n,I} \right]^2 \mathbb{E}\left[ S(\mathbf{X})_{t}^{j_1,\dots,j_n,I}  \right]^2  } } = \frac{ \frac{t^n}{n!} \prod_{k=1}^n \rho_{i_k j_k} \sigma_{i_k} \sigma_{j_k} }{ \sqrt{ \frac{t^n}{n!} \prod_{k=1}^n \sigma_{i_k} \sigma_{i_k} \cdot \frac{t^n}{n!} \prod_{k=1}^n  \sigma_{j_k} \sigma_{j_k} }  } = \prod_{k=1}^n \rho_{i_k j_k},
    \end{equation*}
    implying 
    \begin{equation*}
        \frac{\mathbb{E}\left[   S(\mathbf{X})_{t}^{i_1,I} S(\mathbf{X})_{t}^{j_1,I}  \right]  }{ \sqrt{ \mathbb{E}\left[   S(\mathbf{X})_{t}^{i_1,I} \right]^2 \mathbb{E}\left[ S(\mathbf{X})_{t}^{j_1,I}  \right]^2  } } = \rho_{i_1 j_1}
    \end{equation*}
    and 
    \begin{equation*}
        \frac{\mathbb{E}\left[   S(\mathbf{X})_{t}^{i_1,\dots,i_n,I} S(\mathbf{X})_{t}^{j_1,\dots,j_n,I}  \right]  }{ \sqrt{ \mathbb{E}\left[   S(\mathbf{X})_{t}^{i_1,\dots,i_n,I} \right]^2 \mathbb{E}\left[ S(\mathbf{X})_{t}^{j_1,\dots,j_n,I}  \right]^2  } } = \rho_{i_n j_n} \cdot \frac{\mathbb{E}\left[   S(\mathbf{X})_{t}^{i_1,\dots,i_{n-1},I} S(\mathbf{X})_{t}^{j_1,\dots,j_{n-1},I}  \right]  }{ \sqrt{ \mathbb{E}\left[   S(\mathbf{X})_{t}^{i_1,\dots,i_{n-1},I} \right]^2 \mathbb{E}\left[ S(\mathbf{X})_{t}^{j_1,\dots,j_{n-1},I}  \right]^2  } }.
    \end{equation*}
    This proves the Kronecker product structure given by \eqref{equ: Omega_k}. 

Proposition \ref{prop: moment_signatures} also implies that, for any $m \neq n$,
\begin{equation*}
        \frac{\mathbb{E}\left[   S(\mathbf{X})_{t}^{i_1,\dots,i_n,I} S(\mathbf{X})_{t}^{j_1,\dots,j_m,I}  \right]  }{ \sqrt{ \mathbb{E}\left[   S(\mathbf{X})_{t}^{i_1,\dots,i_n,I} \right]^2 \mathbb{E}\left[ S(\mathbf{X})_{t}^{j_1,\dots,j_m,I}  \right]^2  } } = 0.
\end{equation*}
This proves that It\^o signatures of different orders are uncorrelated and, therefore, the correlation matrix is block diagonal.
\Halmos\endproof

\proof{Proof of Proposition \ref{prop: moment_signatures_S_integral}.}
Equations
\begin{equation*}
    \mathbb{E} \left[S(\mathbf{X})_{t}^{i_1,\dots,i_{2n-1},S}\right] = 0
\end{equation*}
and
\begin{equation*}
\mathbb{E} \left[S(\mathbf{X})_{t}^{i_1,\dots,i_{2n},S} S(\mathbf{X})_{t}^{j_1,\dots,j_{2m-1},S}\right] = 0     
\end{equation*}
can be proven using a similar approach to the proof of Theorem \ref{prop: structure_coef_S_integral}. Now we prove
\begin{equation}\label{equ:proof_prop2}
\mathbb{E} \left[S(\mathbf{X})_{t}^{i_1,\dots,i_{2n},S}\right] = \frac{1}{2^n} \frac{t^{n}}{n!} \prod_{k=1}^{n} \rho_{i_{2k-1}i_{2k}} \prod_{k=1}^{2n} \sigma_{i_k}    
\end{equation}
by induction. If $n=0$, \eqref{equ:proof_prop2} holds because of \eqref{equ: propTEMP_ini1} in Proposition \ref{prop: moment_signatures_S_integral_TEMP}. Now we assume that \eqref{equ:proof_prop2} holds for $n=j$. Then, when $n=j+1$, by Proposition \ref{prop: moment_signatures_S_integral_TEMP}, 
\begin{align*}
\mathbb{E} \left[S(\mathbf{X})_{t}^{i_1,\dots,i_{2(j+1)},S}\right] =& \frac{1}{2} \rho_{i_{2j+1}i_{2j+2}} \sigma_{i_{2j+1}}\sigma_{i_{2j+2}} \int_0^t \frac{1}{2^j}  \frac{s^{j}}{j!} \prod_{k=1}^{j} \rho_{i_{2k-1}i_{2k}} \prod_{k=1}^{2j} \sigma_{i_k} \mathrm{d} s \\
=&    \frac{1}{2^{j+1}} \frac{t^{j+1}}{(j+1)!} \prod_{k=1}^{j+1} \rho_{i_{2k-1}i_{2k}} \prod_{k=1}^{2(j+1)} \sigma_{i_k} .
\end{align*}
Therefore, \eqref{equ:proof_prop2} holds.
\Halmos\endproof

\proof{Proof of Theorem \ref{prop: structure_coef_S_integral}.}
    For the Stratonovich signature of a Brownian motion, this is a direct corollary of Proposition \ref{prop: moment_signatures_S_integral}. For both the It\^o and Stratonovich signatures of an OU process, we only need to prove that, for an odd number $m$ and an even number $n$, we have
\begin{equation*}
    \mathbb{E} \left[ S(\mathbf{X})_t^{i_1,\dots,i_m} S(\mathbf{X})_t^{j_1,\dots,j_n} \right] = 0
\end{equation*}
for any $i_1, \dots, i_m$ and $j_1,\dots,j_n$ taking values in $\{1,2,\dots,d\}$. Here the signatures can be defined in the sense of either It\^o or Stratonovich. 

Consider the reflected OU process, $ \check{\mathbf{X}}_t = -\mathbf{X}_t$. By definition, $ \check{\mathbf{X}}_t$ is also an OU process with the same mean reversion parameter. Therefore, the signatures of $ \check{\mathbf{X}}_t$ and $ {\mathbf{X}}_t$ should have the same distribution. In particular, we have
\begin{equation} \label{equ: proof_alternating_reflect}
    \mathbb{E} \left[ S(\check{\mathbf{X}})_t^{i_1,\dots,i_m} S(\check{\mathbf{X}})_t^{j_1,\dots,j_n} \right] = \mathbb{E} \left[ S(\mathbf{X})_t^{i_1,\dots,i_m} S({\mathbf{X}})_t^{j_1,\dots,j_n} \right] .
\end{equation}
Now we consider the definition of the signature
\begin{equation*}
    S(\mathbf{X})_t^{i_1,\dots,i_m} = \int_{0<t_1<\dots<t_m<t}  \mathrm{d} X_{t_1}^{i_1} \cdots \mathrm{d} X_{t_k}^{i_m},
\end{equation*}
where the integral can be defined in the sense of either It\^o or Stratonovich. We therefore have
\begin{align*}
    S(\check{\mathbf{X}})_t^{i_1,\dots,i_m} = S(-{\mathbf{X}})_t^{i_1,\dots,i_m} &= \int_{0<t_1<\dots<t_m<t}  \mathrm{d} (-X_{t_1}^{i_1}) \cdots \mathrm{d} (-X_{t_k}^{i_m}) \\
    &= (-1)^m \int_{0<t_1<\dots<t_m<t}  \mathrm{d} X_{t_1}^{i_1} \cdots \mathrm{d} X_{t_k}^{i_m} = (-1)^m S(\mathbf{X})_t^{i_1,\dots,i_m}.
\end{align*}
Similarly, we have
\begin{equation*}
    S(\check{\mathbf{X}})_t^{j_1,\dots,j_n} = (-1)^n S(\mathbf{X})_t^{j_1,\dots,j_n}.
\end{equation*}
Therefore, 
\begin{align*}\label{equ: proof_alternating_reflectdef}
        &= -\mathbb{E} \left[ S(\mathbf{X})_t^{i_1,\dots,i_m} S({\mathbf{X}})_t^{j_1,\dots,j_n} \right],
\end{align*}
and combining this with \eqref{equ: proof_alternating_reflect} leads to the result. 
\Halmos\endproof

\proof{Proof of Theorem \ref{prop: sufficient_irrepresentable}.}
    Note that, for a block diagonal correlation matrix $\Delta$, the irrepresentable conditions given by Definition \ref{def: irrepresentable_model} hold if and only if they hold for each block. Thus, the first necessary and sufficient condition for the irrepresentable conditions holds due to Theorem \ref{prop: structure_coef}. The second sufficient condition holds due to \eqref{equ:gamma_range} in the proof of Theorem \ref{th:sampleito} because $\rho < \frac{1}{2q_{\max}-1}$ implies $\gamma > 0$. This completes the proof.
\Halmos\endproof

\proof{Proof of Theorem \ref{prop: irrepresentable_equiv}.}
    Note that, for a block diagonal correlation matrix $\Delta$, the irrepresentable conditions given by Definition \ref{def: irrepresentable_model} hold if and only if they hold for each block. Thus, this result holds because of Theorem \ref{prop: structure_coef_S_integral}. 
\Halmos\endproof

\proof{Proof of Theorem \ref{th:sampleito}.}
%We use the notation as stated in the theorem. 
We use Theorem \ref{th:samplelasso3} to obtain the result. Lemma \ref{le:signaturebound} implies that the finite fourth-moment condition for the It\^o signature of Brownian motion holds. 
By Theorem \ref{prop: structure_coef}, the correlation matrix of the It\^o signature of Brownian motion exhibits a block-diagonal structure
$$\Delta^1 = \mathrm{diag}\{ \Omega_0, \Omega_1, \Omega_2,\dots,\Omega_K \},$$
whose diagonal blocks $\Omega_k$ 
are given by
$$ 
\Omega_k = \underbrace{ \Omega \otimes \Omega \otimes \cdots \otimes \Omega }_{k}, \quad k =1,2,\dots, K,
$$
and $\Omega_0 = 1$. 
Because $\rho = \max_{i\neq j} |\rho_{ij}|$, for any $k=1,2,\dots,K$, we have $\max_{i \neq j} \left\{|\Omega_{k,ij} |\right\}\leq \rho$, where $\Omega_{k,ij}$ is the $(i,j)$-entry of $\Omega_k$. Hence, 
\begin{equation}\label{eq:alpha}
    \|\Omega_{k,A^{*c}_k A^*_k}\|_\infty \leq \#A^*_k \cdot \rho\leq q_{\max}\rho
\end{equation}
and
\begin{equation}\label{eq:cmin}
    \|\Omega_{k,A^*_k A^*_k}\|_2 \leq \sqrt{\#A^*_k} \cdot \|\Omega_{k,A^*_k A^*_k}\|_\infty \leq \sqrt{\#A^*_k} \cdot \left(1+(\#A^*_k-1)\rho\right) \leq \sqrt{q_{\max}} \left(1+(q_{\max}-1)\rho\right).
\end{equation}
Let $X = (X_1,\dots,X_{\#A^*_k})^\top \in \mathbb{R}^{\#A^*_k}$ be any vector of constants satisfying $\|X\|_\infty = 1$. Without loss of generality, we assume $X_1 = 1$. Therefore,
\begin{align*}
    \|\Omega_{k,A^*A^*}X\|_\infty &\geq |(\Omega_{k,A^*A^*})_{1,1}X_1 + \cdots + (\Omega_{k,A^*A^*})_{1,{\#A^*_k}}X_{\#A^*_k}| \\&=
    |1 + (\Omega_{k,A^*A^*})_{1,2}X_2+\cdots + (\Omega_{k,A^*A^*})_{1,{\#A^*_k}} X_{\#A^*_k}| \\&\geq
    1 - |(\Omega_{k,A^*A^*})_{1,2} X_2| - \cdots -|(\Omega_{k,A^*A^*})_{1,{\#A^*_k}} X_{\#A^*_k}| \\&\geq 1-(\#A^*_k-1)\rho \geq 1-(q_{\max}-1)\rho,
\end{align*}
which implies that
\begin{align}
    \|\Omega_{k,A^*A^*}^{-1}\|_\infty &= \frac{1}{\min_{\|X\|_\infty=1}\|\Omega_{k,A^*A^*}X\|_\infty} \leq \frac{1}{1-(q_{\max}-1)\rho},\label{eq:zeta}\\
\|\Omega_{k,A^{*c}A^*}\Omega_{k,A^*A^*}^{-1}\|_\infty &\leq \|\Omega_{k,A^{*c}A^*}\|_\infty \cdot \|\Omega_{k,A^*A^*}^{-1}\|_\infty \leq \frac{q_{\max}\rho}{1-(q_{\max}-1)\rho}.\label{eq:gamma}
\end{align}
Equations \eqref{eq:alpha}, \eqref{eq:cmin},  \eqref{eq:zeta}, and \eqref{eq:gamma} lead to the parameters for Theorem \ref{th:samplelasso3} given by
\begin{align}
\alpha &= \left\|\Delta_{A^{*c}A^{*}} \right\|_{\infty} = \max_{1\leq k\leq K}  \left\|\Omega_{k,A^{*c}A^{*}}\right\|_{\infty}  \leq q_{\max}\rho, \nonumber\\
\zeta &= \left\|\Delta_{A^{*}A^{*}}^{-1}\right\|_{\infty} = \max_{1\leq k\leq K}   \left\|\Omega_{k,A^{*}A^{*}}^{-1}\right\|_{\infty} \leq \frac{1}{1-(q_{\max}-1)\rho}, \nonumber\\
C_{\min} &= \Lambda_{\min}(\Delta_{A^{*}A^{*}}) = \frac{1}{\|\Delta_{A^* A^*}^{-1}\|_2}  = \frac{1}{\max_{1\leq k\leq K} {\|\Omega_{k,A^* A^*}^{-1}\|_2}}\nonumber\\
&\qquad \qquad \qquad \qquad \qquad \qquad \geq \frac{1}{\max_{1\leq k\leq K} \sqrt{q_{\max}}{\|\Omega_{k,A^* A^*}^{-1}\|_\infty}} \geq \frac{ 1-(q_{\max}-1)\rho }{\sqrt{q_{\max}} }, \nonumber\\
\gamma &= \min_{1\leq k\leq K} \left\{1-\left\|\Omega_{k,A^{*c}A^{*}}\Omega_{k,A^{*}A^{*}}^{-1}\right\|_{\infty} \right\} \geq \frac{1-(2q_{\max}-1)\rho}{1-(q_{\max}-1)\rho}.  \label{equ:gamma_range}
\end{align}
Plugging these into Theorem \ref{th:samplelasso3} leads to the result.  
\Halmos\endproof

\proof{Proof of Theorem \ref{th:samplestratonovich}.}
Theorem \ref{prop: structure_coef_S_integral} implies that the correlation structure can be represented by $\mathrm{diag} \{ \Psi_{\mathrm{odd}}, \Psi_{\mathrm{even}} \}$. Lemma \ref{le:signaturebound} implies that the finite fourth-moment condition for Stratonovich signature of Brownian motion holds, while Lemma \ref{le:signaturebound} implies that the finite fourth-moment condition for both It\^o and Stratonovich signature of OU process holds. Combining these with Theorem \ref{th:samplelasso3} leads to the result.  
\Halmos\endproof

\proof{Proof of Proposition \ref{prop: moment_signatures_S_integral_TEMP}.}
   For any $l,t \geq 0$ and $m,n=0,1,\dots$, define
    \begin{align*}
    f_{n,m}(l,t) &:= \mathbb{E}\left[ S(\mathbf{X})_l^{i_1,\dots,i_n,S}  S(\mathbf{X})_t^{j_1,\dots,j_{m},S} \right], \\
    g_{n,m}(l,t) &:= \mathbb{E}\left[ S(\mathbf{X})_l^{i_1,\dots,i_n,S} \int_0^t S(\mathbf{X})_s^{j_1,\dots,j_{m-1},S} \mathrm{d} X_s^{j_m} \right].
    \end{align*}
    Then, by \eqref{eq:SandIto} in the proof of Lemma \ref{le:signaturebound} and Fubini's theorem,
    \begin{align*}
        &f_{n,m}(l,t) = \mathbb{E}\left[ S(\mathbf{X})^{i_1,\dots,i_{n},S}_l S(\mathbf{X})^{j_1,\dots,j_{m},S}_t \right] \\
        =& \mathbb{E}\left[ S(\mathbf{X})^{i_1,\dots,i_{n},S}_l \left(
        \int_0^t S(\mathbf{X})_s^{j_1,\dots,j_{m-1},S} 
         \mathrm{d} X_s^{j_m} + \frac{1}{2} \rho_{j_{m-1}j_{m}} \sigma_{j_{m-1}} \sigma_{j_{m}} \int_0^t S(\mathbf{X})_s^{j_1,\dots,j_{m-2},S} 
         \mathrm{d} s
        \right) \right] \\
        =& g_{n,m}(l,t) + \frac{1}{2} \rho_{j_{m-1}j_{m}} \sigma_{j_{m-1}} \sigma_{j_{m}} \mathbb{E}  \left[S(\mathbf{X})^{i_1,\dots,i_{n},S}_l \int_0^t S(\mathbf{X})_s^{j_1,\dots,j_{m-2},S} 
         \mathrm{d} s \right]\\
        =&g_{n,m}(l,t) + \frac{1}{2} \rho_{j_{m-1}j_{m}} \sigma_{j_{m-1}} \sigma_{j_{m}} \int_0^t \mathbb{E}  \left[S(\mathbf{X})^{i_1,\dots,i_{n},S}_l S(\mathbf{X})_s^{j_1,\dots,j_{m-2},S} \right]
         \mathrm{d} s \\
         =& g_{n,m}(l,t) + \frac{1}{2} \rho_{j_{m-1}j_{m}} \sigma_{j_{m-1}} \sigma_{j_{m}} \int_0^t f_{n,m-2}(l,s) \mathrm{d}s.
    \end{align*}
    This proves \eqref{equ: propTEMP_1} and \eqref{equ: propTEMP_3}.
    In addition, by It\^o isometry and Fubini's theorem,
    \begin{align*}
        & g_{n,m}(l,t) = \mathbb{E}\left[ S(\mathbf{X})_l^{i_1,\dots,i_n,S} \int_0^t S(\mathbf{X})_s^{j_1,\dots,j_{m-1},S} \mathrm{d} X_s^{j_m} \right] \\
        =& \mathbb{E}\Bigg[ \left(  \int_0^l S(\mathbf{X})_s^{i_1,\dots,i_{n-1},S} 
         \mathrm{d} X_s^{i_n} + \frac{1}{2} \rho_{i_{n-1}i_{n}} \sigma_{i_{n-1}} \sigma_{i_{n}} \int_0^l S(\mathbf{X})_s^{i_1,\dots,i_{n-2},S} 
         \mathrm{d} s \right) \\
         &\qquad\qquad\qquad\qquad\qquad\qquad\qquad\qquad\qquad\qquad\qquad\qquad\qquad \cdot \int_0^t S(\mathbf{X})_s^{j_1,\dots,j_{m-1},S} \mathrm{d} X_s^{j_m} \Bigg] \\
         =&\mathbb{E}\left[  \int_0^l S(\mathbf{X})_s^{i_1,\dots,i_{n-1},S} 
         \mathrm{d} X_s^{i_n} \int_0^t S(\mathbf{X})_s^{j_1,\dots,j_{m-1},S} \mathrm{d} X_s^{j_m} \right]\\
         &\qquad\qquad +  \frac{1}{2}  \rho_{i_{n-1}i_{n}} \sigma_{i_{n-1}} \sigma_{i_{n}} \mathbb{E} \left[ \int_0^l S(\mathbf{X})_s^{i_1,\dots,i_{n-2},S} \mathrm{d} s  \int_0^t S(\mathbf{X})_s^{j_1,\dots,j_{m-1},S} \mathrm{d} X_s^{j_m} \right] \\
         =& \rho_{i_nj_m}\sigma_{i_n} \sigma_{j_m} \int_0^{l \wedge t} \mathbb{E} \left[ S(\mathbf{X})_s^{i_1,\dots,i_{n-1},S}  S(\mathbf{X})_s^{j_1,\dots,j_{m-1},S}\right] \mathrm{d}s \\
         & \qquad\qquad +\frac{1}{2}  \rho_{i_{n-1}i_{n}} \sigma_{i_{n-1}} \sigma_{i_{n}}  \int_0^l \mathbb{E} \left[ S(\mathbf{X})_s^{i_1,\dots,i_{n-2},S}  \int_0^t S(\mathbf{X})_u^{j_1,\dots,j_{m-1},S} \mathrm{d} X_u^{j_m} \right] \mathrm{d} s \\
         =& \rho_{i_nj_m}\sigma_{i_n} \sigma_{j_m} \int_0^{l \wedge t} f_{n-1,m-1}(s,s) \mathrm{d}s + \frac{1}{2}  \rho_{i_{n-1}i_{n}} \sigma_{i_{n-1}} \sigma_{i_{n}}  \int_0^l g_{n-2,m}(s,t) \mathrm{d} s .
    \end{align*}
    This proves \eqref{equ: propTEMP_2} and \eqref{equ: propTEMP_4}.

    Now we prove the initial conditions. First, \eqref{equ: propTEMP_ini1} follows from the definition of 0-th order of signature. Second, \eqref{equ: propTEMP_ini2} follows from the property of It\^o integral
    \begin{equation*}
     g_{0,2m}(l,t)= \mathbb{E} \left[\int_0^t S(\mathbf{X})_s^{j_1,\dots,j_{2m-1},S} \mathrm{d} X_s^{j_{2m}} \right]= 0 .
    \end{equation*}
    Third,
    \begin{align*}
        f_{1,1}(l,t) &= \mathbb{E} \left[S(\mathbf{X})_l^{i_1,S} S(\mathbf{X})_t^{j_1,S} \right] = \mathbb{E}\left[\int_0^l 1\circ \mathrm{d} X_s^{i_1} \int_0^t 1\circ \mathrm{d} X_s^{j_1} \right] \\
        &= \mathbb{E}\left[   X_l^{i_1}  X_t^{j_1} \right] = \rho_{i_1 j_1} \sigma_{i_1} \sigma_{j_1} (l \wedge t),
    \end{align*}
    which proves \eqref{equ: propTEMP_ini3}. Fourth, by It\^o isometry,
    \begin{align*}
        g_{1,2m-1}(l,t) &= \mathbb{E} \left[S(\mathbf{X})_l^{i_1,S} \int_0^t S(\mathbf{X})_s^{j_1,\dots,j_{2m-2},S} \mathrm{d} X_s^{j_{2m-1}} \right] \\
        &= \mathbb{E} \left[\int_0^l 1\circ \mathrm{d} X_s^{i_1} \int_0^t S(\mathbf{X})_s^{j_1,\dots,j_{2m-2},S} \mathrm{d} X_s^{j_{2m-1}} \right] \\
        & = \mathbb{E} \left[\int_0^l \mathrm{d} X_s^{i_1} \int_0^t S(\mathbf{X})_s^{j_1,\dots,j_{2m-2},S} \mathrm{d} X_s^{j_{2m-1}} \right] \\
        & = \int_0^{l\wedge t} \mathbb{E} \left[ S(\mathbf{X})_s^{j_1,\dots,j_{2m-2},S} \right] \rho_{i_1 j_{2m-1}} \sigma_{i_1} \sigma_{j_{2m-1}} \mathrm{d} s \\
        &=  \rho_{i_1 j_{2m-1}} \sigma_{i_1} \sigma_{j_{2m-1}} \int_0^{l\wedge t} f_{0,2m-2}(s,s)  \mathrm{d} s.
    \end{align*}
In addition, by using \eqref{equ: propTEMP_1} recursively, we can obtain that 
\begin{equation*}
f_{0,2m-2}(s,s) = \frac{1}{2^{m-1}} \frac{s^{m-1}}{(m-1)!} \prod_{k=1}^{m-1} \rho_{j_{2k-1}j_{2k}} \prod_{k=1}^{2m-2} \sigma_{j_k}.    
\end{equation*}
Therefore, 
    \begin{equation*}
        g_{1,2m-1}(l,t) = \rho_{i_1 j_{2m-1}} \frac{1}{2^{m-1}} \frac{(l \wedge t)^{m-1}}{(m-1)!} \sigma_{i_1} \prod_{k=1}^{2m-1} \sigma_{j_k} \prod_{k=1}^{m-1} \rho_{j_{2k-1}j_{2k}} ,
    \end{equation*}
    which proves \eqref{equ: propTEMP_ini4}.
\Halmos\endproof

\proof{Proof of Example \ref{exmp: 1D_OU}.}
The solution to stochastic differential equation \eqref{equ:Exmp_OU} can be explicitly expressed as
\begin{equation*} \label{equ: OU_process_solution}
    Y_t = \int_0^t e^{-\kappa (t-s)}\mathrm{d}W_s, \quad t\geq 0,
\end{equation*}
where $W_t$ is a standard Brownian motion. Therefore, by It\^o isometry, $Y_t$ is a Gaussian random variable with zero mean and 
\begin{equation*}
    \mathrm{Var}(Y_t) =\mathbb{E} \left[Y_t^2 \right]= \mathbb{E} \left[ \int_0^t e^{-\kappa (t-s)}\mathrm{d}W_s \right]^2 = \int_0^t \left[ e^{-\kappa (t-s)}\right]^2\mathrm{d}s = \frac{1-e^{-2\kappa t}}{2\kappa}.
\end{equation*}
Now we calculate the correlation coefficient for its It\^o and Stratonovich signature, respectively. 
\paragraph{It\^o Signature.} By the definition of signature and \eqref{equ:Exmp_OU}, 
\begin{align}
    \mathbb{E} \left[S(\mathbf{X})_T^{1,1,I}\right] = \mathbb{E} \left[ \int_0^T Y_t \mathrm{d} Y_t \right] &= - \kappa \mathbb{E} \left[ \int_0^T Y_t^2 \mathrm{d} t \right] + \mathbb{E} \left[ \int_0^T Y_t \mathrm{d} W_t \right] = - \kappa   \int_0^T \mathbb{E} \left[Y_t^2 \right] \mathrm{d} t \nonumber\\ 
    &= - \kappa \int_0^T \frac{1-e^{-2\kappa t}}{2 \kappa} \mathrm{d}t = -\frac{T}{2} + \frac{ 1-e^{-2\kappa T} }{4\kappa}. \label{equ: EXMP_ITO_firstmoment}
\end{align}
For the second moment, by It\^o isometry, 
\begin{align*}
    \mathbb{E}\left[S(\mathbf{X})_T^{1,1,I} \right]^2 &= \mathbb{E} \left[ \int_0^T Y_t \mathrm{d} Y_t \right]^2 = \mathbb{E} \left[ -\kappa \int_0^T Y_t^2 \mathrm{d} t + \int_0^T Y_t \mathrm{d} W_t  \right]^2 \\
    &= \kappa^2 \int_0^T \int_0^T \mathbb{E} \left[Y_t^2 Y_s^2 \right] \mathrm{d} t \mathrm{d} s - 2 \kappa \mathbb{E} \left[ \int_0^T Y_t^2 \mathrm{d} t \int_0^T Y_t \mathrm{d} W_t \right] + \int_0^T \mathbb{E} \left[Y_t^2\right] \mathrm{d} t \\
    &=: \text{(a)} - \text{(b)} + \text{(c)}.
\end{align*}
It is easy to calculate Term (c):
\begin{equation} \label{equ: TermC}
   \text{(c)}= \int_0^T \mathbb{E} \left[Y_t^2\right] \mathrm{d} t = \int_0^T \frac{1-e^{-2\kappa t}}{2 \kappa} \mathrm{d}t = \frac{T}{2\kappa} + \frac{ e^{-2\kappa T} - 1 }{4\kappa^2}.
\end{equation}
To derive Term (a), we need to calculate $\mathbb{E}\left[Y_t^2 Y_s^2\right]$. Assume that $s<t$ and denote $M_t = \int_0^t e^{\kappa u } \mathrm{d} W_u $, we have $Y_t = e^{-\kappa t} M_t$, and therefore
\begin{align*}
    \mathbb{E}\left[Y_t^2 Y_s^2\right] &= e^{-2\kappa(t+s)} \mathbb{E}\left[M_t^2 M_s^2\right] = e^{-2\kappa(t+s)} \mathbb{E}\left[(M_t-M_s+M_s)^2 M_s^2\right] \\
    &=e^{-2\kappa(t+s)} \left[ \mathbb{E}\left[(M_t-M_s)^2 M_s^2\right] + 2 \mathbb{E}\left[(M_t-M_s) M_s^3\right] + \mathbb{E}\left[M_s^4\right] \right].
\end{align*}
Because $M_t - M_s = \int_s^t e^{\kappa u } \mathrm{d} W_u$ is a Gaussian random variable with mean 0 and variance
\begin{equation*}
    \mathrm{Var}(M_t - M_s) =\mathbb{E} \left[(M_t - M_s)^2 \right]= \mathbb{E} \left[ \int_s^t e^{\kappa u } \mathrm{d} W_u \right]^2 = \int_s^t \left[ e^{\kappa u }\right]^2\mathrm{d}u = \frac{e^{2\kappa t}-e^{2\kappa s}}{2\kappa},
\end{equation*}
and  $M_t$ has independent increments, we have
\begin{align*}
    \mathbb{E} \left[Y_t^2 Y_s^2\right] &= e^{-2\kappa(t+s)} \left[ \mathbb{E}\left[(M_t-M_s)^2\right]  \mathbb{E}\left[M_s^2\right] + 2 \mathbb{E}\left[M_t-M_s\right] \mathbb{E}\left[ M_s^3\right] + \mathbb{E}\left[M_s^4\right] \right]\\
    &= e^{-2\kappa(t+s)} \left[ \frac{e^{2\kappa t}-e^{2\kappa s}}{2\kappa} \cdot \frac{e^{2\kappa s}-1}{2\kappa} + 0 + 3\left(\frac{e^{2\kappa s}-1}{2\kappa} \right)^2 \right] \\
    &= \frac{ 1+ 2 e^{-2\kappa t + 2\kappa s} - e^{-2\kappa s} - 5e^{-2\kappa t} + 3 e^{-2\kappa t - 2 \kappa s}  }{ 4\kappa^2 }
\end{align*}
when $s<t$. One can similarly write the corresponding formula for the case of $s > t$ and therefore 
\begin{align*}
    \text{(a)} &= \kappa^2 \int_0^T \int_0^T \mathbb{E} \left[Y_t^2 Y_s^2 \right] \mathrm{d} t \mathrm{d} s \\
    &= \frac{1}{4} \left(  T^2 + \frac{T}{\kappa} + \frac{ 10T e^{-2\kappa T  } }{ 2 \kappa } + \frac{ 3e^{-4\kappa T} }{ 4\kappa^2} - \frac{9}{4\kappa^2} + \frac{3 e^{-2\kappa T}}{2\kappa^2}  \right).
\end{align*}
For Term (b), note that
\begin{equation*}
    2 \kappa \mathbb{E} \left[ \int_0^T Y_t^2 \mathrm{d} t \int_0^T Y_t \mathrm{d} W_t \right] = 2 \kappa  \int_0^T \mathbb{E} \left[ Y_s^2 \int_0^T Y_t \mathrm{d} W_t \right]   \mathrm{d} s ,
\end{equation*}
By It\^o's lemma, 
\begin{equation*}
    \mathrm{d} Y_s^2 = 2 Y_s \mathrm{d} Y_s + \mathrm{d}[Y,Y]_s = - 2 \kappa Y_s^2 \mathrm{d} s + 2 Y_s \mathrm{d} W_s +  \mathrm{d} s,
\end{equation*}
which implies that
\begin{equation*}
    Y_s^2 = - 2 \kappa \int_0^s Y_u^2 \mathrm{d} u + 2 \int_0^s Y_u \mathrm{d} W_u + \int_0^s  \mathrm{d} u.
\end{equation*}
Therefore, for $s<T$, with the help of It\^o isometry and \eqref{equ: TermC}, we have
\begin{align*}
    f(s) &= \mathbb{E} \left[ Y_s^2 \int_0^T Y_t \mathrm{d} W_t \right]  \\
    &= \mathbb{E} \left[ \left( - 2 \kappa \int_0^s Y_u^2 \mathrm{d} u + 2 \int_0^s Y_u \mathrm{d} W_u + \int_0^s  \mathrm{d} u \right) \int_0^T Y_t \mathrm{d} W_t \right] \\
    &= - 2 \kappa \int_0^s \mathbb{E} \left( Y_u^2 \int_0^T Y_t \mathrm{d} W_t \right) \mathrm{d} u + 2 \int_0^s \mathbb{E} \left[Y_t^2 \right] \mathrm{d} t + 0 \\
    &= - 2 \kappa \int_0^s f(u) \mathrm{d} u + \frac{s}{\kappa} + \frac{ e^{-2\kappa s} - 1 }{2 \kappa^2},
\end{align*}
and taking derivatives of both sides leads to 
\begin{equation*}
    \frac{\mathrm{d}f}{\mathrm{d}s} = - 2\kappa f(s) + \frac{1}{\kappa} - \frac{e^{-2\kappa s}}{\kappa}.
\end{equation*}
Solving this ordinary differential equation with respect to $f$ with initial condition $f(0) = 0$, we obtain that
\begin{equation*}
    f(s) = \frac{1}{2\kappa^2} - \frac{s e^{-2\kappa s}}{\kappa} - \frac{e^{-2\kappa s}}{2\kappa^2}.
\end{equation*}
Therefore,
\begin{equation*}
    \text{(b)} = 2 \kappa \int_0^T f(s) \mathrm{d} s = \frac{T}{\kappa} + \frac{Te^{-2\kappa T}}{\kappa} + \frac{ e^{-2\kappa T}-1 }{\kappa^2} .
\end{equation*}
Finally, 
\begin{equation}
    \mathbb{E}\left[S(\mathbf{X})_T^{1,1,I}\right]^2 =  \text{(a)} - \text{(b)} + \text{(c)} = \frac{Te^{-2\kappa T}}{4\kappa} + \frac{3e^{-4\kappa T}}{16\kappa^2} - \frac{ 3e^{-2\kappa T} }{8 \kappa^2} - \frac{T}{4\kappa} + \frac{3}{16\kappa^2} + \frac{T^2}{4}. \label{equ: EXMP_ITO_secondmoment}
\end{equation}
Therefore, 
\begin{align*}
    \frac{ \mathbb{E} \left[S(\mathbf{X})_T^{0,I} S(\mathbf{X})_T^{1,1,I}\right] }{ \sqrt{ \mathbb{E} \left[S(\mathbf{X})_T^{0,I}\right]^2 \mathbb{E}\left[S(\mathbf{X})_T^{1,1,I}\right]^2 } } &= \frac{ \mathbb{E} \left[S(\mathbf{X})_T^{1,1,I}\right] }{ \sqrt{ \mathbb{E}\left[S(\mathbf{X})_T^{1,1,I}\right]^2 } }  \\
    &= \frac{ -2\kappa T - e^{-2\kappa T} + 1 }{ 
 \sqrt{  4\kappa T e^{-2\kappa T} + 3 e^{-4\kappa T} - 6 e^{-2\kappa T} - 4 \kappa T + 3 + 4\kappa^2 T^2 } } ,
\end{align*}
where the 0-th order of signature is defined as 1.

\paragraph{Stratonovich Signature.} The Stratonovich integral and the It\^o integral are related by
\begin{equation*}
    \int_0^t A_s \circ \mathrm{d} B_s = \int_0^t A_s \mathrm{d} B_s + \frac{1}{2} [A, B ]_t.
\end{equation*}
Therefore, 
\begin{equation*}
   S(\mathbf{X})_T^{1,S} = \int_0^T 1 \circ \mathrm{d} Y_t = \int_0^T 1  \mathrm{d} Y_t + \frac{1}{2} [1, Y ]_T = \int_0^T 1  \mathrm{d} Y_t =  S(\mathbf{X})_T^{1,I} = Y_T,
\end{equation*}
and 
\begin{equation*}
   S(\mathbf{X})_T^{1,1,S} = \int_0^T S(\mathbf{X})_T^{1,S} \circ \mathrm{d} Y_t = \int_0^T Y_t \circ \mathrm{d} Y_t = \int_0^T Y_t \mathrm{d} Y_t + \frac{1}{2} [Y, Y ]_T = S(\mathbf{X})_T^{1,1,I} + \frac{T}{2},
\end{equation*}
where we use the fact that $ [1, Y ]_T = 0$ and $ [Y, Y ]_T = T$. Now by \eqref{equ: EXMP_ITO_firstmoment} and \eqref{equ: EXMP_ITO_secondmoment}, we have
\begin{equation*}
     \mathbb{E}\left[S(\mathbf{X})_T^{1,1,S}\right] = \mathbb{E}\left[S(\mathbf{X})_T^{1,1,I}\right] + \frac{T}{2} = \frac{ 1-e^{-2\kappa T} }{4\kappa},
\end{equation*}
and
\begin{align*}
     \mathbb{E}\left[S(\mathbf{X})_T^{1,1,S}\right]^2 &= \mathbb{E} \left[S(\mathbf{X})_T^{1,1,I} + \frac{T}{2} \right]^2 \\
     &= \mathbb{E}\left[S(\mathbf{X})_T^{1,1,I}\right]^2 + T \mathbb{E}\left[S(\mathbf{X})_T^{1,1,I}\right] + \frac{T^2}{4} = \frac{3 ( 1-e^{-2\kappa T})^2}{16\kappa^2}.
\end{align*}
Therefore, 
\begin{equation*}
     \frac{ \mathbb{E} \left[S(\mathbf{X})_T^{0,S} S(\mathbf{X})_T^{1,1,S}\right] }{ \sqrt{ \mathbb{E}\left[S(\mathbf{X})_T^{0,S}\right]^2 \mathbb{E}\left[S(\mathbf{X})_T^{1,1,S}\right]^2 } } = \frac{\sqrt{3}}{3}.\tag*{\Halmos}
\end{equation*}
\endproof

\proof{Proof of Proposition \ref{prop: sufficient_irrepresentable_tight}.}
    Let $a = \# A_1^*$ and $b = \# A_1^{*c}$. Under the equal inter-dimensional correlation assumption, we have $\Sigma_{A^*, A^*} = (1-\rho) I_a+\rho \mathbf{1}_a \mathbf{1}_a^\top$, where $I_a$ is an $a\times a$ identity matrix and $\mathbf{1}_a$ is an $a$-dimensional all-one vector. In addition, $\Sigma_{A^{*c}, A^{*}} = \rho \mathbf{1}_b \mathbf{1}_a^\top$, where $\mathbf{1}_b$ is a $b$-dimensional all-one vector. By the Sherman--Morrison formula, 
    \begin{equation*}
        \Sigma_{A^*, A^*}^{-1} = \frac{1}{1-\rho} I_a - \frac{\rho}{(1-\rho)(1+(a-1)\rho) }\mathbf{1}_a \mathbf{1}_a^\top.
    \end{equation*}
    Therefore, since all true beta coefficients are positive, we have
    \begin{equation*}
        \Sigma_{A^{*c}, A^{*}}  \Sigma_{A^*, A^*}^{-1} \mathrm{sign}({\boldsymbol \beta}_{A^*}) = \frac{ a \rho}{ 1 + (a-1) \rho } \mathbf{1}_a . 
    \end{equation*}
    Hence, the irrepresentable condition 
    \begin{equation*}
        \left\| \Sigma_{A^{*c}, A^{*}}  \Sigma_{A^*, A^*}^{-1} \mathrm{sign}({\boldsymbol \beta}_{A^*}) \right\| =  \frac{ a |\rho|}{ 1 + (a-1) \rho }  < 1 
    \end{equation*}
    holds if and only if $\frac{ a |\rho|}{ 1 + (a-1) \rho } < 1$. One can easily verify that this holds if $ \rho  \in( -\frac{1}{2 \# A_1^* }, 1)$, and does not hold if $\rho \in ( -\frac{1}{\# A_1^* } , -\frac{1}{2 \# A_1^* } ]$. This completes the proof.     
\Halmos\endproof

\proof{Proof of Theorem \ref{th:samplelasso3}.}
For $\xi = \min\left\{g_\Sigma^{-1}\left(\frac{\gamma}{\zeta(2+2\alpha\zeta+\gamma)}\right),g_\Sigma^{-1}\left(\frac{C_{\min}}{2\sqrt{p}}\right)\right\}>0$, Lemmas \ref{th:samplelasso1} and \ref{th:samplelasso2} imply that
\begin{align*}
    &\mathbb{P}\left( \Lambda_{\min}(\hat{\Delta}_{A^*A^*})\geq \frac{1}{2} C_{min} \right) \geq 1-\frac{4p^4\sigma_{\max}^4(\sigma_{\min}^4+K)}{N\xi^2\sigma_{\min}^4} ,\\&\mathbb{P}\left(\left\|\hat{\Delta}_{A^{*c}A^{*}}\hat{\Delta}_{A^{*}A^{*}}^{-1} \right\|_{\infty} \leq 1-\frac{\gamma}{2} \right) \geq 1-\frac{4p^4\sigma_{\max}^4(\sigma_{\min}^4+K)}{N\xi^2\sigma_{\min}^4} .
\end{align*}
Hence, 
\begin{equation}\label{eq:wainright}
    \mathbb{P}\left(\Lambda_{\min}(\hat{\Delta}_{A^*A^*})\geq \frac{1}{2} C_{\min},\left\|\hat{\Delta}_{A^{*c}A^{*}}\hat{\Delta}_{A^{*}A^{*}}^{-1} \right\|_{\infty} \leq 1-\frac{\gamma}{2}\right) 
 \geq 1-\frac{8p^4\sigma_{\max}^4(\sigma_{\min}^4+K)}{N\xi^2\sigma_{\min}^4}.
\end{equation}
Equation \eqref{eq:wainright} gives the probability that the conditions of \citet[Theorem 1]{wainwright2009sharp} hold. Therefore, applying \citet[Theorem 1]{wainwright2009sharp} yields the result.
\Halmos\endproof

\clearpage
\singlespacing
\clearpage
\bibliography{mybib}

\end{document}